\documentclass[10pt, journal,compsoc]{IEEEtran} 

\usepackage{amsmath,amssymb}
\usepackage{color}
\usepackage{graphics,graphicx}
\usepackage{multirow}
\usepackage{subfigure}
\usepackage{array}
\usepackage{adjustbox}
\usepackage{booktabs}
\usepackage{flushend}

\ifCLASSOPTIONcompsoc
\usepackage[nocompress]{cite}
\else
\usepackage{cite}
\fi

\definecolor{limegreen}{rgb}{0.2, 0.8, 0.2}
\newcolumntype{C}[1]{>{\centering\let\newline\\\arraybackslash\hspace{0pt}}m{#1}}
\newcolumntype{L}[1]{>{\raggedright\let\newline\\\arraybackslash\hspace{0pt}}m{#1}}

\newcommand{\fs}{0.15}
\newcommand{\as}{0.10}
\newcommand{\sep}{1.5}
\DeclareMathOperator*{\argmax}{arg\,max}

\makeatletter
\def\paragraph{\@startsection{paragraph}{4}{\z@}{1.5ex plus 1.5ex minus 0.5ex}%
	{0ex}{\normalfont\normalsize\bfseries}}
\def\@IEEEsectpunct{.\ \,}
\let\@fnsymbol\@arabic
\makeatother

\begin{document}
	
\title{Long-term Temporal Convolutions \\for Action Recognition}

        
\author{G\"{u}l~Varol, Ivan~Laptev, and~Cordelia~Schmid,~\IEEEmembership{Fellow,~IEEE}
\IEEEcompsocitemizethanks{\IEEEcompsocthanksitem G\"{u}l Varol and Ivan Laptev are with Inria, WILLOW project-team, D\'{e}partement d'Informatique de l'\'{E}cole Normale Sup\'{e}rieure, ENS/Inria/CNRS UMR 8548, Paris, France.\protect\\
E-mail: gul.varol@inria.fr
\IEEEcompsocthanksitem Cordelia Schmid is with Inria, Thoth project-team, Inria Grenoble Rh\^{o}ne-Alpes, Laboratoire Jean Kuntzmann, France.}
}

\IEEEtitleabstractindextext{
\begin{abstract}
Typical human actions last several seconds and exhibit characteristic spatio-temporal structure. Recent methods attempt to capture this structure and learn action representations with convolutional neural networks. Such representations, however, are typically learned at the level of a few video frames failing to model actions at their full temporal extent. In this work we learn video representations using neural networks with long-term temporal convolutions (LTC). We demonstrate that LTC-CNN models with increased temporal extents improve the accuracy of action recognition. We also study the impact of different low-level representations, such as raw values of video pixels and optical flow vector fields and demonstrate the importance of high-quality optical flow estimation for learning accurate action models.  We report state-of-the-art results on two challenging benchmarks for human action recognition UCF101~(92.7\%) and HMDB51 (67.2\%).
\end{abstract}

}

\maketitle

\IEEEdisplaynontitleabstractindextext
\IEEEpeerreviewmaketitle

\IEEEraisesectionheading{\section{Introduction}\label{sec:introduction}}

\IEEEPARstart{H}{uman}
 actions and events can be seen as spatio-temporal objects. Such a view finds 
 support both in psychology~\cite{Tversky02} and in computer vision approaches 
 to action recognition in video~\cite{Laptev08,Niebles08,Schuldt04,wang_idt}. 
 Successful methods for action recognition, indeed, share similar techniques 
 with object recognition and represent actions by statistical models of local 
 video descriptors. Differently to objects, however, actions are characterized 
 by the temporal evolution of appearance governed by motion. Consistent with 
 this fact, motion-based video descriptors such as HOF and MBH~\cite{Laptev08,wang_idt} 
 as well as recent CNN-based motion representations~\cite{simonyan_twostream} 
 have shown most gains for action recognition in practice.
 
\begin{figure}[t!]
  \centering
  \includegraphics[width=0.7\linewidth]{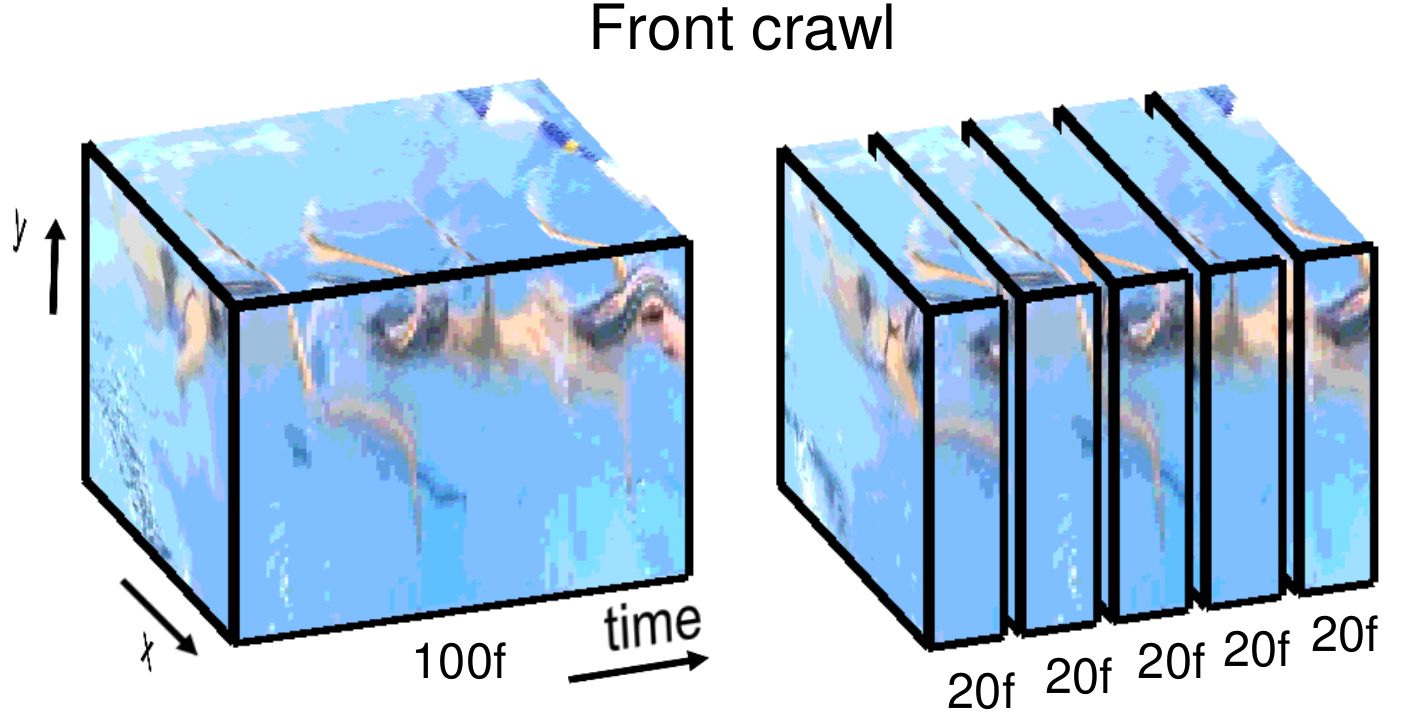}
  \hspace{2.5cm}(a)
  \hspace{2.5cm}(b) \\
  \includegraphics[width=0.7\linewidth]{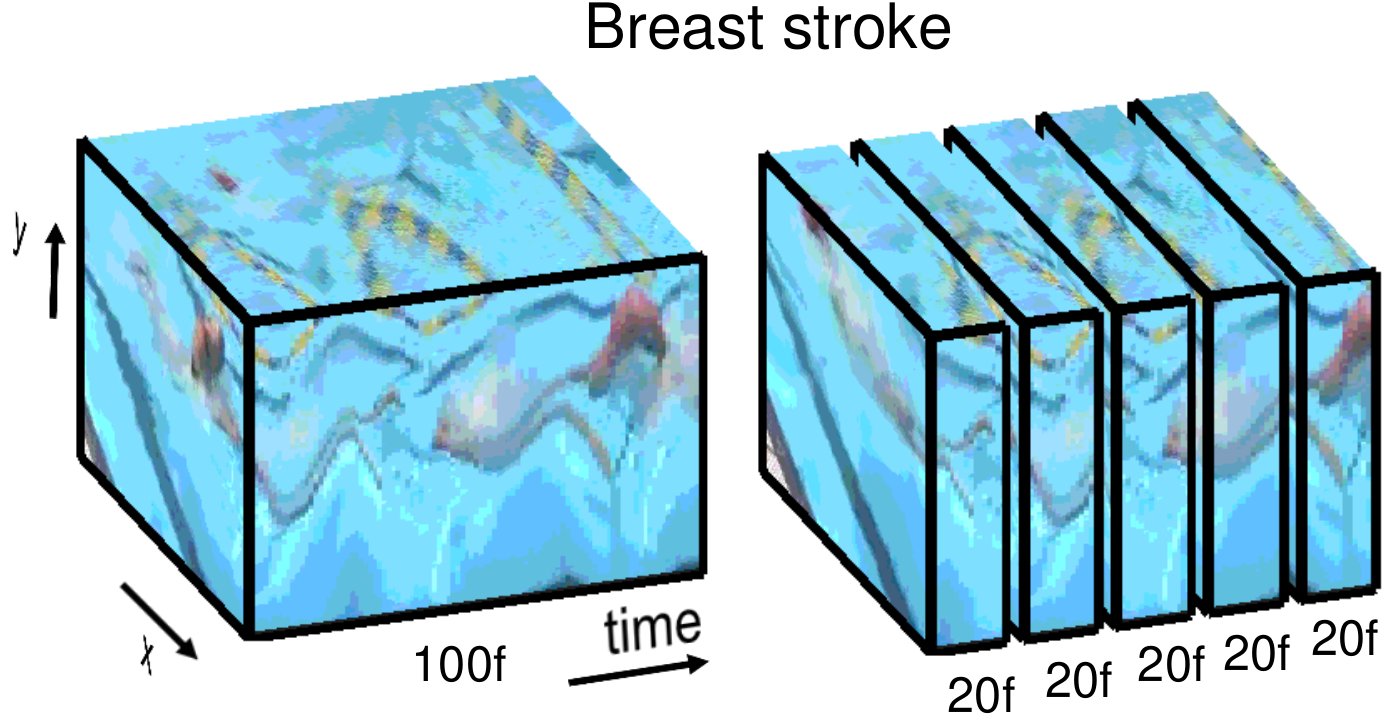}
  \hspace{2.5cm}(c)
  \hspace{2.5cm}(d)
  \mbox{}\vspace{-.3cm}\\
  \caption{Video patches for two classes of swimming actions. (a),(c): 
  Actions often contain characteristic, class-specific space-time patterns 
  that last for several seconds. (b),(d):~Splitting videos into short temporal 
  intervals is likely to destroy such patterns making recognition more difficult. 
  Our neural network with Long-term Temporal Convolutions (LTC) learns video 
  representations over extended periods of time.}
  \mbox{}\vspace{-.6cm}\\
  \label{fig:teaser}
\end{figure}

The recent rise of convolutional neural networks (CNNs) convincingly demonstrates 
the power of learning visual representations~\cite{krizhevsky_imagenet}. Equipped 
with large-scale training datasets~\cite{Deng09,Zhou14}, CNNs have quickly taken 
over the majority of still-image recognition tasks such as object, scene and face recognition~\cite{Girshick14,Taigman14,Zhou14}.
Extensions of CNNs to action recognition in video have been proposed in several 
recent works~\cite{karpathy_sports1m,simonyan_twostream,tran_c3d}. Such methods, 
however, currently show only moderate improvements over earlier methods using 
hand-crafted video features~\cite{wang_idt}.

Current CNN methods for action recognition often extend CNN architectures for 
static images~\cite{krizhevsky_imagenet} and learn action representations for 
short video intervals ranging from 1 to 16 frames~\cite{karpathy_sports1m,simonyan_twostream,tran_c3d}. 
Yet, typical human actions such as hand-shaking and drinking, as well as cycles 
of repetitive actions such as walking and swimming often last several seconds 
and span tens or hundreds of video frames. As illustrated in Figure~\ref{fig:teaser}(a),(c), 
actions often contain characteristic patterns with specific spatial as well 
as {\em long-term} temporal structure. Breaking this structure into short 
clips (see Figure~\ref{fig:teaser}(b),(d)) and aggregating video-level 
information by the simple average of clip scores~\cite{simonyan_twostream,tran_c3d} 
or more sophisticated schemes such as LSTMs~\cite{Donahue15} is likely to be suboptimal.

\begin{figure*} [t]
\begin{center}
  \includegraphics[width=0.77\linewidth]{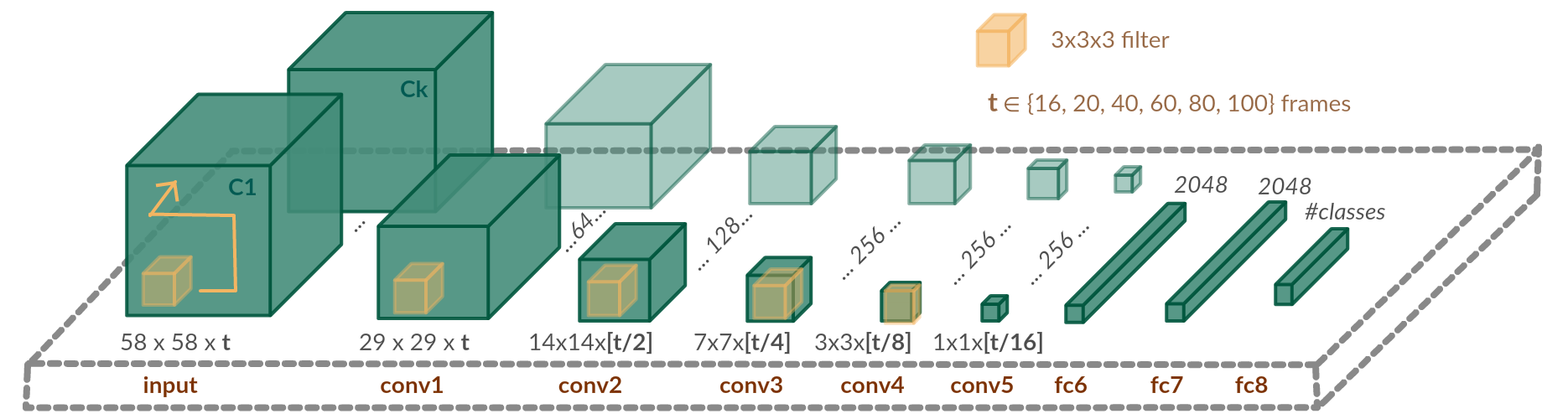}
  \vspace{-.6cm}\\
\end{center}
   \caption{Network architecture. Spatio-temporal convolutions with 3x3x3 
   	filters are applied in the first 5 layers of the network.
     Max pooling and ReLU are applied in between all convolutional layers. 
     Network input channels $C1...Ck$ are defined for different temporal
     resolutions $t\in\{20,40,60,80,100\}$ and either two-channel motion 
     ({\em{flow-x}}, {\em{flow-y}}) or three-channel appearance ({\em{R}},{\em{G}},{\em{B}}).
     The spatio-temporal resolution of convolution layers decreases with the pooling operations.
}
\mbox{}\vspace{-.9cm}\\
\label{fig:architecture}
\end{figure*}

In this work, we investigate the learning of long-term video representations. 
We consider space-time convolutional neural networks~\cite{Ji10,Taylor10,tran_c3d} 
and study architectures with Long-term Temporal Convolutions (LTC), see Figure~\ref{fig:architecture}.
To keep the complexity of networks tractable, we increase the temporal extent 
of representations at the cost of decreased spatial resolution. We also study 
the impact of different low-level representations, such as raw values of 
video pixels and optical flow vector fields. 
Our experiments confirm the advantage of motion-based representations and 
highlight the importance of good quality motion estimation for learning 
efficient representations for human action recognition. We report state-of-the-art 
performance on two recent and challenging human action benchmarks: UCF101 and HMDB51.

The contributions of this work are twofold. We demonstrate {\em (i)} the 
advantages of long-term temporal convolutions and {\em (ii)}  the importance 
of high-quality optical flow estimation for learning accurate video 
representations for human action recognition. In the remaining part of 
the paper we discuss related work in Section~\ref{sec:relatedwork}, 
describe space-time CNN architectures in Section~\ref{sec:method} and 
present an extensive experimental study of our method in Section~\ref{sec:experiments}.
Our implementation and pre-trained CNN models (compatible with Torch) 
are available on the project web page~\cite{projectpage}. 
\vspace{-.3cm}
\section{Related Work}
\label{sec:relatedwork}
Action recognition in the last decade has been dominated 
by local video features~\cite{Schuldt04,Laptev08,wang_idt} 
aggregated with Bag-of-Features histograms~\cite{Csurka04} 
or Fisher Vector representations~\cite{Perronnin10}. While 
typical pipelines resemble earlier methods for object 
recognition, the use of local motion features, in particular
Motion Boundary Histograms~\cite{wang_idt}, has been 
found important for action recognition in practice. Explicit 
representations of the temporal structure of actions have 
rarely beed used with some exceptions such as the recent work~\cite{Fernando15}.

Learning visual representations with CNNs~\cite{krizhevsky_imagenet,lecun-89e} 
has shown clear advantages over ``hand-crafted'' features 
for many recognition tasks in static images~\cite{Girshick14,Taigman14,Zhou14}. 
Extensions of CNN representations to action recognition in video 
have been proposed in several recent works~\cite{Donahue15,Ji10,karpathy_sports1m,
simonyan_twostream,Taylor10,tran_c3d,wang_tdd,
wang_deeptwostream,bilen_dynamic,feichtenhofer_twostreamfusion}. 
Some of these methods encode single video frames with static CNN features~\cite{Donahue15,karpathy_sports1m,simonyan_twostream}. 
Extensions to short video clips where video frames are treated 
as multi-channel inputs to 2D CNNs have also been investigated 
in~\cite{karpathy_sports1m,simonyan_twostream,
feichtenhofer_twostreamfusion,wang_deeptwostream}. 

Learning CNN representations for action recognition has been addressed 
for raw pixel inputs and for pre-computed optical flow features. 
Consistent with previous results obtained with hand-crafted representations, 
motion-based CNNs typically outperform CNN representations learned 
for RGB inputs~\cite{simonyan_twostream,wang_deeptwostream}. In 
this work we investigate multi-resolution representations of motion 
and appearance where for motion-based CNNs we demonstrate the 
importance of high-quality optical flow estimation. Similar findings 
have been recently confirmed by~\cite{zhang_emv}, where the authors 
transfer knowledge from high quality optical flow algorithms to 
motion vector encoding representation.

Most of the current CNN methods use architectures with 2D convolutions, 
enabling shift-invariant representations in the image plane. Meanwhile, 
the invariance to translations in time is also important for action 
recognition since the beginning and the end of actions is unknown in general. 
CNNs with 3D spatio-temporal convolutions address this issue and provide 
a natural extension of 2D CNNs to video. 3D CNNs have been investigated 
for action recognition in~\cite{Ji10, karpathy_sports1m,Taylor10,tran_c3d}. 
All of these methods, however, learn video representations for RGB input. 
Moreover, they typically consider very short video intervals, for example, 
16-frame video clips are used in~\cite{tran_c3d} and 2, 7, 15 frames in~\citeleft\citen{Taylor10}\citepunct\citen{Ji10}\citepunct\citen{karpathy_sports1m}\citeright~respectively. 
In this work we extend 3D CNNs to significantly longer temporal 
convolutions that enable action representation at their full 
temporal scale.  We also explore the impact of optical flow input. 
Both of these extensions show clear advantages in our experimental 
comparison to previous methods.
\section{Long-term Temporal Convolutions}
\label{sec:method}

In this section we first present the network architecture. We then
specify the different inputs to networks used in this work.
We finally provide details on learning and testing procedures.

\begin{figure*} [t]
\centering
\begin{tabular}{cc}  \footnotesize
  \begin{tabular}{cccc}
    \frame{\includegraphics[trim = 15mm 0mm 0mm 0mm, clip, width=0.145\linewidth] {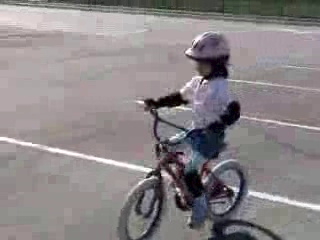}} &
    \frame{\includegraphics[trim = 15mm 0mm 0mm 0mm, clip, width=0.145\linewidth]{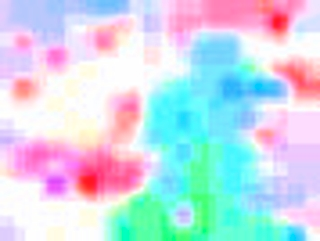}}&
    \frame{\includegraphics[trim = 15mm 0mm 0mm 0mm, clip, width=0.145\linewidth]{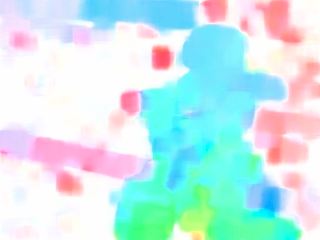}} &
    \frame{\includegraphics[trim = 15mm 0mm 0mm 0mm, clip, width=0.145\linewidth]{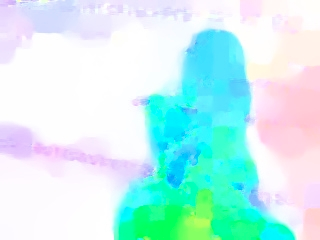}} \\
    RGB &  MPEG flow~\cite{kantorov_efficient}& Farneback~\cite{Farneback}& Brox~\cite{Brox2004} \\
  \end{tabular}
  &
  \begin{adjustbox}{width=0.24\linewidth}
  \begin{tabular}{lcc}
    \toprule
    Input & Clip & Video\\
    \midrule
    RGB		 	&	 57.0	& 59.9 \\
    MPEG flow 	&	 58.5	& 63.8 \\
    Farneback 	&	 66.3	& 71.3 \\
    Brox 		&	 \textbf{74.8}	& \textbf{79.6} \\
    \bottomrule
    \multicolumn{3}{c}{}
  \end{tabular}
  \end{adjustbox}
  \mbox{}\vspace{-.21cm}\\
\end{tabular}
\caption{Illustration of the three optical flow methods and 
	comparison of corresponding recognition performance. From 
	left to right: original image, MPEG, Farneback and Brox 
	optical flow. The color coding indicates the orientation 
	of the flow. The table on the right presents accuracy of 
	action recognition in UCF101 (split~1) for different 
	inputs. Results are obtained with 60f networks and training 
	from scratch (see text for more details).}
\mbox{}\vspace{-.4cm}\\
\label{fig:flowsamples}
\end{figure*}

\subsection{Network architecture}
Our network  architecture with long-term temporal
convolutions is illustrated in Figure~\ref{fig:architecture}. 
The network has 5 space-time convolutional layers with 
64, 128, 256, 256 and 256 filter response maps, followed by 3 fully 
connected layers of sizes 2048, 2048 and
the number of classes. Following~\cite{tran_c3d} we use $3\times3\times3$ 
space-time filters for all convolutional layers. 
Each convolutional layer is followed by a rectified linear unit
(ReLU) and a space-time max pooling layer.  Max pooling filters are of size
$2\times2\times2$ except in the first layer, where it is $2\times2\times1$.
The size of convolution output is kept constant by padding 1 pixel in all three
dimensions. Filter stride for all dimensions
is 1 for convolution and 2 for pooling operations.  We use dropout
for the first two fully connected layers. Fully connected layers are
followed by ReLU layers. Softmax layer at the end of
the network outputs class scores.

\subsection{Network input}
To investigate the impact of long-term temporal convolutions, we
here study network inputs with different temporal extents.
We depart from the recent C3D work~\cite{tran_c3d} and
first compare inputs of 16 frames (16f) and 60 frames (60f).
We then systematically analyze implications of the increased temporal
and spatial resolutions for input signals in terms of motion and
appearance.
For the 16-frame network we crop input patches of size $112\times112\times16$ 
from videos with spatial resolution $171\times128$ pixels.
We choose this baseline architecture to enable direct comparison
with~\cite{tran_c3d}. For the 60-frames networks we decrease spatial resolution
to preserve network complexity and use input patches of size $58\times58\times60$
randomly cropped from videos rescaled to $89\times67$ spatial resolution.

As illustrated in Figure~\ref{fig:architecture}, the temporal 
resolution in our 60f network corresponds to 60, 30, 15, 7 and 3 
frames for each of the five convolutional layers. In comparison, 
the temporal resolution of the 16f network is reduced more 
drastically to 16, 8, 4, 2 and 1 frame at each convolutional layer.
We believe that preserving the temporal resolution at higher 
convolutional layers should enable learning more complex temporal patterns.
The space-time resolution for the outputs of the fifth convolutional 
layers is $3\times3\times1$ and $1\times1\times3$ for the 16f and 60f networks respectively.
The two networks have a similar number of parameters in the \emph{fc6}
layer and the same number of parameters in all other layers.
For a systematic study of networks with different input resolutions 
we also evaluate the effect of increased temporal resolution 
$t \in \{20,40,60,80,100\}$ and varying spatial resolution of $\{58\times58,71\times71\}$ pixels.

In addition to the input size, we experiment with different types of 
input modalities. First, as in~\cite{tran_c3d}, we use raw RGB values from 
video frames as input. To explicitly learn motion representations,
we also use flow fields in $x$ and $y$ directions as input to our networks. 
Flow is computed for original videos. To maintain correct flow values 
for network inputs with reduced spatial resolution, the magnitude of 
the flow is scaled by the factor of spatial subsampling.
In other words, if a point moves 2 pixels in a $320\times240$ video frame, 
its motion will be 1 pixel when the frame is resized to $160\times120$ resolution.
Moreover, to center the input data, we follow 
\cite{simonyan_twostream} and subtract the mean flow vector for each frame.

To investigate the dependency of action recognition on the quality of
motion estimation, we experiment with three types of flow inputs obtained
either directly from the video encoding, referred to as MPEG
flow~\cite{kantorov_efficient},
or from two optical flow estimators, namely Farneback~\cite{Farneback} and
Brox~\cite{Brox2004}. Figure~\ref{fig:flowsamples} shows results for
the three flow algorithms.
MPEG flow is a fast substitute for optical flow which we obtain from the
original video encoding.
Such flow, however, has low spatial resolution.
It also misses flow
vectors at some frames (I-frames) which we interpolate from neighboring 
frames.
Farneback flow is also relatively fast and obtains rather noisy flow 
estimates. The approach of Brox flow is the most sophisticated of the three 
and is known to perform well in various flow estimation benchmarks.
 
\subsection{Learning}
We train our networks on the training set of
each split independently for both UCF101 and HMDB51 datasets, which
contain 9.5K and 3.7K videos, respectively. We use stochastic gradient 
descent applied to mini-batches with negative log likelihood criterion. For 16f networks we
use a mini-batch size of 30 video clips. We reduce the batch size to 15 
video clips for 60f networks, and 10 clips for 100f networks due to 
limitations of our GPUs. The initial learning rate
for networks learned from scratch is $3\times10^{-3}$ and $3\times10^{-4}$ 
for networks fine-tuned from pre-trained models. For
UCF101, the learning rate is decreased twice with a factor of $10^{-1}$. 
For 16f networks, the first decrease is after 80K iterations and the 
second one after 45K additional iterations. The optimization is completed after 20K more
iterations. Convergence is faster for HMDB51, so the learning rate is
decreased once after 60K iterations and completed after 10K more
iterations. These numbers are doubled for 60f networks and tripled for 100f networks,
since their batch sizes are twice and three times smaller compared to 16f nets.
The above schedule is used together with 0.9 dropout ratio. Our 
experimental setups with 0.5 dropout ratio have less iterations due 
to faster convergence. The momentum is set to 0.9 and weight decay 
is initialized with $5\times 10^{-3}$ and reduced by a factor 
of $10^{-1}$ at every decrease of the learning rate.

Inspired by the random spatial cropping during training,
we apply the corresponding augmentation to the temporal dimension 
as in~\cite{simonyan_twostream}, which we call
{\em random clipping}. During training, given an input video, we 
randomly select a point $(x, y, t)$ to sample a video clip of fixed size.
A common alternative is to preprocess the data by using a sliding 
window approach to have pre-segmented clips of fixed size; however, 
this approach limits the amount of data when the windows are not 
overlapped as in~\cite{tran_c3d}. Another data augmentation method 
that we evaluate is to have a multiscale cropping similar to~\cite{wang_deeptwostream}. 
For this, we randomly select a coefficient for width and height 
separately from (1.0, 0.875, 0.75, 0.66) and resize the cropped region 
to the size of the network input. Finally, we horizontally flip the input with 50\% probability.

At test time, a video is divided into $t$-frame clips with a 
temporal stride of 4 frames. Each clip is further tested
with 10 crops, namely the 4 corners and the center, together with their
horizontal flips. The video score is obtained by averaging over clip scores
and crop scores. If the number of frames in a video is less than the clip
size, we pad the input by repeating the last frames to fill the
missing volume.

\section{Experiments}
\label{sec:experiments}

We perform experiments on two widely used and challenging benchmarks for action recognition: 
UCF101 and HMDB51 (Sec.~\ref{ss:datasets}).  
We first examine the effect of network parameters (Sec.~\ref{ss:eval_param}).
We then compare to the state-of-the-art (Sec.~\ref{ss:comp_soa}) 
and present a visual analysis of the spatio-temporal filters
(Sec.~\ref{ss:visualization}). Finally we report runtime analysis (Sec.~\ref{ss:runtime}).

\vspace{-.3cm}
\subsection{Datasets and evaluation metrics}
\label{ss:datasets}

UCF101 ~\cite{UCF101} is a widely-used benchmark for action recognition 
with 13K clips from YouTube videos lasting 7 seconds on average. 
The total number of frames is 2.4M distributed among 101 categories. 
The videos have spatial resolution of $320\times240$ pixels and 25 fps frame rate.

The HMDB51 dataset~\cite{HMDB51} consists of 7K videos of 51 actions. 
The videos have $320\times240$ pixels spatial resolution and 30 fps 
frame rate. Although this dataset has been considered a large-scale 
benchmark for action recognition for the past few years, the amount 
of data for learning deep networks is limited.

We rely on two evaluation metrics. The first one measures 
per-clip accuracy, i.e.\ we assign each clip the class label with
the maximum softmax output and measure the number of
correctly assigned labels over all clips. 
The second metric measures video accuracy, i.e.\ the standard 
evaluation protocol. To obtain a video score we
average the per-clip softmax scores and take the maximum value of
this average as class label. We average over all videos to obtain video
accuracy. We report our final results according to the standard evaluation 
protocol, which is the mean video accuracy across the three test splits. 
To evaluate the network parameters we use the first split.

\vspace{-.3cm}
\subsection{Evaluation of LTC network parameters}
\label{ss:eval_param}

In the following we first examine the impact of optical flow
and data augmentation. We then evaluate gains provided by long-term
temporal convolutions for the best flow and data augmentation
techniques by comparing 16f and 60f networks. 
We also investigate the advantage of pre-training on one dataset (UCF101)
and fine-tuning on a smaller dataset (HMDB51).
Furthermore, we study the effect of systematically increased temporal
resolution for flow and RGB inputs as well as the combination of networks.  

\paragraph*{Optical flow}
The impact of the flow quality on action
recognition and a comparison to RGB is shown in
Figure~\ref{fig:flowsamples} for UCF101 (split~1). 
The network is trained from scratch and with a 60-frame video volume as 
input. 
We first observe that even the low-quality MPEG flow outperforms RGB.
The increased  quality of optical flow leads to further improvements.
The use of Brox flow allows nearly 20\% increase in performance.
The improvements are consistent when classifying individual clips and full videos.
This suggests that action recognition is easier to learn from
motion compared to raw pixel values. While results in Figure~\ref{fig:flowsamples} were 
obtained for 60f networks, the same holds for 16f networks (see Table~\ref{table:ucf_16_60}).
We also conclude that the high accuracy of optical flow estimation plays an
important role for learning competitive video representations for action
recognition.
Given the results in Figure~\ref{fig:flowsamples}, we choose Brox flow for all
remaining experiments in this paper. 

\paragraph*{Data augmentation}
Table~\ref{table:augmentation}
demonstrates the contribution of data augmentation 
when training a large CNN with limited amount of
data. Our baseline uses sliding window clips with 75\% overlap
and a dropout of 0.5 during training. We gain 3.1\% with random
clipping, 1.6\% with multiscale cropping and 2\% with higher dropout ratio.
When combined, the data augmentation and a higher dropout results in a
4\% gain for video classification on  UCF101 split~1.
High dropout, multiscale cropping and random clipping are used in the 
remaining experiments, unless stated
otherwise. 
\begin{table}
\begin{center}
\begin{tabular}{lcc}
\toprule
Method & Clip accuracy & Video accuracy\\
\midrule
Baseline augmentation & 71.6 & 76.5 \\
Random clipping & 74.8 & 79.6 \\
Multiscale cropping & 72.5 & 78.1 \\
High dropout (0.9) & 74.4 & 78.5 \\
Combined & \textbf{76.3} & \textbf{80.5} \\
\bottomrule
\end{tabular}
\end{center}
\caption{Data augmentations on UCF101 (split~1). All results are with 
	60-frame Brox flow and training from scratch. All three modifications 
	(random clipping, multiscale cropping and high dropout) give an 
	improvement when used alone, the best performance is obtained when combined.} 
\mbox{}\vspace{-1.7cm}\\
\label{table:augmentation}
\end{table}

\begin{table}[b]
	\mbox{}\vspace{-0.9cm}\\
	\begin{center}
		\begin{tabular}{ccccC{1cm}C{1cm}L{1cm}}
			\toprule
			Input& MS & D & Test & 16f & 60f & gain  \\
			\midrule
			RGB		& x          		& 0.5 & Clip & 48.4 & 57.0 & + 8.6 \\
			& & &  Video & 51.9 & 59.9 & + 8.0 \\ \midrule
			Flow	& x          	 	 & 0.5 & Clip & 66.8 & 74.8 & + 8.0 \\
			& & &  Video & 77.4 & 79.6 & + 2.2 \\ \midrule
			Flow 	& \checkmark & 0.9 &  Clip & 67.1 & \textbf{76.3} & + 9.1 \\
			& & & Video & 78.7 & \textbf{80.5} & + 1.8 \\
			\bottomrule
		\end{tabular}
	\end{center}
	\caption{Results for networks with different temporal resolutions
		and under variation of data augmentation (MS: multiscale cropping) and dropout (D) for
		UCF101 (split~1), trained from scratch. Random clipping is used in
		all experiments. Evaluations are on individual clips and on
		full videos.} 
	\mbox{}\vspace{-1.3cm}\\
	\label{table:ucf_16_60}
\end{table}
\begin{table}[b]
	\begin{center}
		\begin{tabular}{ccC{0.6cm}C{0.6cm}L{0.8cm}C{1.5cm}}
			\toprule 
			Pre-training & Test & 16f & 60f & gain & {\scriptsize 2D CNN\cite{simonyan_twostream}} \\
			\midrule 
			x & Clip & 37.0 & 52.6 & + 15.6 & \\
			& Video & 43.9 & 52.9 & +  9.0 & 46.6 \\
			\midrule
			\checkmark & Clip & 40.6 & \textbf{56.1} & + 15.5 & \\
			& Video & 48.3 & \textbf{57.1} & +  8.8 & 49.0 \\
			\bottomrule
		\end{tabular}
	\end{center}
	\caption{Results for networks with different temporal resolutions for 
		HMDB51 (split~1) with or without pre-training on UCF101. Flow input, random clipping, multiscale cropping 
		and 0.9 dropout are used in all setups.}
	\mbox{}\vspace{-1.5cm}\\
	\label{table:hmdb_16_60}
\end{table}
\vspace{-0.2cm}
\paragraph*{Comparison of 16f and 60f networks}
Our 16-frame and 60-frame networks have similar complexity in terms of 
input sizes and the number of network parameters (see Section~\ref{sec:method}).
Moreover, the 16-frame network resembles the C3D architecture and enables 
direct comparison with~\cite{tran_c3d}.
We therefore study the gains provided by the 60-frame inputs before
analyzing performance with systematically increasing temporal
resolution (from 20 to 100 frames by steps of 20) in the next paragraph. 

\begin{figure*}
	\centering
	\begin{tabular}{c@{\hspace{-.05cm}}c}
	\begin{adjustbox}{valign=t}
		\begin{tabular}{ccc}
		{\em (a) overall clip accuracy} & {\em (b) overall video accuracy} & {\em (c) per-action clip accuracy}
		\\ \mbox{}\vspace{-.4cm}\\
		\includegraphics[trim=0.5cm 0cm 1cm 1cm, clip, width=0.22\linewidth]{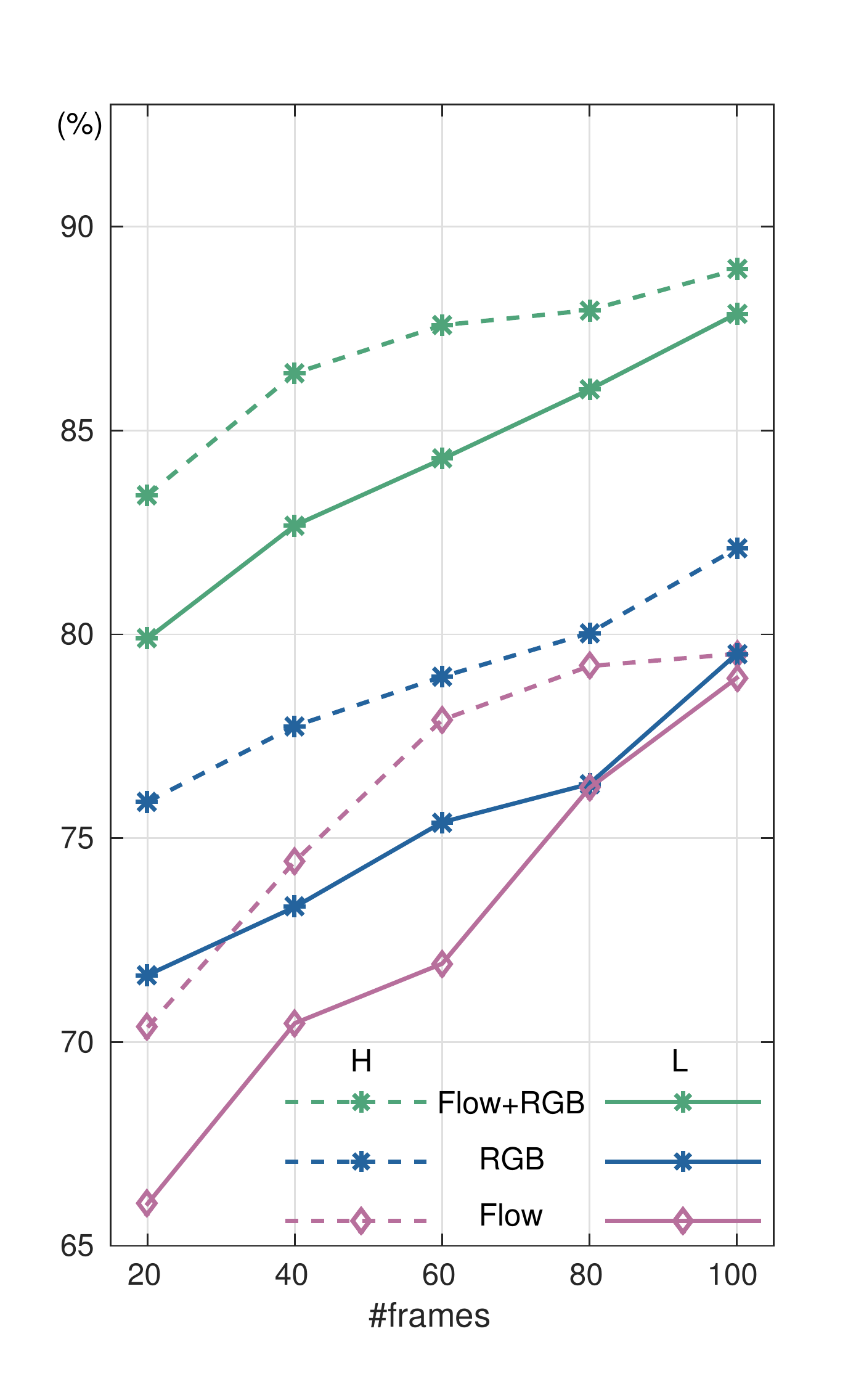}&
		\includegraphics[trim=0.5cm 0cm 1cm 1cm, clip, width=0.22\linewidth]{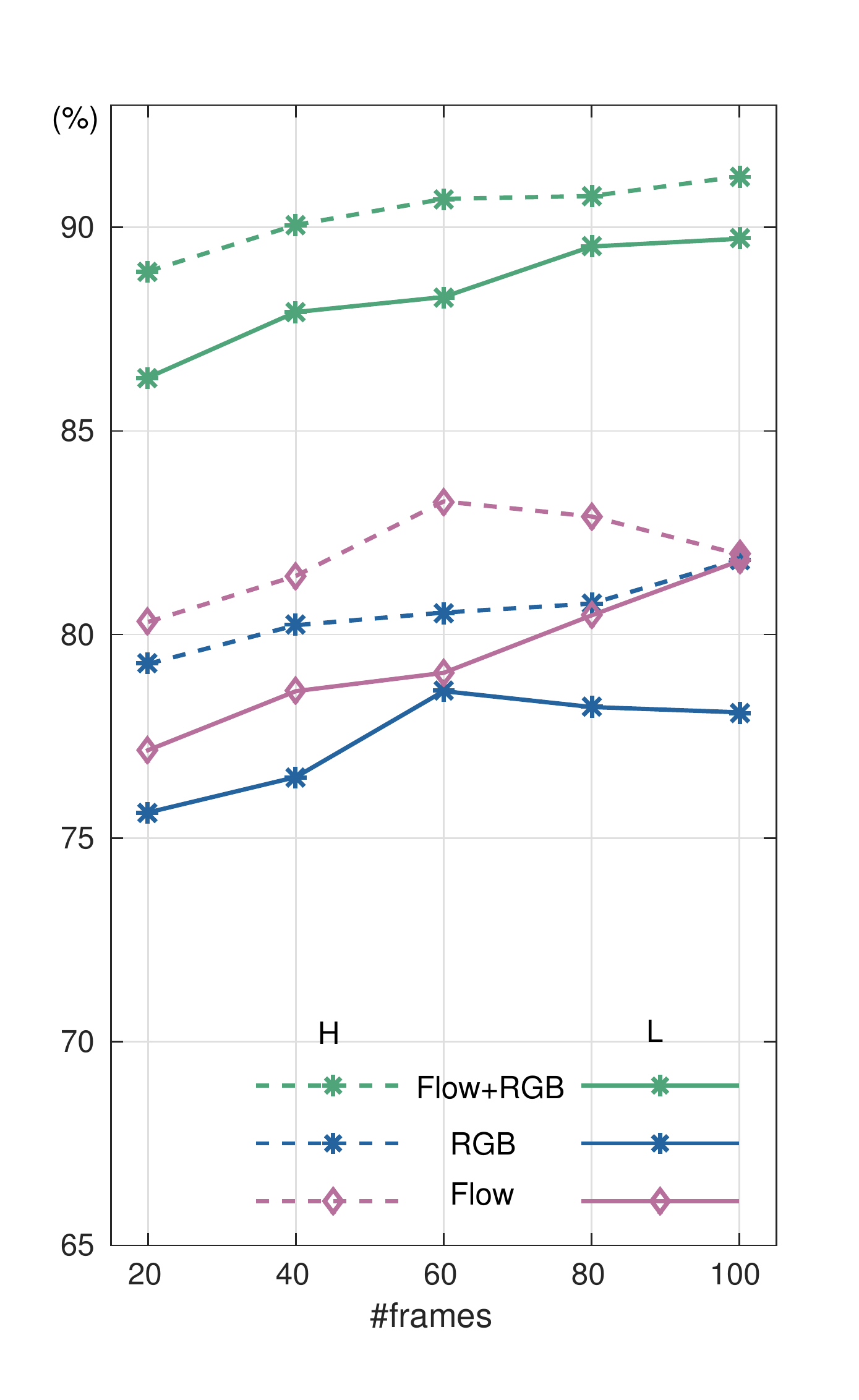}&
		\includegraphics[ trim=0.5cm 0cm 1cm 1cm, clip, width=0.22\linewidth]{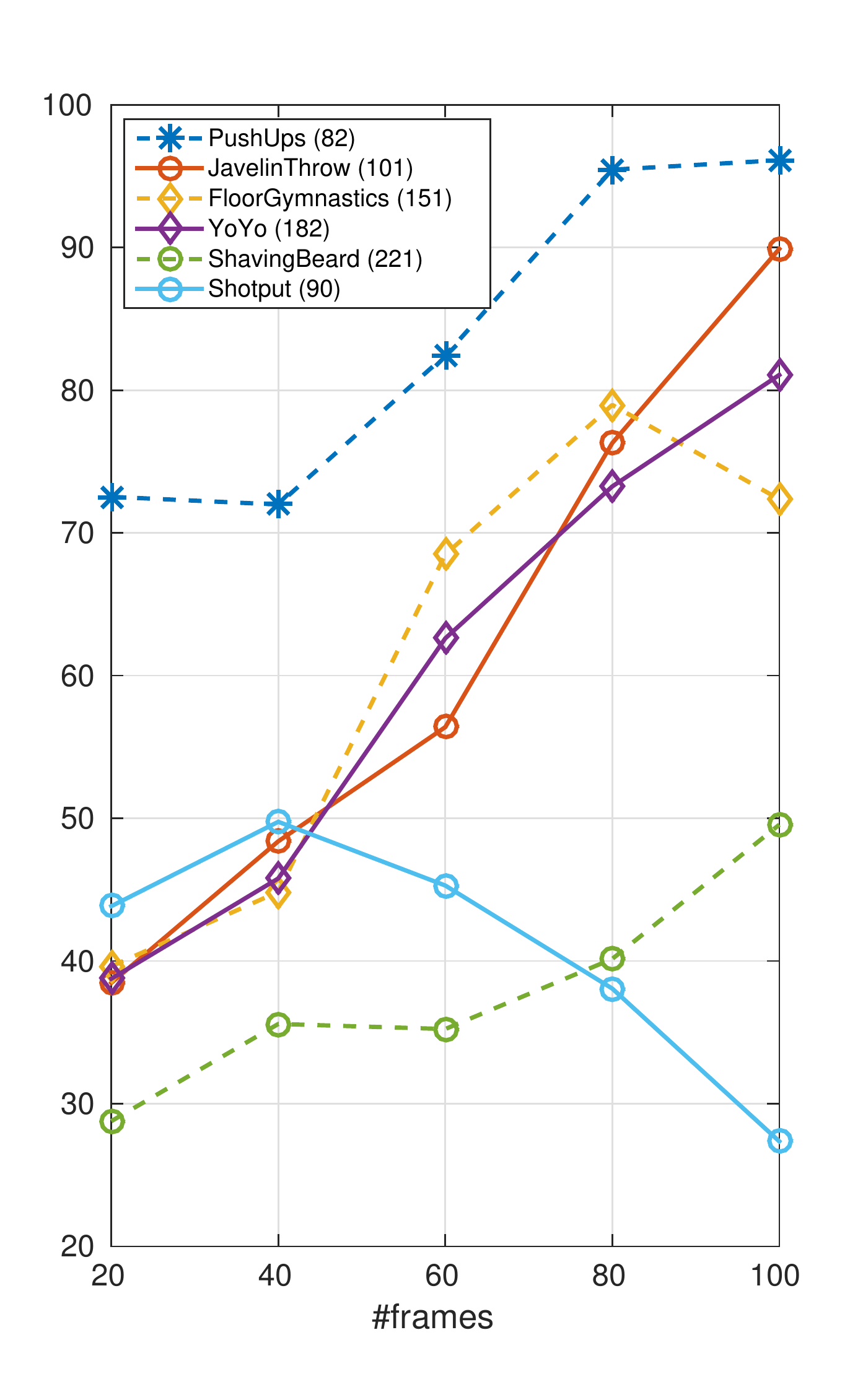}
		\end{tabular}
	\end{adjustbox}
	&\
	\begin{adjustbox}{valign=t}
		\begin{tabular}{@{}c@{}}
			{
					\em (d) optimal temporal extent
			}
			\mbox{}\vspace{-.4cm}\\ \vspace{.25cm} \\ 
			\includegraphics[width=0.24\linewidth]{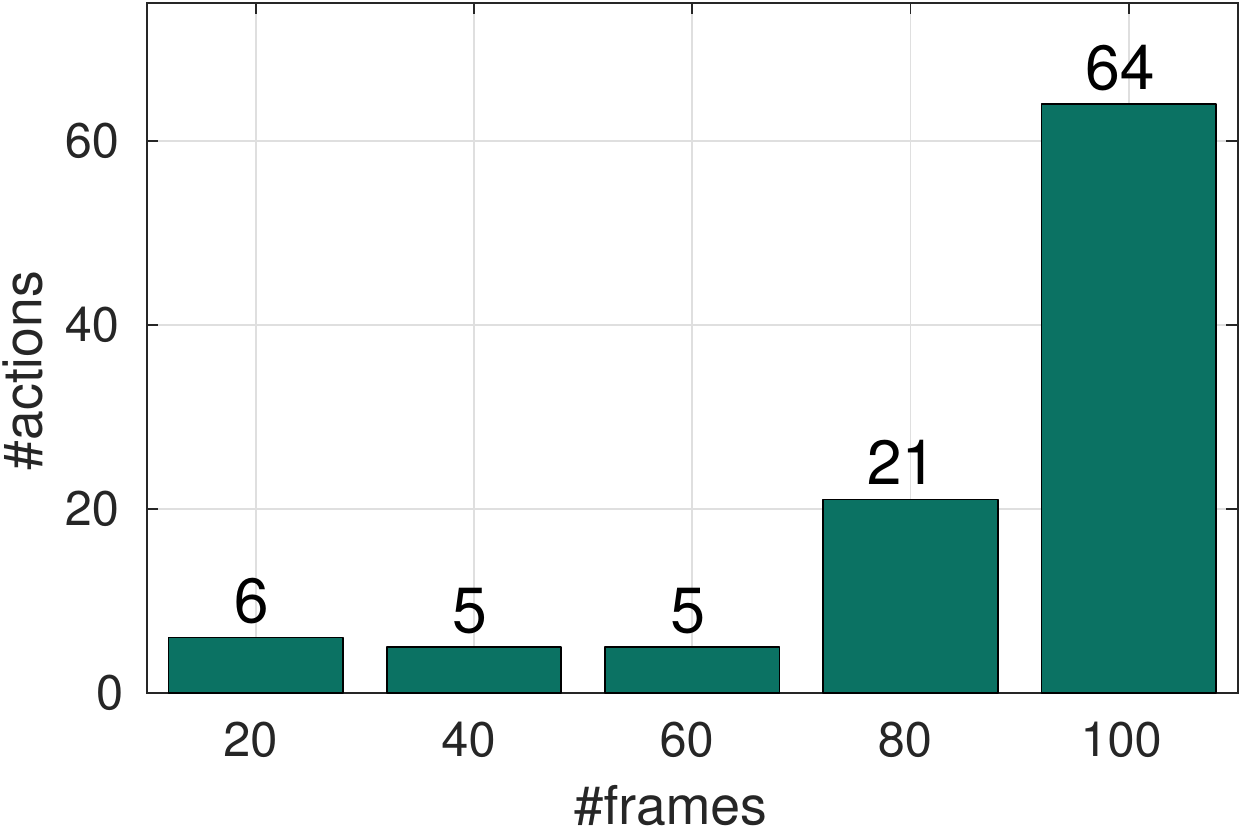}  \vspace{.1cm} \\
			{
					\em (e) average accuracy gain
			}
			\vspace{.15cm} \\
			\includegraphics[width=0.24\linewidth]{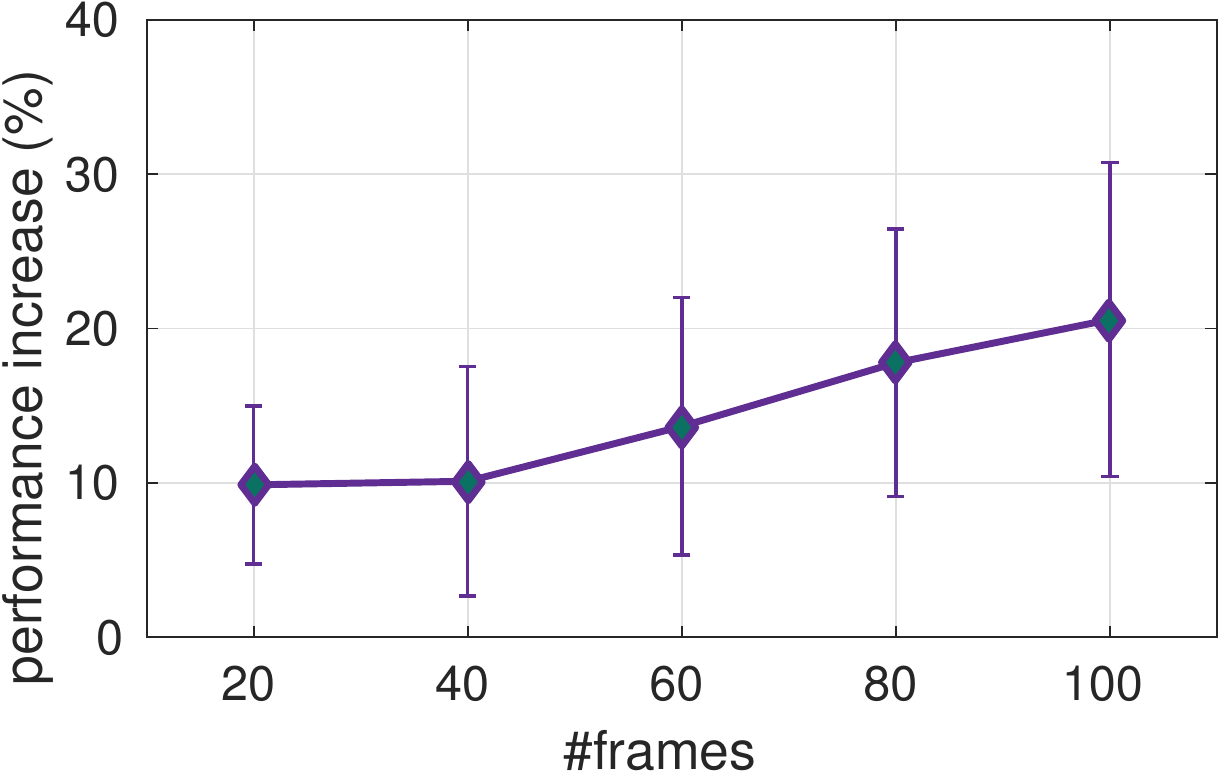}\\
		\end{tabular}
	\end{adjustbox}
	\end{tabular}
	\mbox{}\vspace{-.82cm}\\
	\caption{Results for the split 1 of UCF101 using LTC networks
		of {\em i.} varying temporal extents $t$, {\em ii.} varying spatial
		resolutions [high (H), low (L)] and {\em iii.} different input 
		modalities (RGB pre-trained on 
		Sports-1M, flow trained from scratch). For
		faster convergence all networks were trained using 0.5
		dropout and a fixed batch size of 10. Classification results are 
		shown for clips (a) and videos (b)
		computed over all classes and presented for a subset of individual
		classes for flow input of low spatial resolution (c). The average 
		number of frames in the training set for a class is denoted in parenthesis.
		(d) shows a distribution of action classes over the optimal temporal extent and
        (e) indicates correspondnig improvements (see text for details).
		With the exception of a few classes, most of the classes benefit 
		from larger temporal extents.}
	\mbox{}\vspace{-.6cm}\\
	\label{fig:nframes}
\end{figure*}

Table~\ref{table:ucf_16_60} compares the performance of 16f and 60f networks 
for RGB and flow inputs as well as for different data augmentation and dropout ratios
for UCF101 split 1.
We observe consistent and significant improvement of long-term temporal convolutions
in 60f networks for all tested setups, when measured in terms of clip and video accuracies.
Our 60f architecture significantly improves for both RGB and flow-based networks.
As expected, the improvement is more prominent for clips since video evaluation aggregates
information over the whole video.

We repeat similar experiments for the split 1 of HMDB51 and report
results in Table~\ref{table:hmdb_16_60}. Similar to UCF101, flow-based networks with
long-term temporal convolutions lead to significant improvements over the 16f 
network, in terms of clip and video accuracies.
Given the small size of HMDB51, we follow~\cite{simonyan_twostream} and also 
fine-tune networks that have been pre-trained on UCF101.
As illustrated in the 2nd row of Table~\ref{table:hmdb_16_60}, such pre-training 
gives significant improvement. Moreover, our 60f flow networks significantly 
outperform results of the 2D CNN temporal stream~(\cite{simonyan_twostream}, Table 2)
evaluated in a comparable setup, both with and without pre-training.

\paragraph*{Varying temporal and spatial resolutions}   
Given the benefits of long-term temporal convolutions above, it is 
interesting to study networks
for increasing temporal extents and varying spatial resolutions
systematically. In particular, we investigate if accuracy 
saturates for networks with larger temporal extents, if higher spatial 
resolution impacts the performance
of long-term temporal convolutions and if LTC is equally beneficial
for flow and RGB networks.
 
To study these questions, we evaluate networks with
increasing temporal extent $t\in\{20,40,60,80,100\}$ and two spatial resolutions $\{58\times58,71\times71\}$ for both RGB and flow.
We also 
investigate combining RGB and flow by averaging their class scores.
Preliminary experiments with alternative fusion
techniques did not improve over such a late fusion.

Flow networks have our previous architecture as in Figure~\ref{fig:architecture}, 
except slightly more connections in {\em fc6} for $71\times71$ resolution.
For flow input, we train our networks from scratch. For RGB input, 
learning appears to be difficult from scratch. Even if we extend the 
temporal extent from 60 frames (see Table~\ref{table:ucf_16_60}) 
to 100 frames, we obtain 68.4\% on UCF101 split~1, which is still 
below frame-based 2D convolution methods fine-tuned from ImageNet 
pre-training~\cite{simonyan_twostream}. Although longer 
extent boosts the performance significantly, we conclude 
that one needs to pre-train RGB network on larger data.

Given the large improvements provided by the pre-training of C3D 
RGB network on the large-scale Sports-1M dataset
in~\cite{tran_c3d}, we use this 16-frame pre-trained network and 
extend it to longer temporal convolutions in 2 steps.\footnote{We 
have also tried to pre-train our flow-based networks on Sports-1M 
but did not obtain significant improvements.}
The first step is fine-tuning the 16f C3D network. A randomly 
initialized fully connected ({\em fc}) layer of size 101 (number of classes) 
is added at the end of the network. Only the {\em fc} layers 
are fine-tuned by freezing the convolutional layers. We start with 
a learning rate of $3\times 10^{-4}$ and decrease it to $3\times 10^{-5}$ 
after 30K iterations for 1K more iterations. In the second step, 
we input longer clips to the network and fine-tune all the layers. 
Convolutional layers are applied to longer video clips of $t$ frames. 
This results in outputs from {\it conv5} layer with $\lfloor{t/16}\rfloor$ 
temporal resolution. To re-cycle pre-trained {\em fc} layers of 
C3D, we max-pool {\it conv5} outputs over time and pass results to 
{\it fc6}. We use a subset of the {\it fc6} weights for inputs of 
lower spatial resolution. For this phase, we run for same number of 
iterations, but we decrease the learning rate from 
$3\times 10^{-5}$ to $3\times 10^{-6}$. We keep dropout 
ratio 0.5 as in the pre-trained network.

Figure~\ref{fig:nframes}(a)(b) illustrates results of networks with 
varying temporal and spatial resolutions for clips and videos of UCF101, split~1.
We observe significant improvements over $t$ for LTC networks 
using flow (trained from scratch), RGB (with pre-training on 
Sports-1M), as well as combination of both modalities.
Networks with higher spatial resolutions give better results 
for lower $t$, however, the gain of increased
spatial resolution is lower for networks with long temporal 
extents. Given the large number of parameters in
high-resolution networks, such behavior can be explained by 
the overfitting due to the insufficient amount of training 
data in UCF101.
We believe that larger training sets could
lead to further 
improvements. Moreover, flow benefits more
from the averaging 
of clip scores than RGB. This could be an indication of static 
RGB information over different time intervals of the video, whereas flow is dynamic.

\begin{figure}
	\centering
	\includegraphics[trim = 0mm 38mm 0mm 0mm, clip, width=0.99\linewidth]{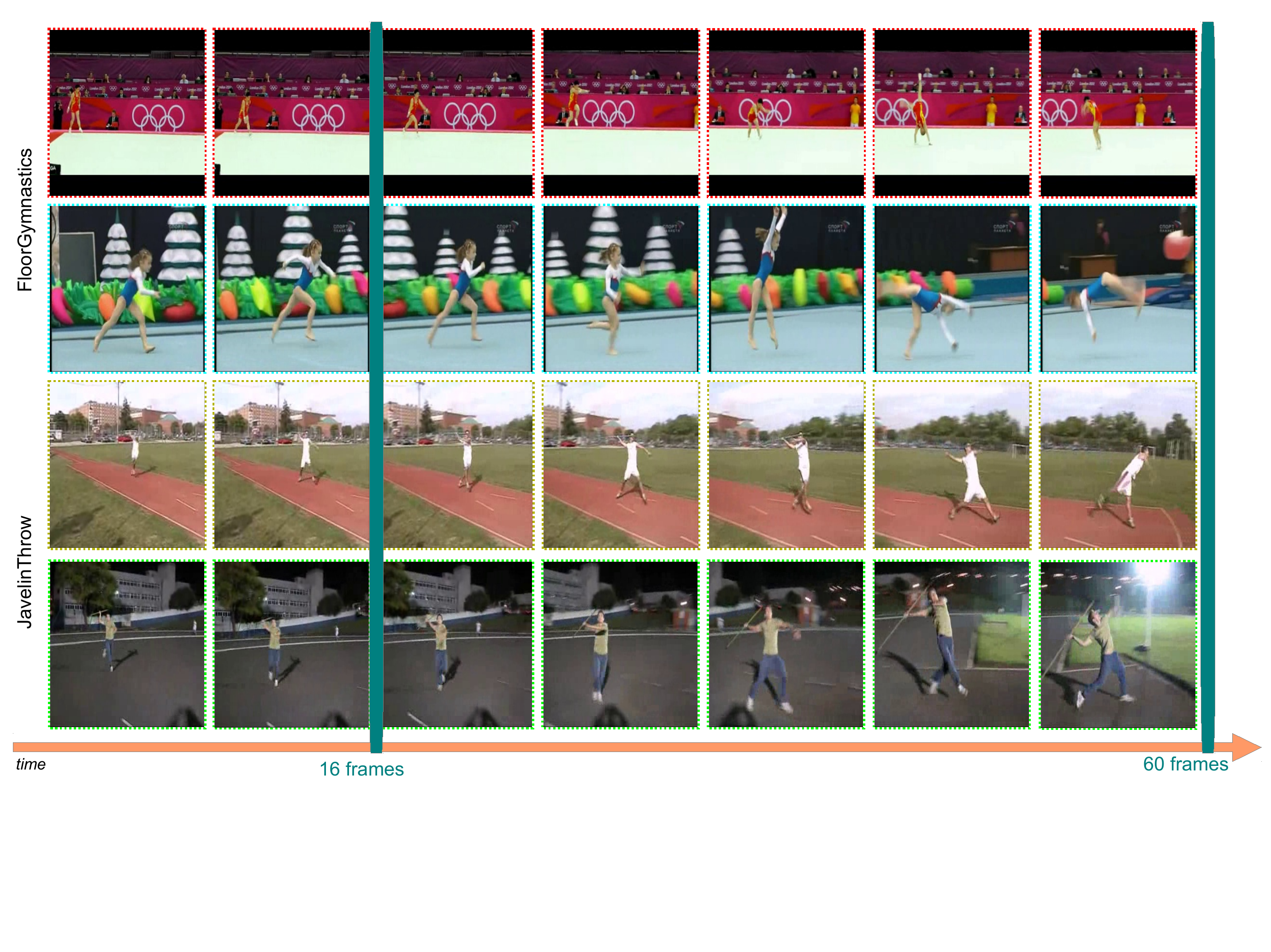}
	\mbox{}\vspace{-0.4cm}\\
	\caption{The highest improvement of long-term temporal 
		convolutions in terms of class accuracy is for {\em JavelinThrow}. 
		For 16-frame network, it is mostly confused with the 
		{\em FloorGymnastics} class. Here, we visualize sample 
		videos with 7 frames extracted at every 8 frames. The 
		intuitive explanation is that both classes start by running 
		for a few seconds and then the actual action takes place. 
		LTC can capture this interval, whereas 16-frame networks 
		fail to recognize such long-term activities.}
	\mbox{}\vspace{-0.6cm}\\
	\label{fig:frames_improvement}
\end{figure}

Figure~\ref{fig:nframes}(c) presents
results of LTC for a few action classes
demonstrating a variety of accuracy patterns over different temporal extents.
Out of all 101 classes, no action has monotonic decrease with
the increasing temporal extent, whereas the performance
of 25 action classes increased monotonically.
{\em{PushUps}}, {\em{YoYo}} and {\em{ShavingBeard}} are
examples of classes with high, medium and low performance
that all benefit from larger temporal extents.
{\em{Shotput}} is an example of a class with lower performance for 
longer temporal extents.
A possible explanation is that samples of the {\em{Shotput}} class are
relatively short and have 90 frames on average (we pad short clips).
Two additional examples with a significant 
gain for larger temporal extents are {\em{FloorGymnastics}} and
{\em{JavelinThrow}}, see Figure~\ref{fig:frames_improvement} for sample frames
from these two classes.
We observe that both actions are composed of running followed by 
throwing a javelin or the actual gymnastics action. Short-term
networks, thus, easily confuse the two actions, while 
LTC can capture such long and complex actions.
For both classes, we provide snapshots at
every 8th frame. It is clear that one needs to look at more than 16
frames to distinguish these actions.

Let the performance of class $c$ for temporal extent $t$ be $p_c(t)$.
A set of classes with the maximum performance at $t$ is then 
$M(t) := \{c \mid t \in \argmax\limits_{t'}(p_c(t'))\}$.
Figure~\ref{fig:nframes}(d) plots $|M(t)|$ with respect to $t$. 
The majority of classes (64 out of 101)
obtain maximum performance when trained with 100f networks.
To further check if there exists an ``ideal temporal extent''
for different actions, Figure~\ref{fig:nframes}(e) illustrates the
average performance increase $d(t)$:
\vspace{-.2cm}
\begin{equation}
d(t) := \frac{1}{|M(t)|}\sum_{M(t)}{\max_{t'}(p_c(t')) -
	 \min_{t'}(p_c(t'))}
	\vspace{-.2cm}
\end{equation}
We can observe that values of $d(t)$ are lower for shorter extents
  and larger for longer extents. 
That means actions scoring best at short extents
 score similar at all scales, so we cannot conclude that certain
actions favor certain extents. Most actions favor long extents
 as the difference is largest for 100f.
A possible explanation is that making the interval too long for
short actions does not have much impact, whereas making the interval
too short for long actions does impact the performance, see
Figure~\ref{fig:frames_improvement}.

\paragraph*{Combining networks of varying temporal resolutions}  
We evaluate combining different
networks with late fusion.
For final results on flow, $58\times 58$ spatial resolution and 0.9 
dropout are used for both UCF101 and HMDB51 datasets. The flow 
networks are learned from scratch for UCF101 and fine-tuned for HMDB51.
For final results on UCF101 with RGB input, we use $71 \times 71$ spatial resolution networks fine-tuned from 
C3D network~\cite{tran_c3d}. However, we do not further fine-tune it for HMDB51 because of overfitting, 
and use C3D network as a feature extractor in combination with SVM for obtaining RGB scores. 
Our implementation of C3D as a feature extractor and a SVM classifier achieved 80.2\% and 49.7\% 
average performance on 3 splits of UCF101 and HMDB51, respectively. We get similar result when fine-tuning 
C3D on 16-frames (80.5\% on UCF101).

\begin{figure}
	\parbox[]{0.4\linewidth}{\null
		\centering
		\includegraphics[trim = 1.2cm 1cm 1.5cm 0.8cm, clip, clip, width=0.99\linewidth]{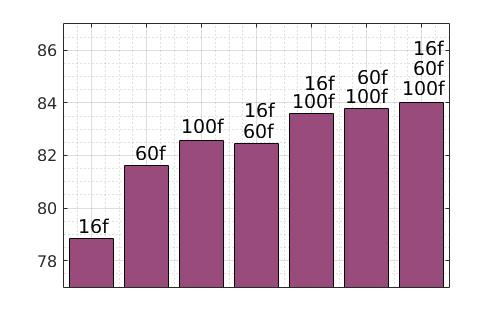}
	} \quad
	\parbox[]{0.43\linewidth}{\null
		\begin{adjustbox}{max width=1.27\linewidth}
			\begin{tabular}{lcc}
				\toprule
				&  UCF101   & HMDB51 \\ \midrule
				LTC$_{Flow}$ (100f)		& 82.6      & 56.7   \\		
				LTC$_{Flow}$ (60f+100f)	& 83.8 		& 60.5   \\
				LTC$_{RGB}$  (100f)		& 81.8		& -	 	 \\
				LTC$_{RGB}$  (60f+100f)	& 81.5 		& -		 \\
				LTC$_{Flow+RGB}$  		& 91.0		& 65.6	 \\ 
				LTC$_{Flow+RGB}$+IDT 		& \textbf{91.8}		& \textbf{67.7} 	 \\
				\bottomrule
			\end{tabular}
		\end{adjustbox}
	}
	\mbox{}\vspace{-0.4cm}\\
	\caption{Results for network combinations.
		(Left): Combination of LTC flow networks with different 
		temporal extents on UCF101-split~1.
		(Right): Combination of flow and RGB networks together 
		with IDT features on UCF101 and 
		HMDB51-splits~1. For UCF101, flow is trained from scratch 
		and RGB is pre-trained on Sports-1M. For HMDB51, flow is 
		pre-trained on UCF101 and RGB scores are obtained using C3D feature extractor.}
	\mbox{}\vspace{-1cm}\\
	\label{fig:actionlength}
\end{figure}

Figure~\ref{fig:actionlength} (left) shows results for
combining outputs of flow networks with different temporal extents. The
combination is performed by averaging video-level class scores
produced by each network. We observe that combinations of two networks
with different temporal extents provides significant improvement for flow. The
gains of combining more than two resolutions appear to be
marginal. For final results, we report combining 60f and 100f 
networks for both flow and RGB, with the exception of HMDB51 RGB scores 
for which we use a SVM classifier on 16f feature extractor.
Figure~\ref{fig:actionlength} (right) shows results for
combining multiscale networks of different modalities together with
the IDT+FV baseline classifier~\cite{wang_idt} on split~1 of both datasets. We observe
complementarity of different networks and IDT+FV where the best result
is obtained by combining all classifiers.

\vspace{-.5cm}
\subsection{Comparison with the-state-of-the-art}  \label{ss:comp_soa}
In Table~\ref{table:stateoftheart}, we compare to the state-of-the-art
on HMDB51 and UCF101 datasets. Note that the numbers do not 
directly match with previous tables and figures, 
which are reported only on first splits. Different methods are grouped together
according to being hand-crafted, using only RGB or
optical flow input to CNNs and combining any of these.
Trajectory features perform already well, especially with
higher-order encodings. CNNs on RGB perform very poor if trained from
scratch, but strongly benefits from static image pre-training such as 
ImageNet. Recently~\cite{tran_c3d} trained space-time filters from a
large collection of videos; however, their method is not end-to-end,
given that one has to train a SVM on top of the CNN features. 
Although we fine-tune LTC$_{RGB}$ based on a network learned with a short temporal span 
and we reduce the spatial resolution, we are able to improve 
by 2.2\% on UCF101 (80.2\% versus 82.4\%) by extending the pre-trained network to 100 frames.

We observe that LTC outperforms 2D convolutions on both datasets.
Moreover, LTC$_{Flow}$ outperforms LTC$_{RGB}$ despite no pre-training. 
Our results using LTC$_{Flow+RGB}$ with average fusion significantly 
outperform the Two-stream average fusion baseline~\cite{simonyan_twostream} 
by 4.8\% and 6.8\% on UCF101 and HMDB51 datasets, respectively. Moreover, 
the SVM fusion baseline in~\cite{simonyan_twostream} is still 
significantly below LTC$_{Flow+RGB}$.
Overall, the combination of our best networks LTC$_{Flow+RGB}$ together 
with the IDT method\footnote{Our implementation of IDT+FV~\cite{wang_idt} 
obtained 84.5\% and 57.3\% for UCF101 and HMDB51, respectively.} provides 
best results on both UCF101~(92.7\%) and HMDB51~(67.2\%) datasets.
Notably both of these results outperform previously published results on these datasets, except~\cite{feichtenhofer_twostreamfusion} which studies best ways to 
combine RGB and flow streams, hence complementary to our method.

\begin{table}
	\begin{center}
		\begin{adjustbox}{max width=0.426\textwidth}
			\begin{tabular}{cclcc}
				\toprule
				& 			 				& Method 	& UCF101 & HMDB51 	\\
				\midrule
				\multirow{2}{*}{\scriptsize IDT}& \cite{wang_idt}				& IDT+FV 					& 85.9 	& 57.2 		\\
			& \cite{MIFS}					& IDT+MIFS					& 89.1 	& 65.1 		\\
				\midrule
				\multirow{6}{*}{\scriptsize RGB}& \cite{karpathy_sports1m} 	& Slow fusion (from scratch)& 41.3 	& - 		\\
				& \cite{tran_c3d} 			& C3D (from scratch) 		& 44\textsuperscript{1} & -  \\
				& \cite{karpathy_sports1m} 	& Slow fusion 				& 65.4  & -  		\\
			    & \cite{simonyan_twostream} 	& Spatial stream 			& 73.0 	& 40.5 		\\ 
			   	& \cite{tran_c3d} 			& C3D (1 net) 				& 82.3	& -			\\
				& \cite{tran_c3d} 			& C3D (3 nets) 				& 85.2	& -			\\
				\midrule
								{\scriptsize Flow} 	  & \cite{simonyan_twostream} 	& Temporal stream				 & 83.7 & 54.6 \\
				\midrule
				\multirow{4}{*}{\scriptsize RGB}
				& \cite{simonyan_twostream}	& Two-stream (avg. fusion) 	& 86.9 	& 58.0 		\\ 
				\multirow{4}{*}{\scriptsize +}& \cite{simonyan_twostream}	& Two-stream (SVM fusion)  	& 88.0 	& 59.4 		\\ 
				\multirow{4}{*}{\scriptsize Flow}& \cite{yue_beyondshort} 		& LSTM						& 88.6 	& -			\\
				& \cite{wang_tdd}		 		& TDD 						& 90.3 	& 63.2		\\
				& \cite{wang_transformations} & Transformations			& 92.4	& 62.0 		\\
				& \cite{feichtenhofer_twostreamfusion} & Two-stream (conv. fusion) & \textbf{92.5} & \textbf{65.4} \\
			    \midrule 
				\multirow{3}{*}{\scriptsize +IDT}& \cite{tran_c3d}				& C3D+IDT 					& 90.4  & -			\\
				& \cite{wang_tdd} 			& TDD+IDT 					& 91.5 	& 65.9		\\
				& \cite{feichtenhofer_twostreamfusion} & Two-str. (conv. fusion)+IDT & \textbf{93.5} & \textbf{69.2} \\
				\midrule
				& 							& LTC$_{RGB}$		& 82.4 & - \textsuperscript{2} \\
				&								& LTC$_{Flow}$		& 85.2 & 59.0 \\
				& 							& LTC$_{Flow+RGB}$		& \textbf{91.7} & \textbf{64.8} \\ 
				&								& LTC$_{Flow+RGB}$+IDT & \textbf{92.7} & \textbf{67.2}	\\
				\bottomrule
			\end{tabular}
		\end{adjustbox}
	\end{center}
	\caption{Comparison with the state-of-the-art on UCF101 and HMDB51 (mean accuracy across 3 splits).
		\textsuperscript{1}This number is read from the plot in 
		figure~2~\cite{tran_c3d} and is clip-based, therefore not directly comparable. 
		\textsuperscript{2}We use C3D+SVM scores (49.7\%) for HMDB51.}
	\mbox{}\vspace{-1.65cm}\\
	\label{table:stateoftheart}
\end{table}

\vspace{-0.2cm}
\subsection{Analysis of the 3D spatio-temporal filters}
\label{ss:visualization}

\begin{figure}
\begin{center}
\frame{\includegraphics[width=\fs\columnwidth, height=\fs\columnwidth]{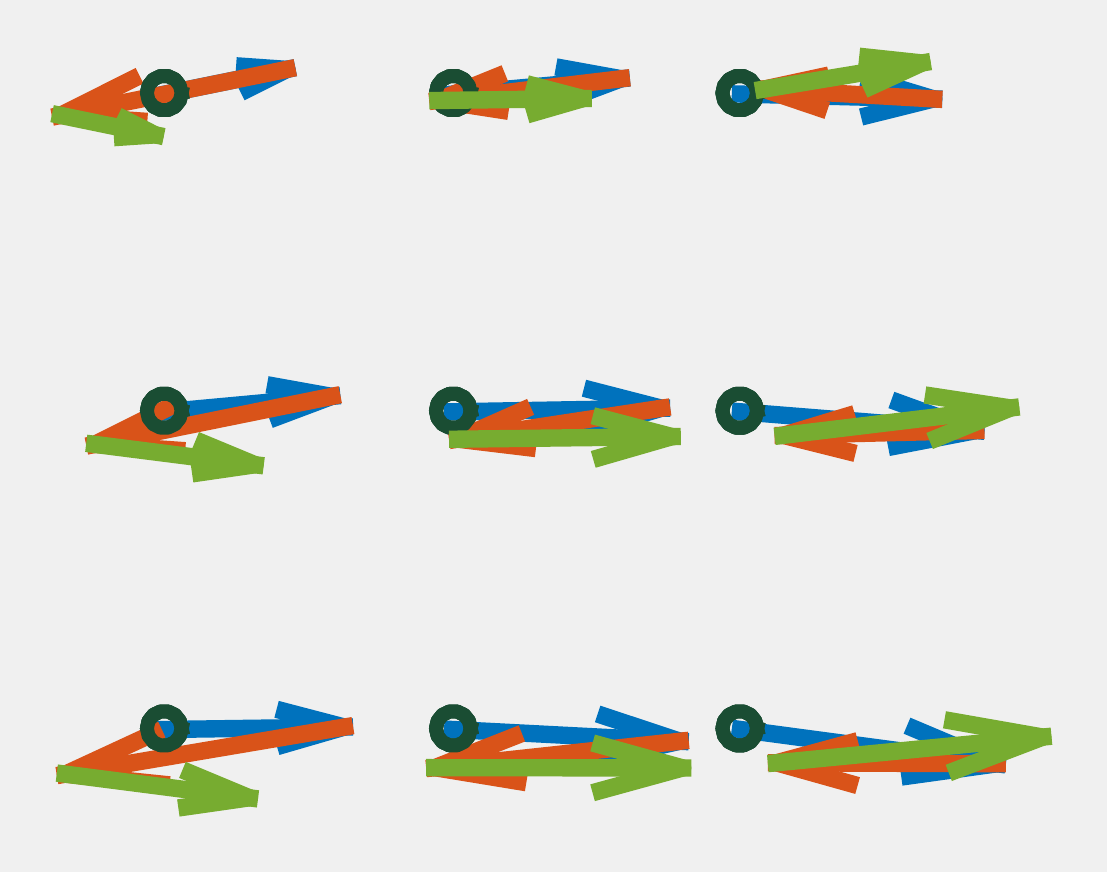}}
\frame{\includegraphics[width=\fs\columnwidth, height=\fs\columnwidth]{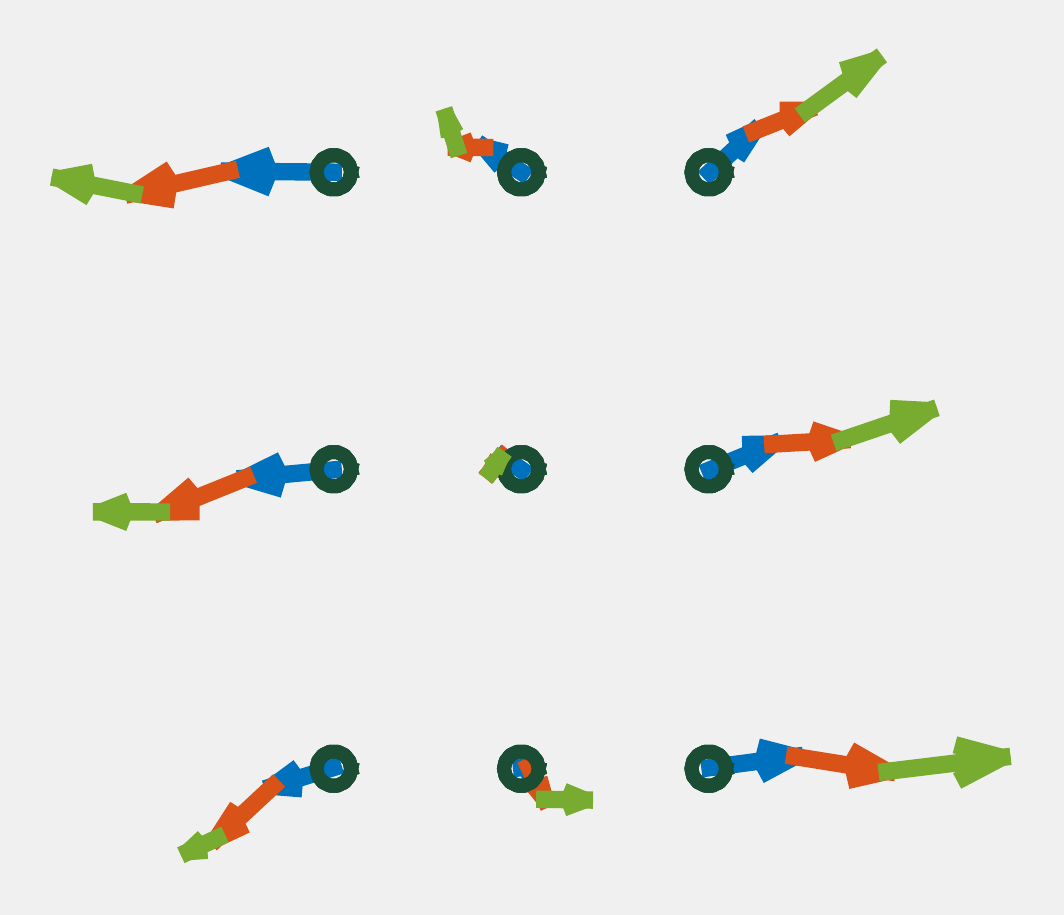}}
\frame{\includegraphics[width=\fs\columnwidth, height=\fs\columnwidth]{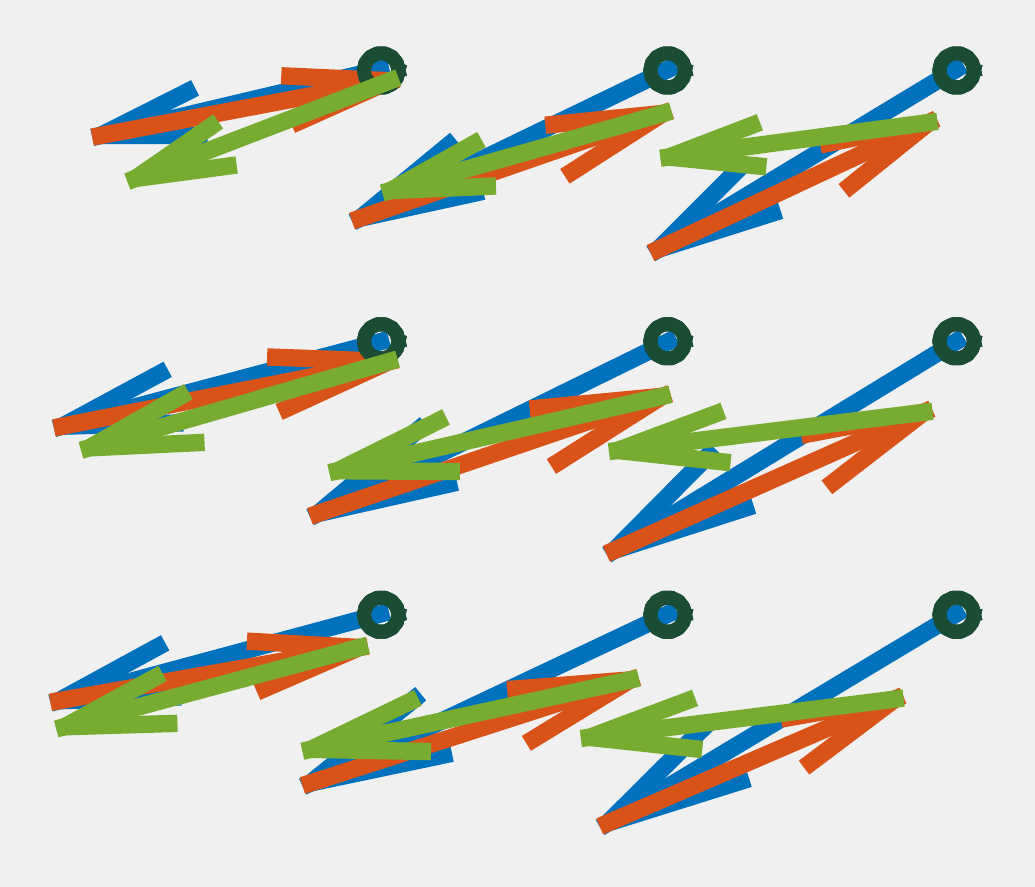}}
\frame{\includegraphics[width=\fs\columnwidth, height=\fs\columnwidth]{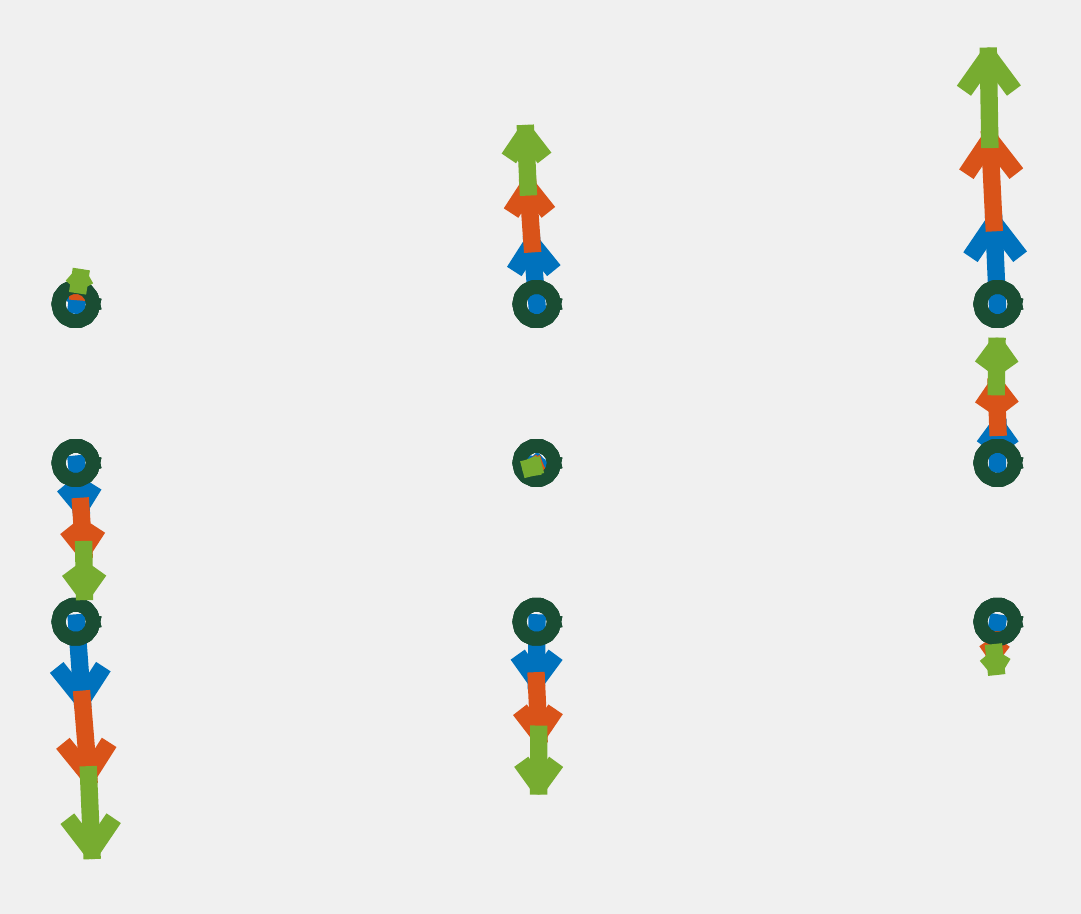}}
\frame{\includegraphics[width=\fs\columnwidth, height=\fs\columnwidth]{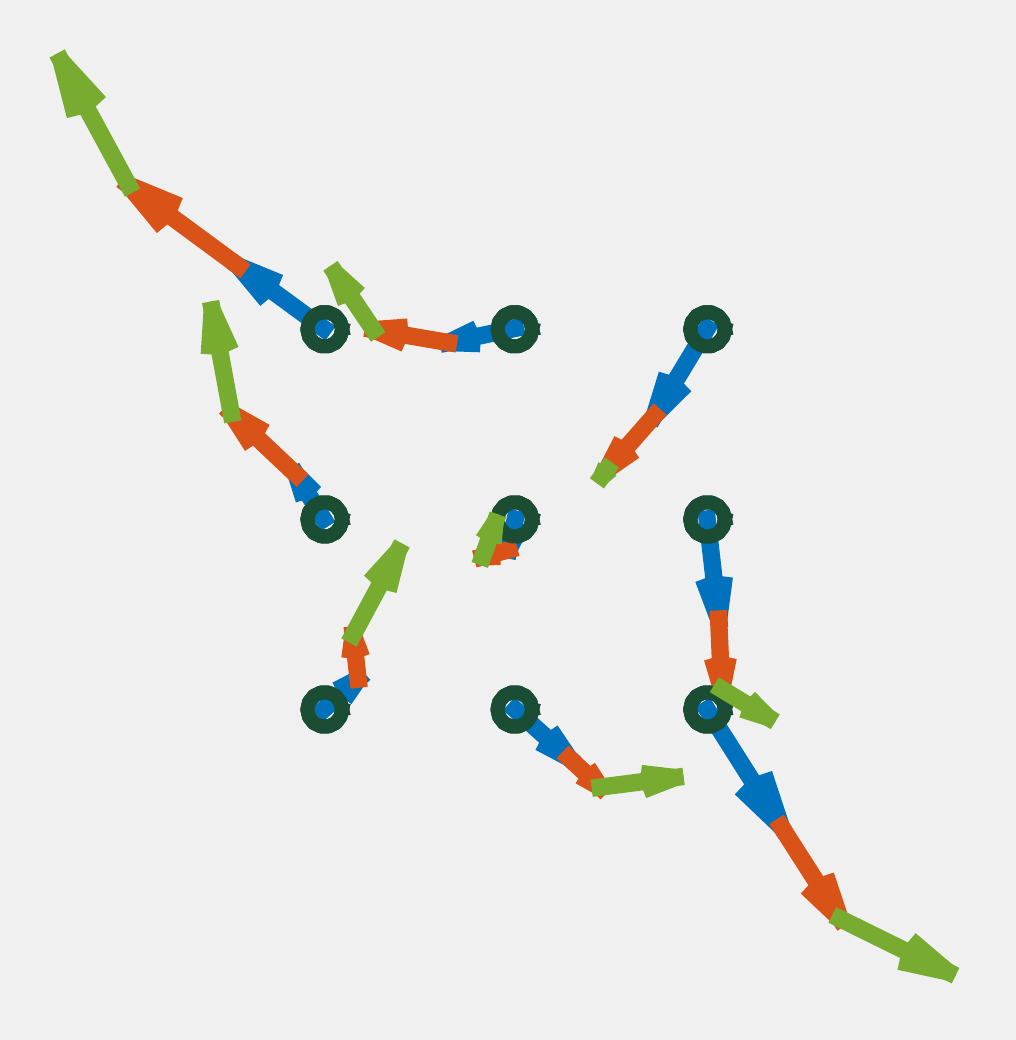}}
\frame{\includegraphics[width=\fs\columnwidth, height=\fs\columnwidth]{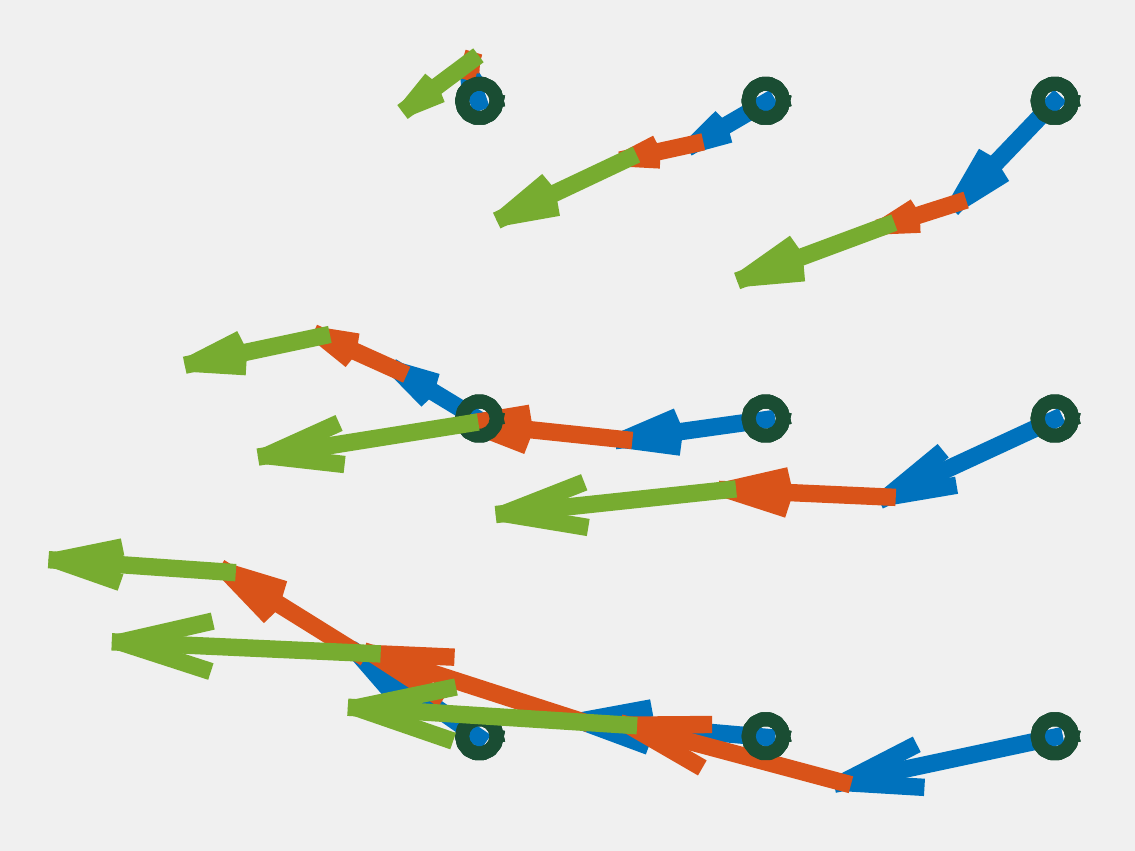}}\\
\frame{\includegraphics[width=\fs\columnwidth, height=\fs\columnwidth]{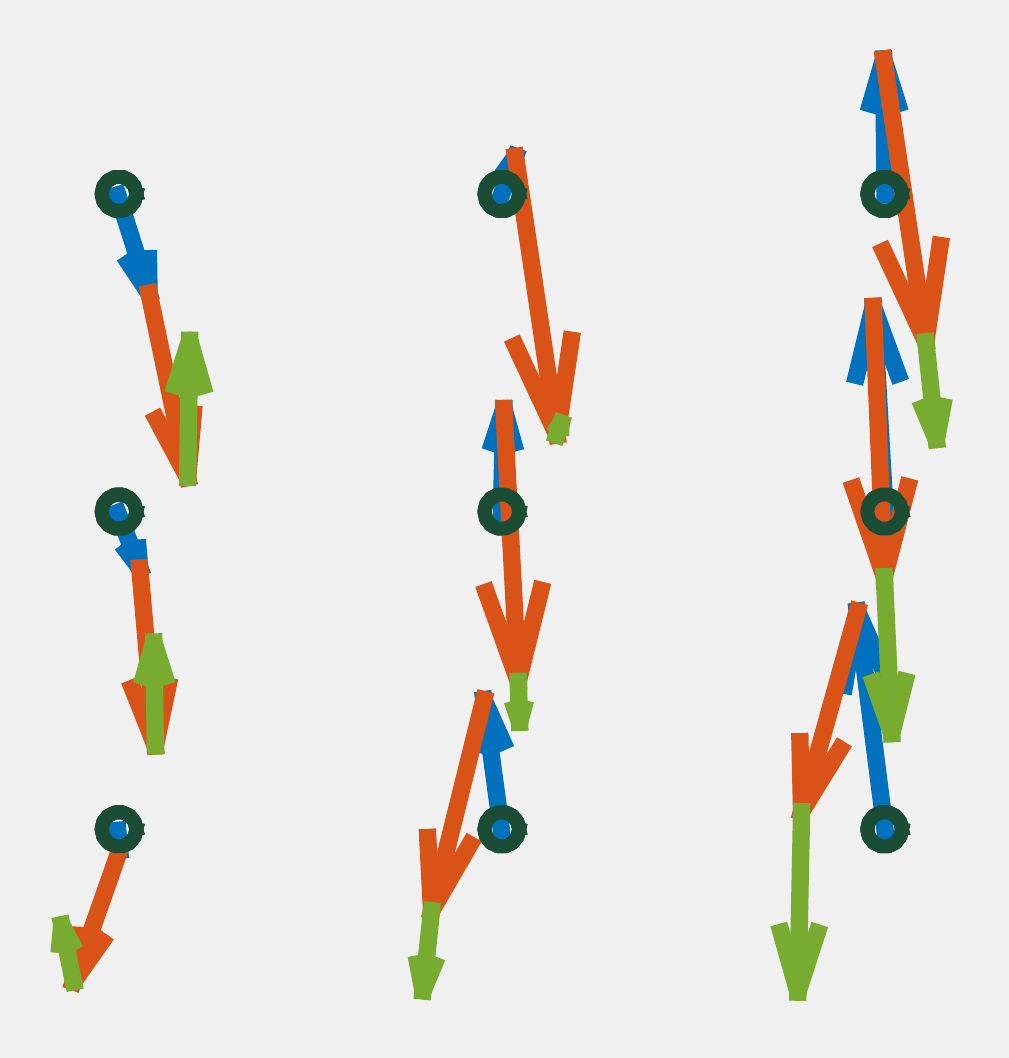}}
\frame{\includegraphics[width=\fs\columnwidth, height=\fs\columnwidth]{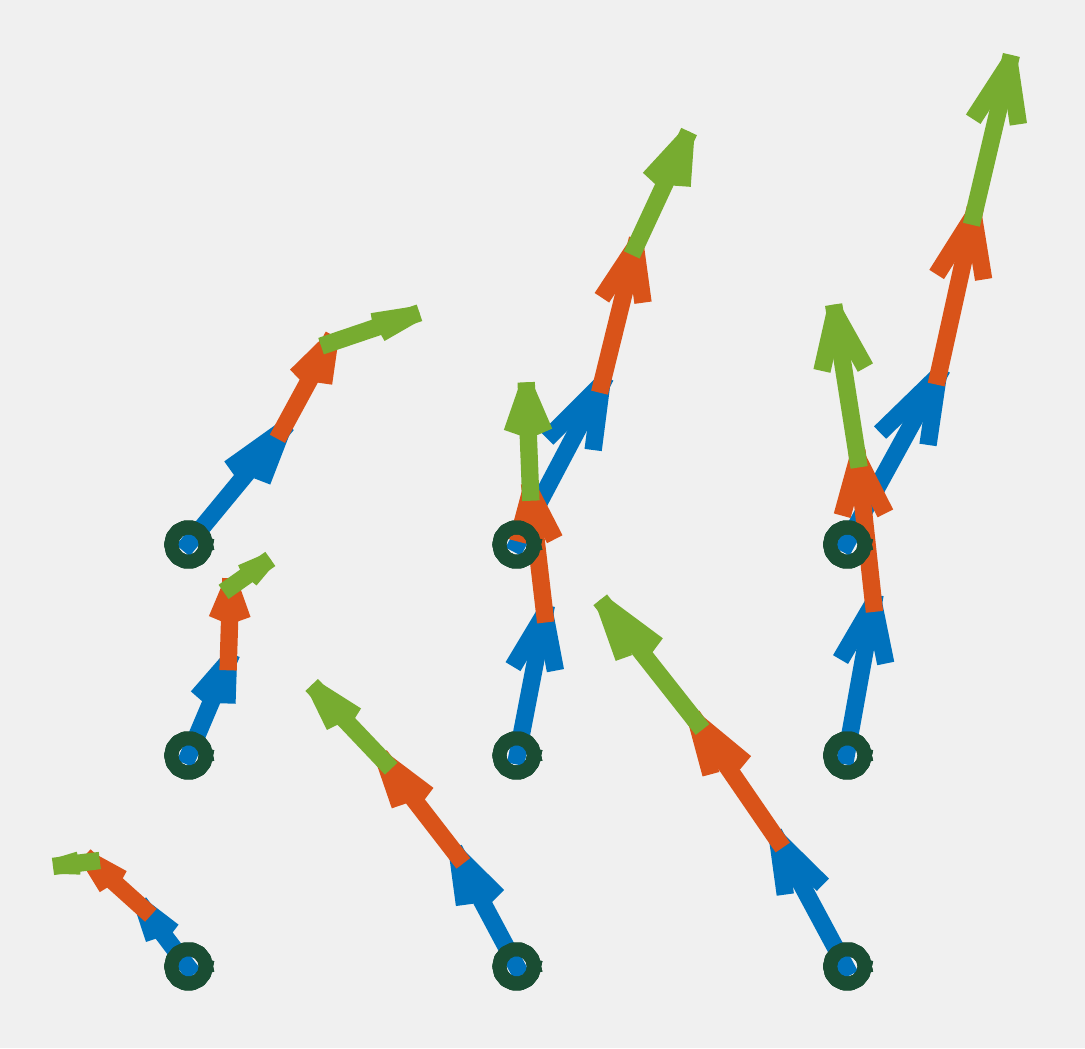}}
\frame{\includegraphics[width=\fs\columnwidth, height=\fs\columnwidth]{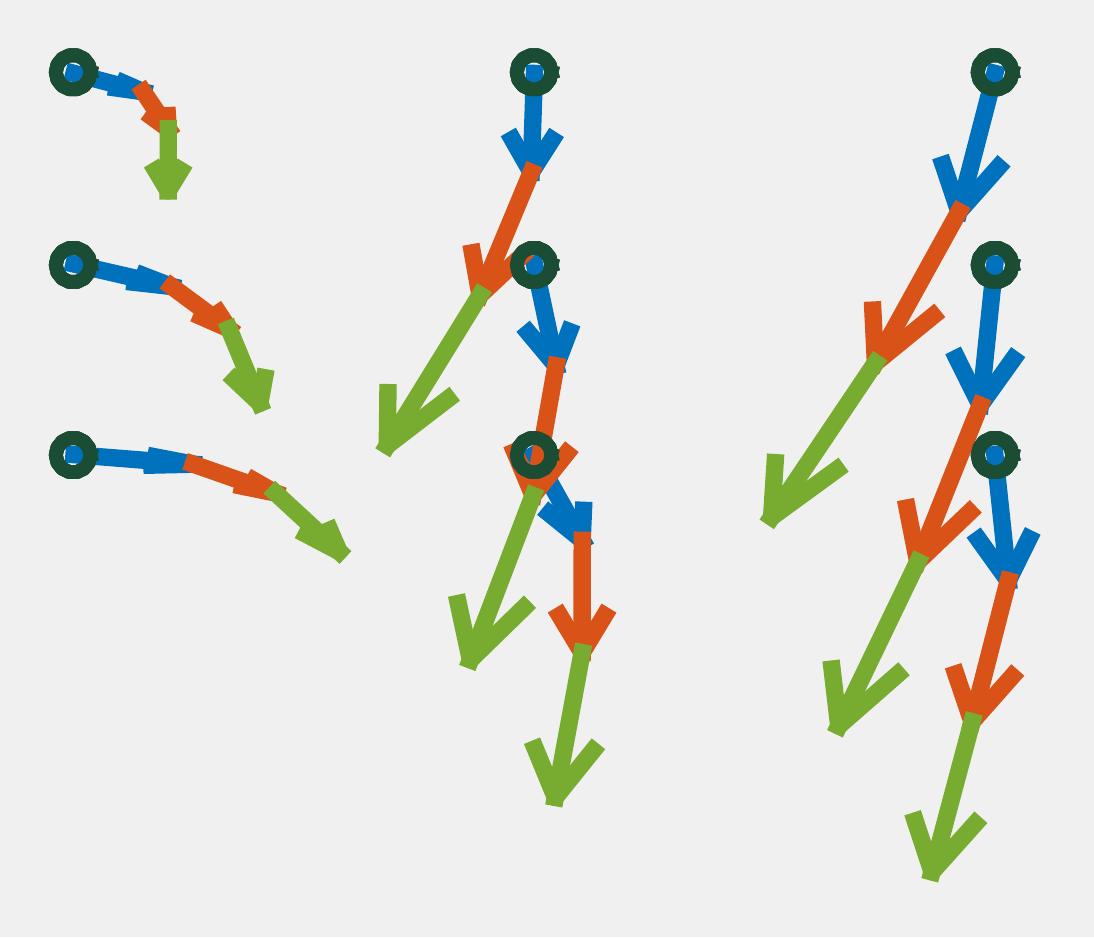}}
\frame{\includegraphics[width=\fs\columnwidth, height=\fs\columnwidth]{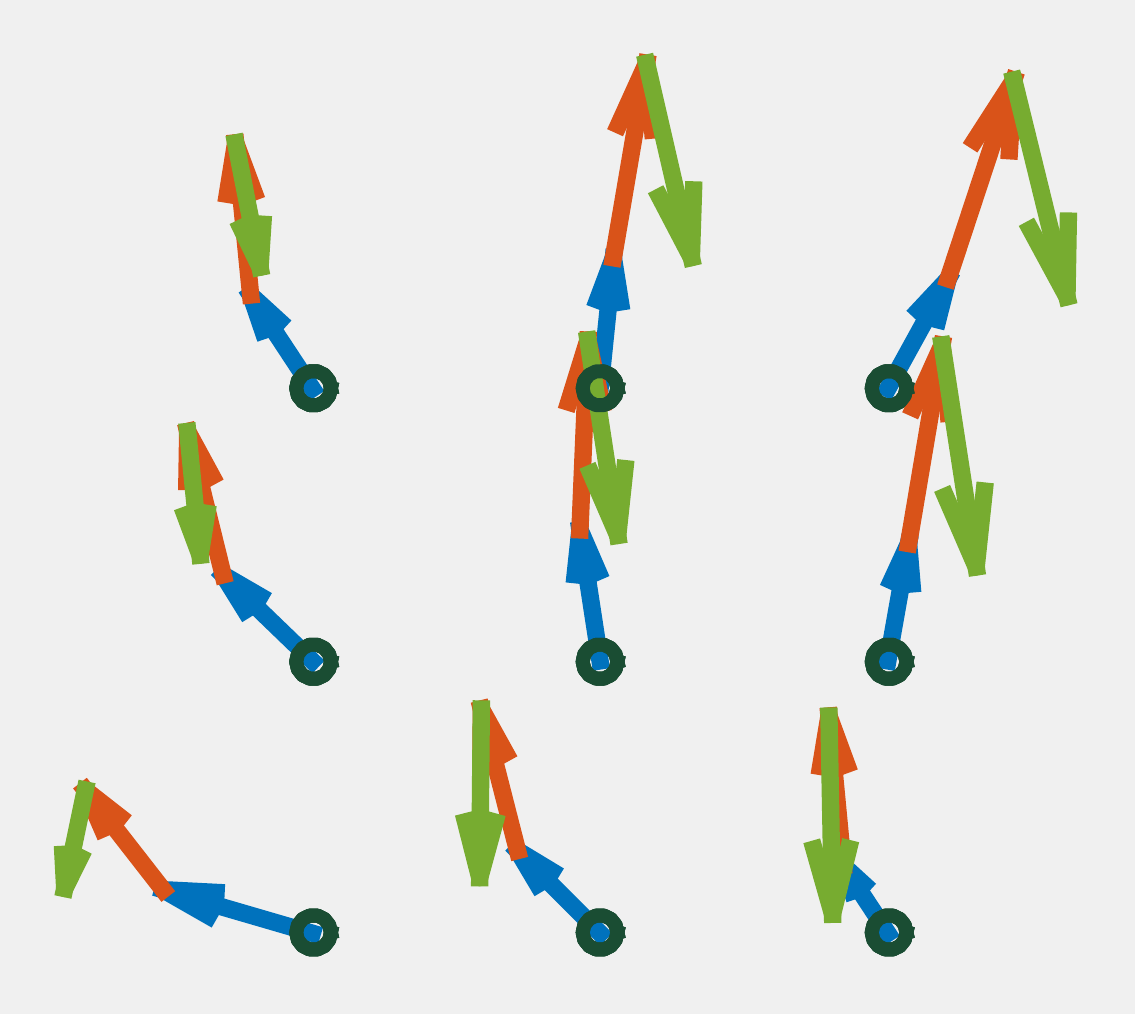}}
\frame{\includegraphics[width=\fs\columnwidth, height=\fs\columnwidth]{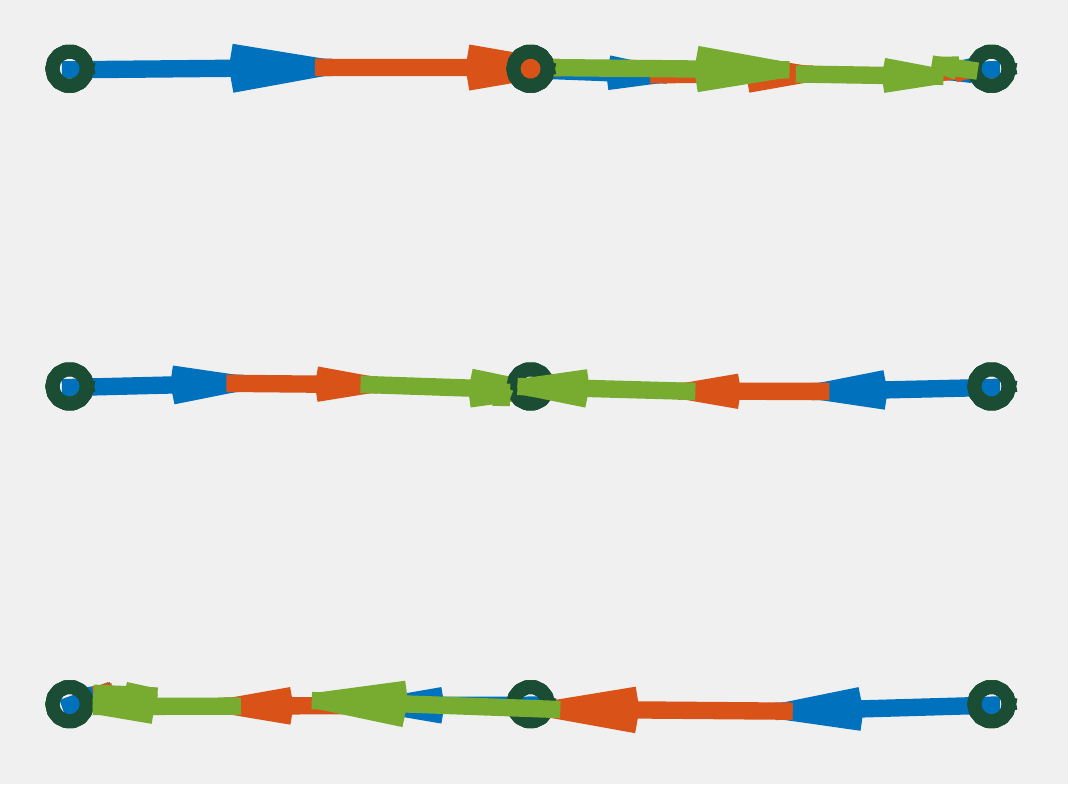}}
\frame{\includegraphics[width=\fs\columnwidth, height=\fs\columnwidth]{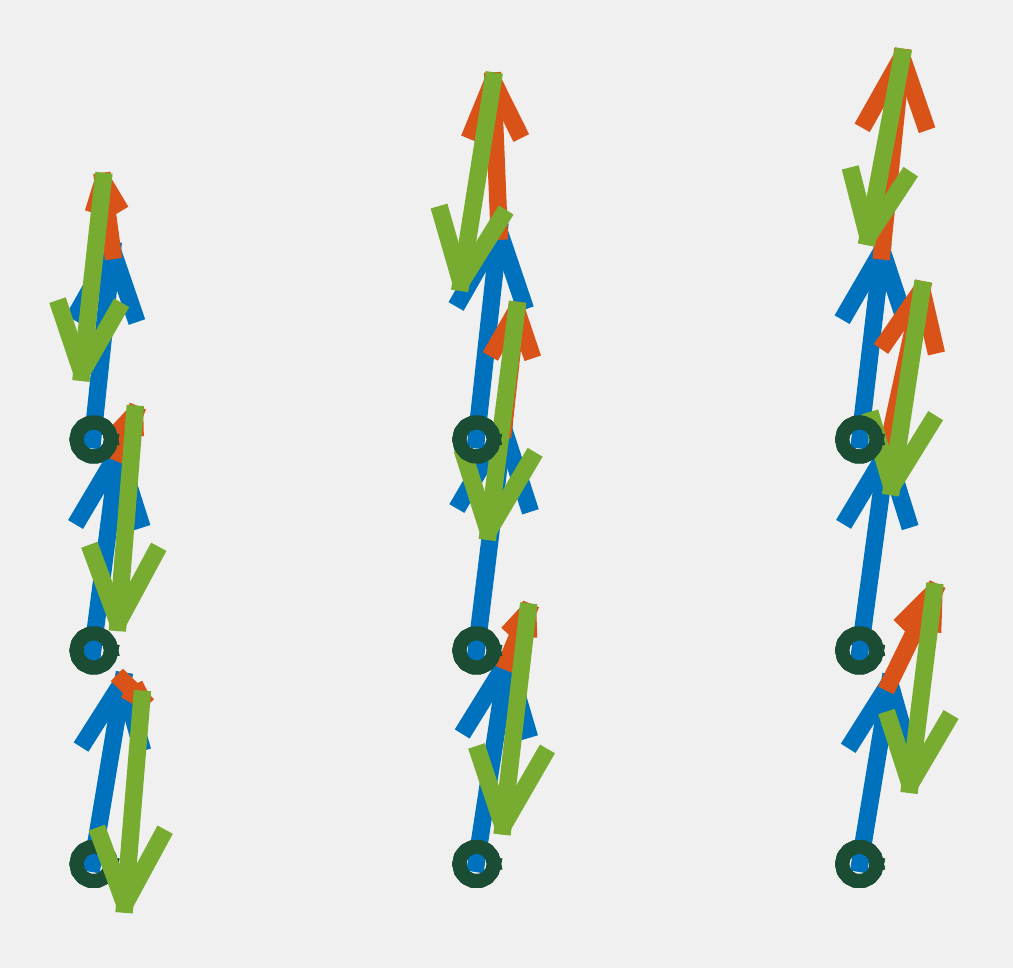}}\\
\frame{\includegraphics[width=\fs\columnwidth, height=\fs\columnwidth]{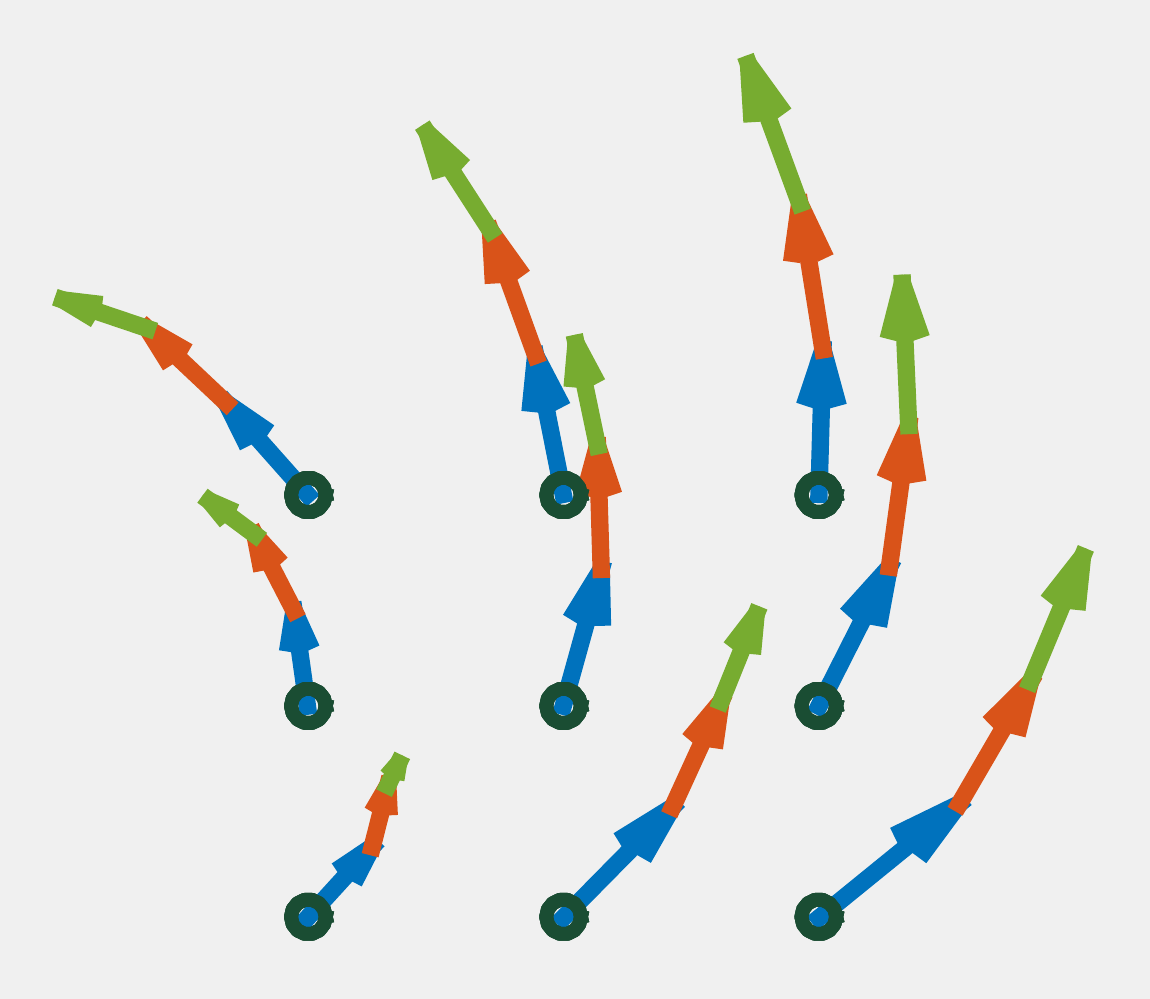}}
\frame{\includegraphics[width=\fs\columnwidth, height=\fs\columnwidth]{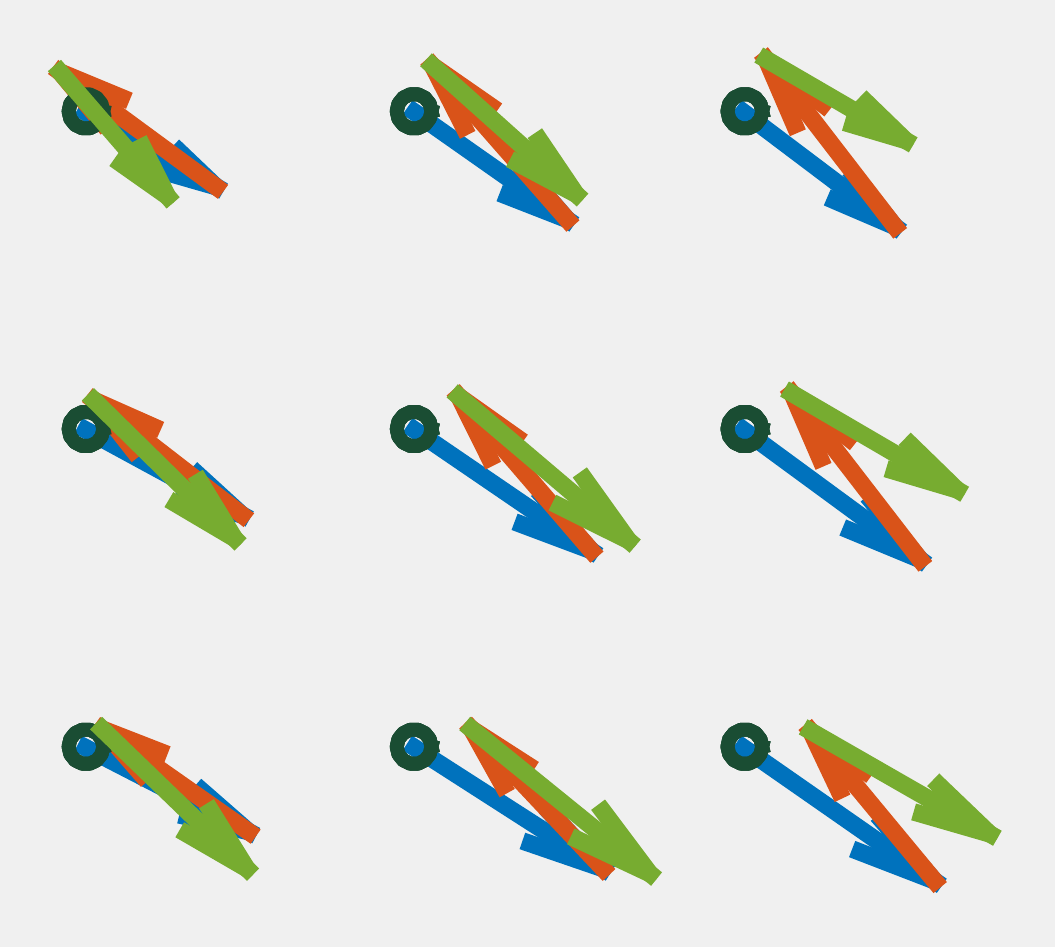}}
\frame{\includegraphics[width=\fs\columnwidth, height=\fs\columnwidth]{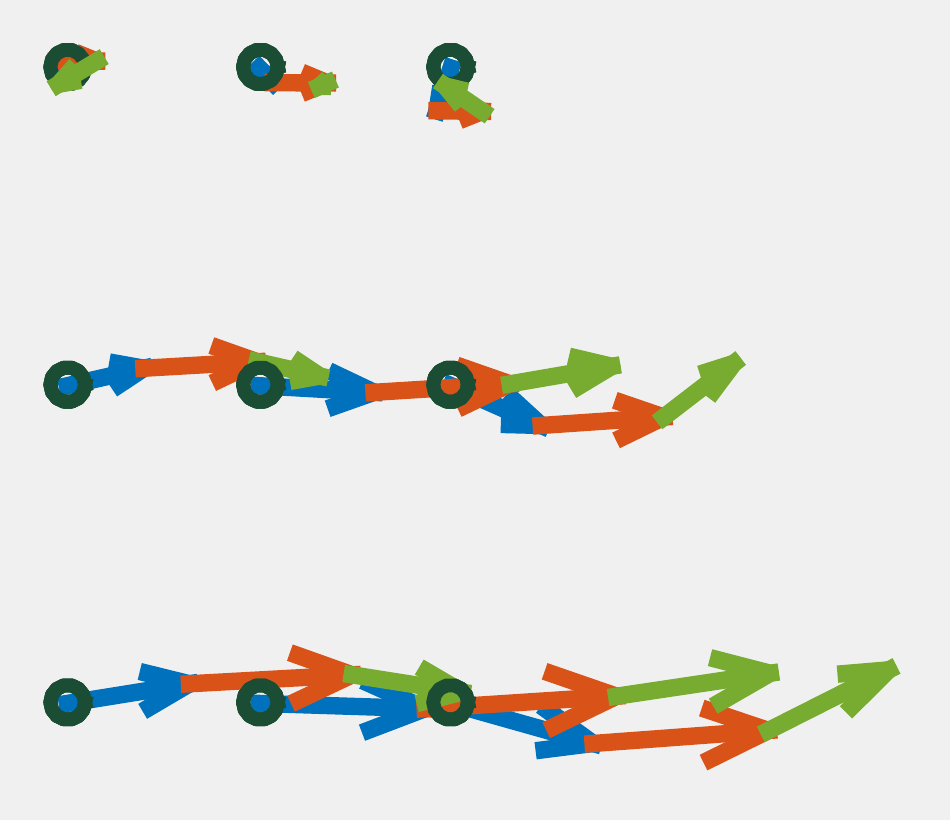}}
\frame{\includegraphics[width=\fs\columnwidth, height=\fs\columnwidth]{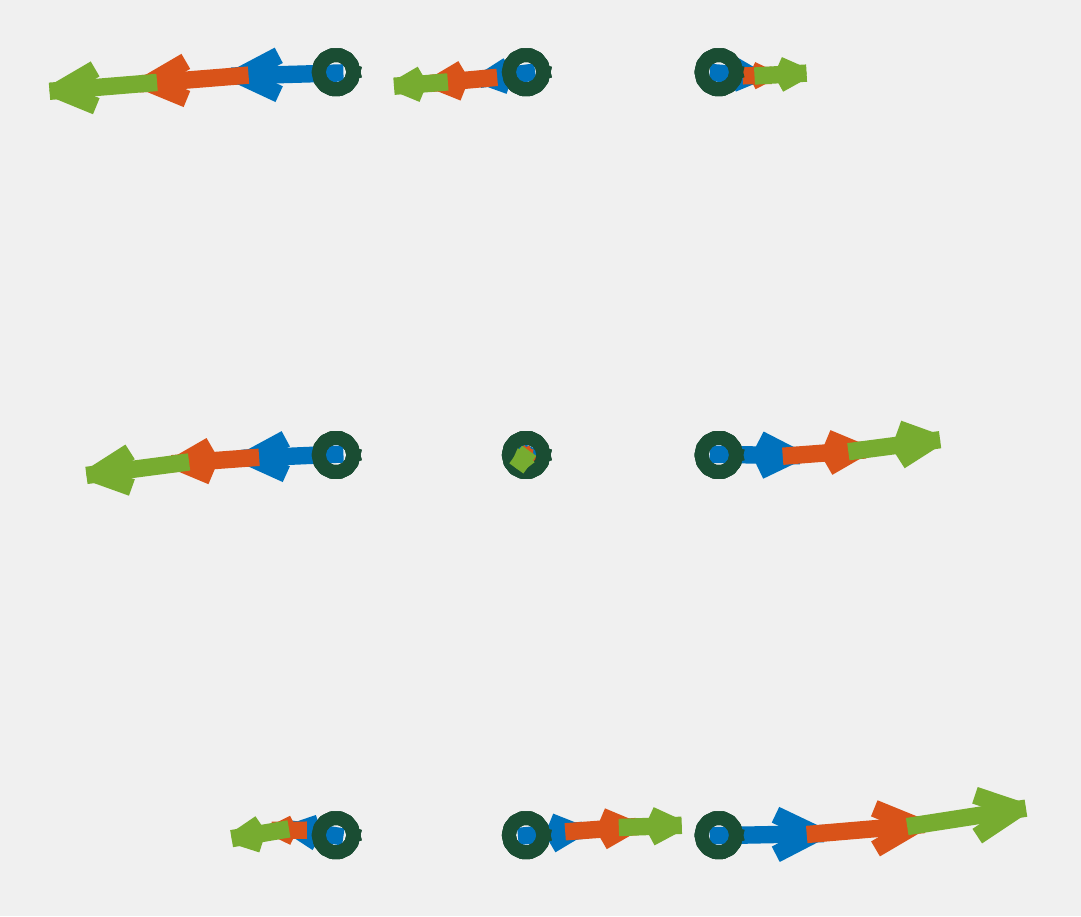}}
\frame{\includegraphics[width=\fs\columnwidth, height=\fs\columnwidth]{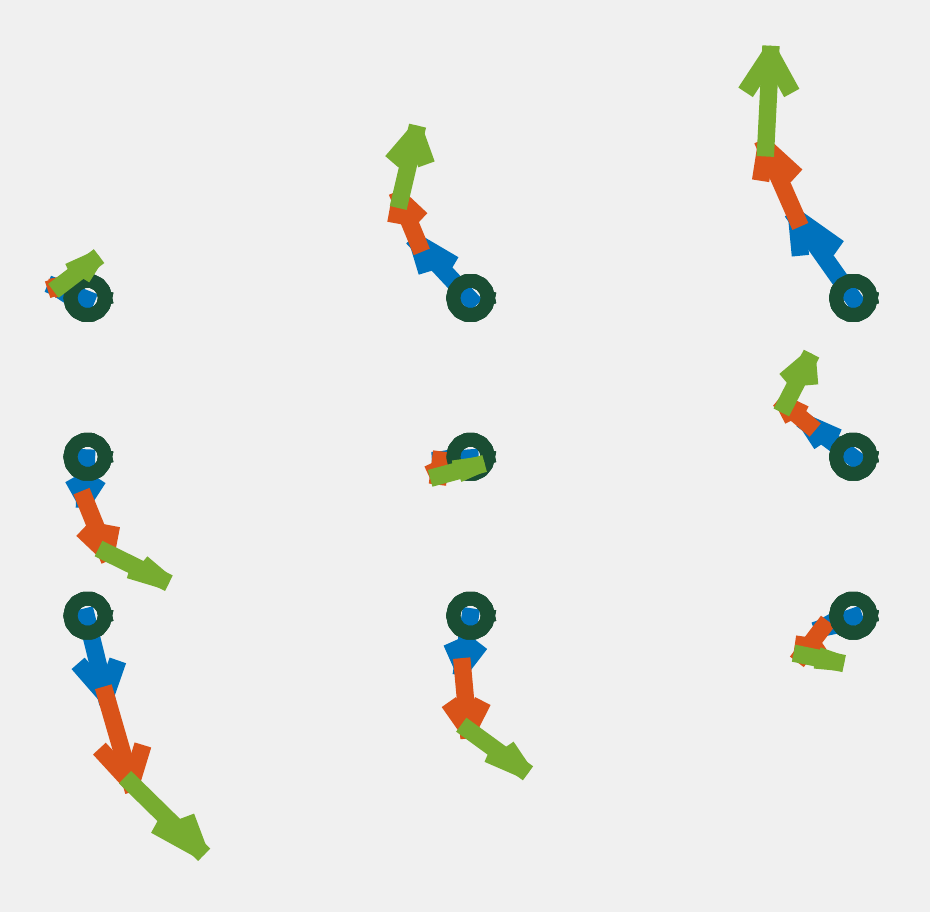}}
\frame{\includegraphics[width=\fs\columnwidth, height=\fs\columnwidth]{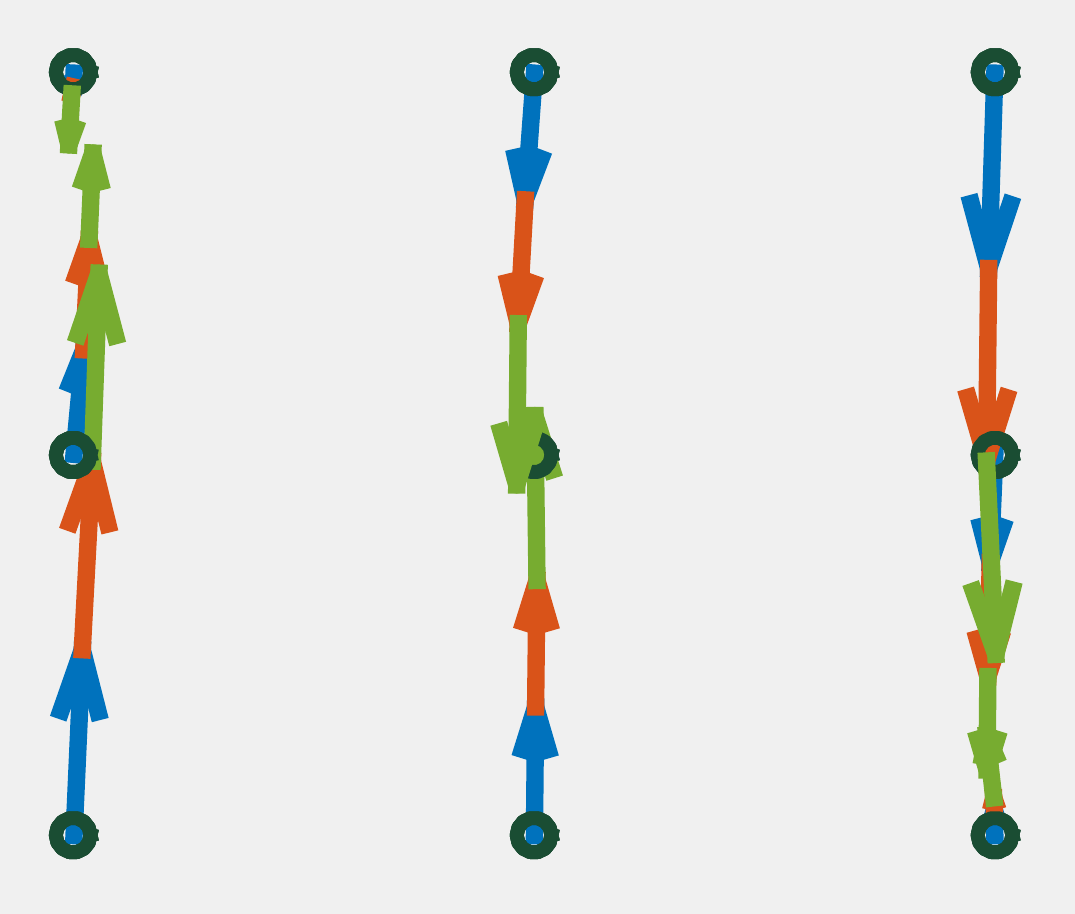}}
\mbox{}\vspace{-.5cm}\\
\end{center}
   \caption{Spatio-temporal filters from the first layer of the network learned with 2-channel, Brox optical flow and 60 frames on UCF101. 18 out of 64 filters are presented. Each cell in the grid represents two $3\times3\times3$ filters for 2-channel flow input (one for $x$ and one for $y$). $x$ and $y$ intensities are converted into vectors in 2D. Third dimension (time) is denoted by putting vectors one after the other in different colors for better visualization ($t$=1 blue, $t$=2 red, $t$=3 green). We see that LTC is able to learn complex motion patterns for video representation. Better viewed in color.}
\label{fig:filters}
\end{figure}

\paragraph*{First layer weights}
In order to have an intuition of
what an LTC network learns, we visualize the first layer space-time
convolutional filters in the vector-field form. Filters learned on
2-channel optical flow vectors have dimension $2\times3\times3\times3$
in terms of
channels, width, height and time. For each filter, we take
the two channels in each $3\times3\times3$ volume and visualize them
as vectors using x- and y-components. Figure~\ref{fig:filters} shows
18 example filters out of the 64 from a network learned on UCF101 with
60 frames flow input. Since our filters are spatio-temporal, they have a
third dimension in time. We find it convenient to show them as vectors
concatenated one after the other with regard to the time steps.
We denote each time step with
different colors and see that the filters learned by long-term
temporal convolutions are able to represent complex motions in local
neighborhoods, which enables to incorporate even more complex patterns
in later stages of the network.

\begin{figure}
\centering
\subfigure[Top activations of filters at {\em conv3-conv5} layers. Each row is another layer, indicated by L3-L5. Left is for 100 frames and right is for 16 frames networks. Colors indicate different action classes. Each color plot illustrates distribution of classes for seven top activations of 30 selected filters. Rows are maximum responding test videos and columns are filters.]{
\begin{tabular}{@{\hskip -0.1ex}c@{\hskip 1ex}cc}
& \textbf{100f} & \textbf{16f} \\
\rotatebox[origin=l]{90}{\textbf{L5}} &
\includegraphics[width=0.21\textwidth]{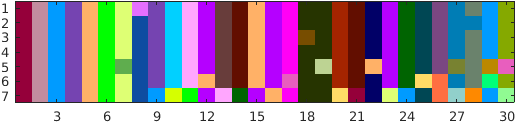}
&
\includegraphics[width=0.21\textwidth]{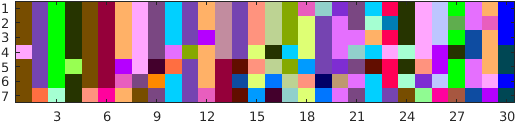}\\
\rotatebox[origin=l]{90}{\textbf{L4}} &
\includegraphics[width=0.21\textwidth]{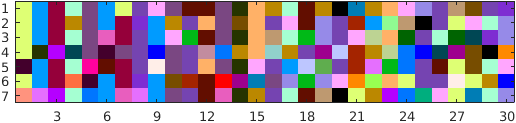}
&
\includegraphics[width=0.21\textwidth]{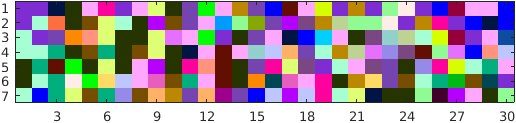}\\
\rotatebox[origin=l]{90}{\textbf{L3}} &
\includegraphics[width=0.21\textwidth]{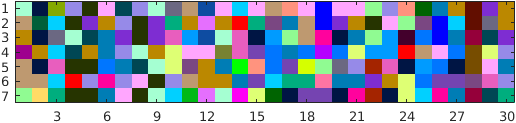}
&
\includegraphics[width=0.21\textwidth]{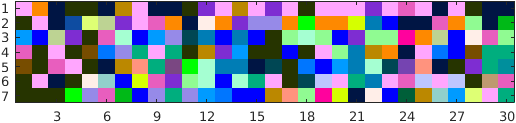}\\
\end{tabular}
\label{fig:topactivationscolors}
} \\
\mbox{}\vspace{-.3cm}\\
\subfigure[Frames corresponding to videos with top activations at {\em conv4} and {\em conv5} layers. Circles indicate the spatial location of the maximum response. The visualized frames correspond to the maximum response in time.]{
\begin{tabular}{@{\hskip -0.5ex}cc@{\hskip 1ex}cc@{\hskip 1ex}c@{\hskip 1ex}c}
& \multicolumn{2}{c}{\textbf{100f}} && \multicolumn{2}{c}{\textbf{16f}} \\ \cline{2-3} \cline{5-6}

& F1 & F2 & & F1 & F2\\
\parbox[t]{2mm}{\multirow{3}{*}{\rotatebox[origin=c]{90}{\textbf{Layer 5}}}} &
\includegraphics[width=\as\columnwidth, height=\as\columnwidth]{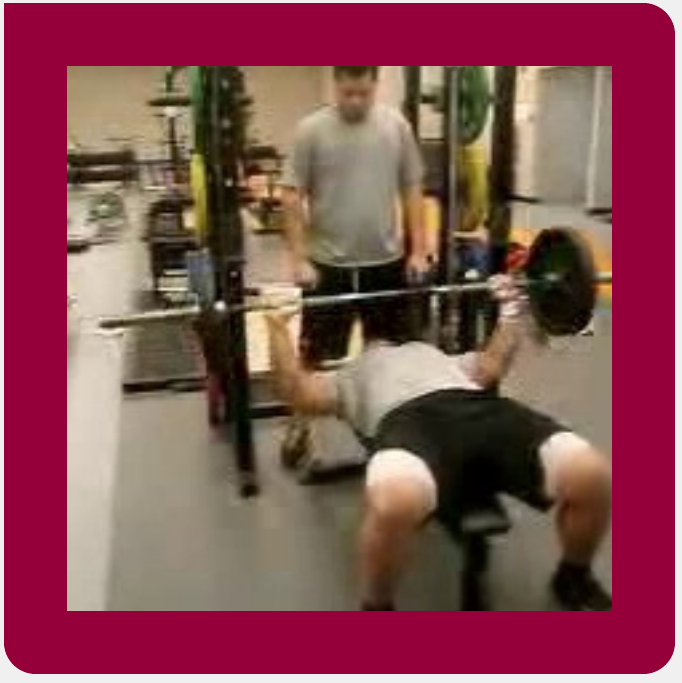} \hspace{-\sep ex}
\includegraphics[width=\as\columnwidth, height=\as\columnwidth]{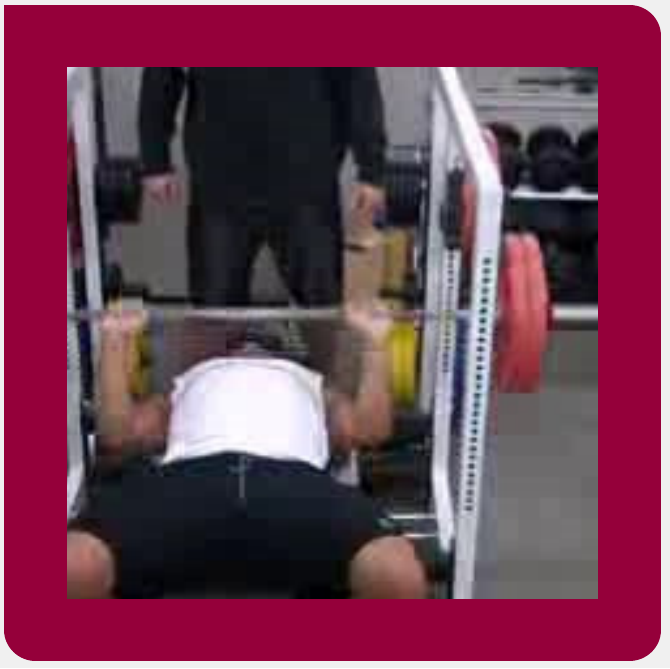} &
\includegraphics[width=\as\columnwidth, height=\as\columnwidth]{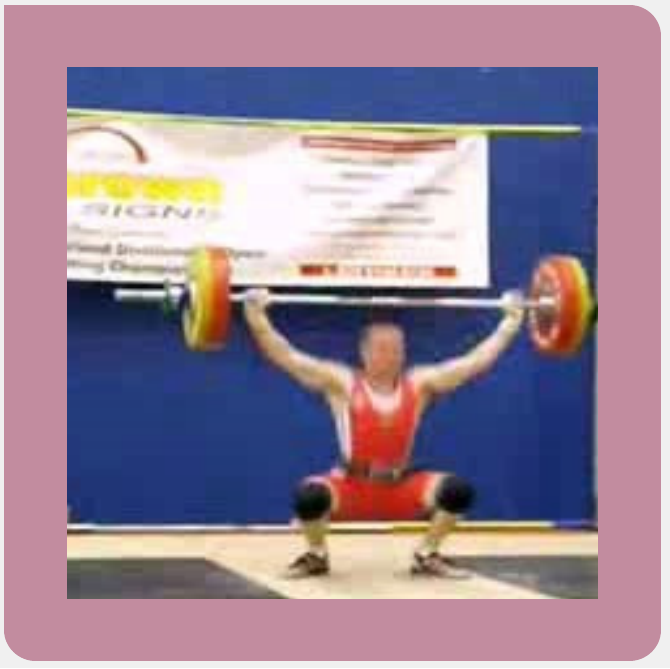} \hspace{-\sep ex}
\includegraphics[width=\as\columnwidth, height=\as\columnwidth]{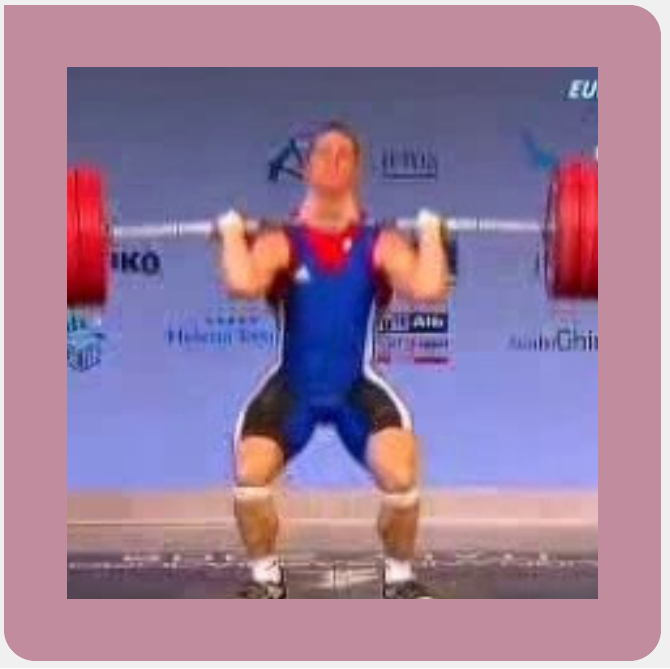}  &&
\includegraphics[width=\as\columnwidth, height=\as\columnwidth]{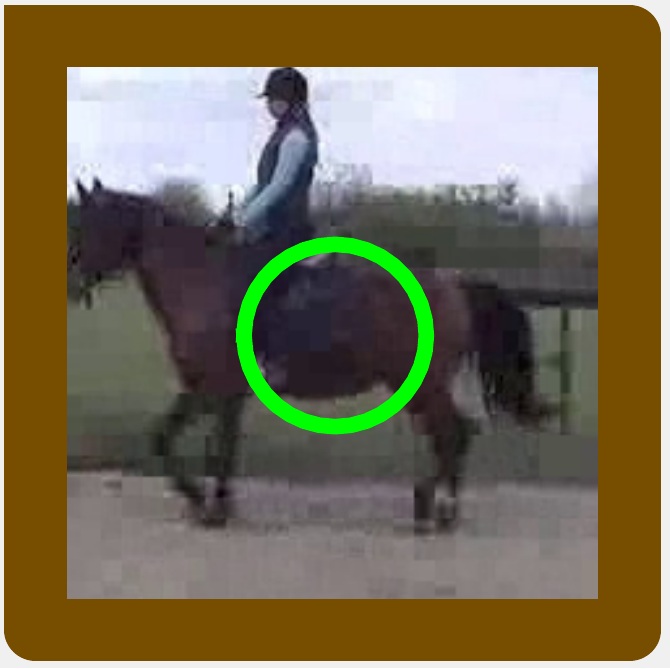} \hspace{-\sep ex}
\includegraphics[width=\as\columnwidth, height=\as\columnwidth]{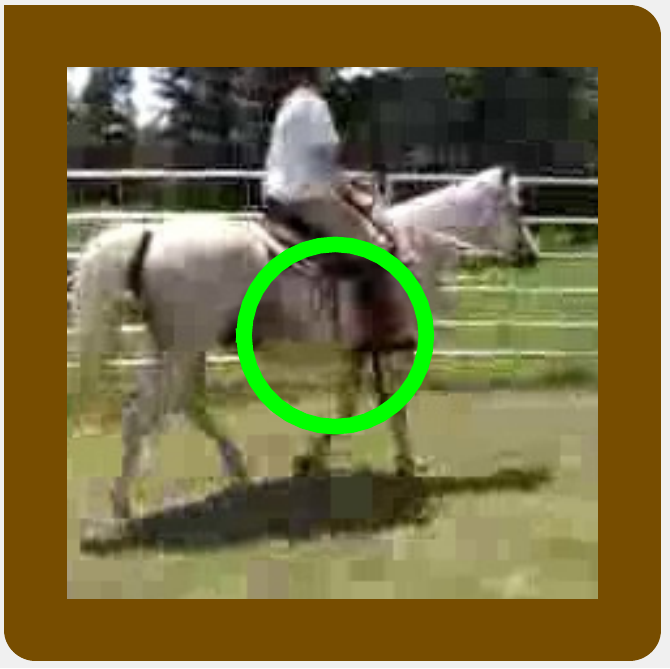} &
\includegraphics[width=\as\columnwidth, height=\as\columnwidth]{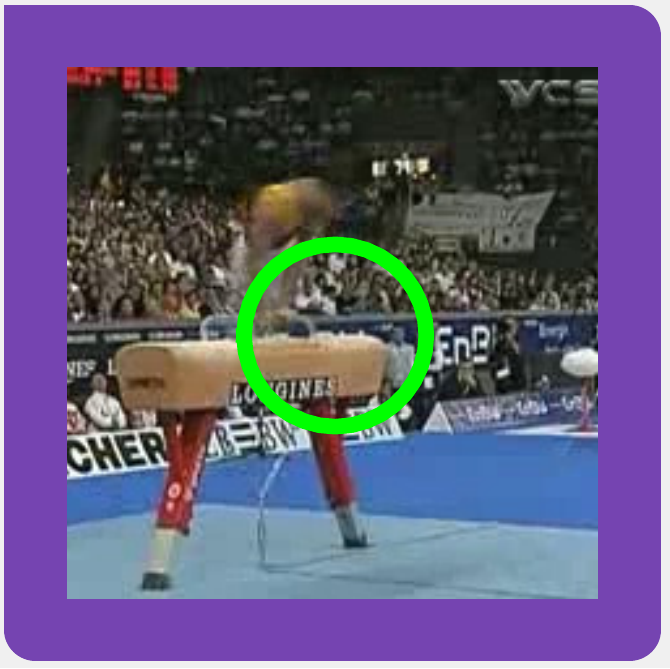} \hspace{-\sep ex}
\includegraphics[width=\as\columnwidth, height=\as\columnwidth]{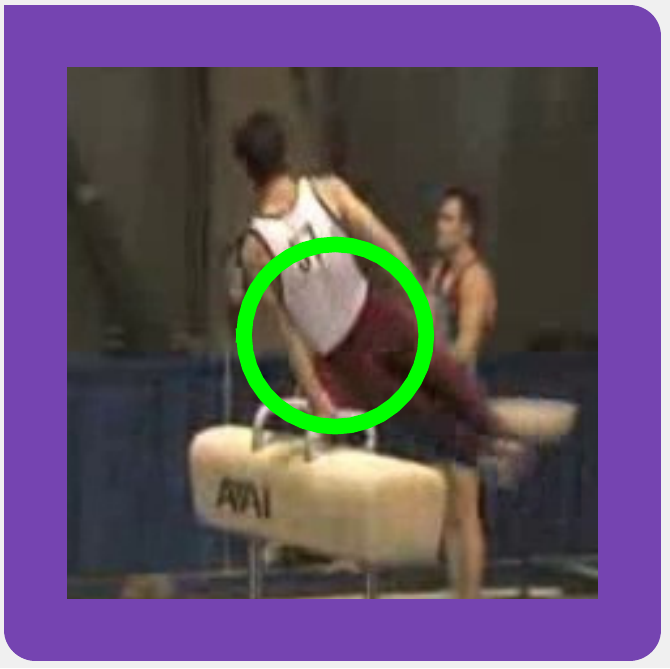} \\[-1ex]
&
\includegraphics[width=\as\columnwidth, height=\as\columnwidth]{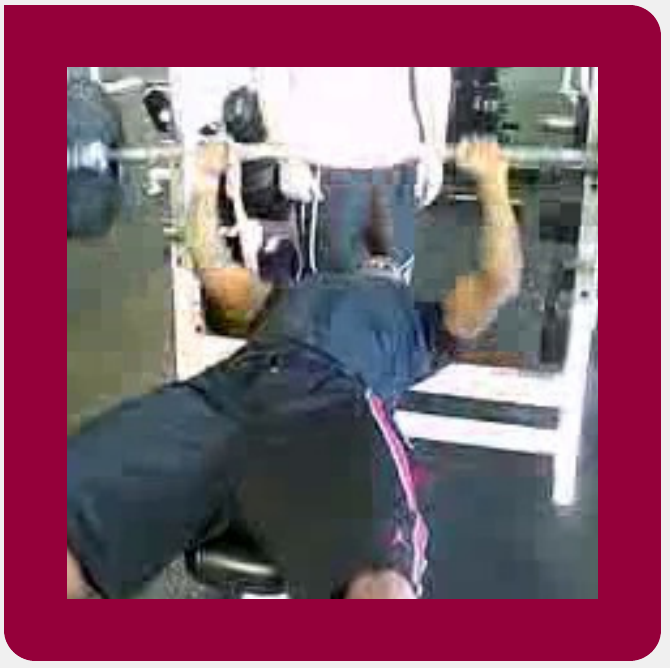} \hspace{-\sep ex}
\includegraphics[width=\as\columnwidth, height=\as\columnwidth]{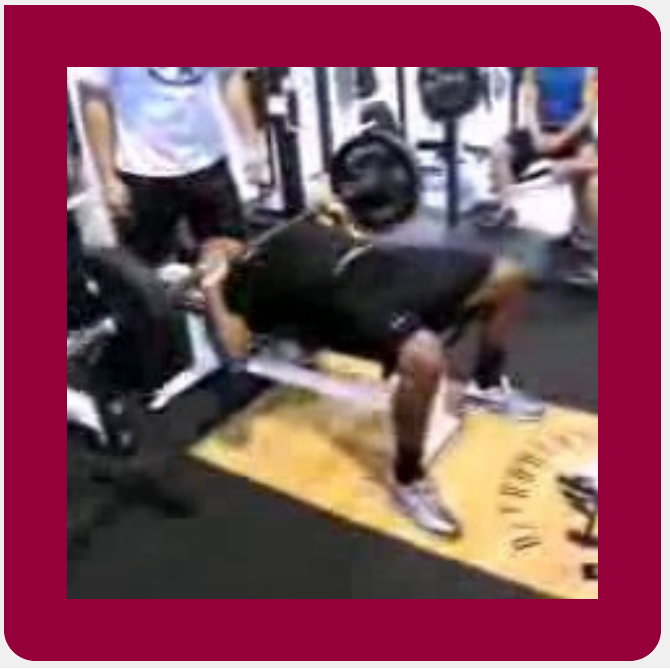} &
\includegraphics[width=\as\columnwidth, height=\as\columnwidth]{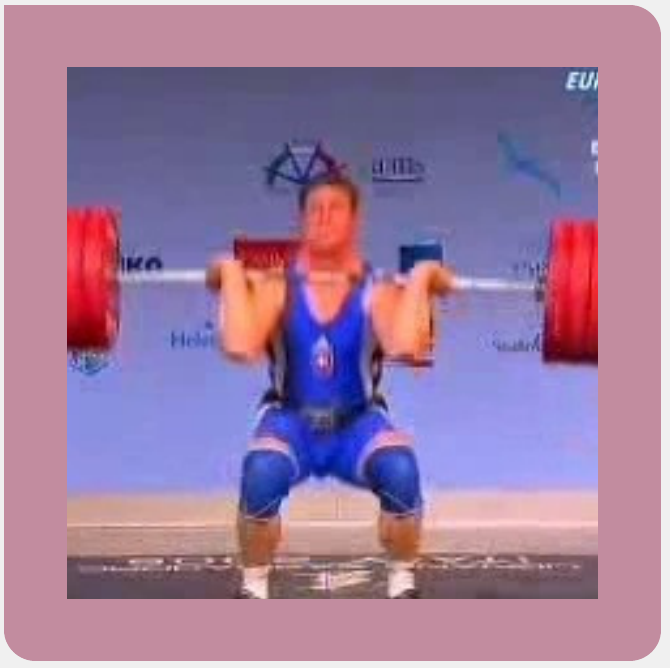} \hspace{-\sep ex}
\includegraphics[width=\as\columnwidth, height=\as\columnwidth]{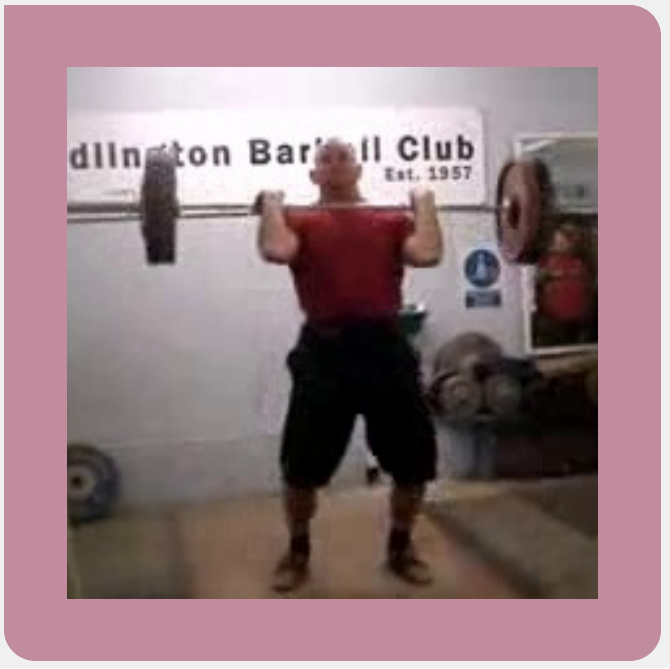}  &&
\includegraphics[width=\as\columnwidth, height=\as\columnwidth]{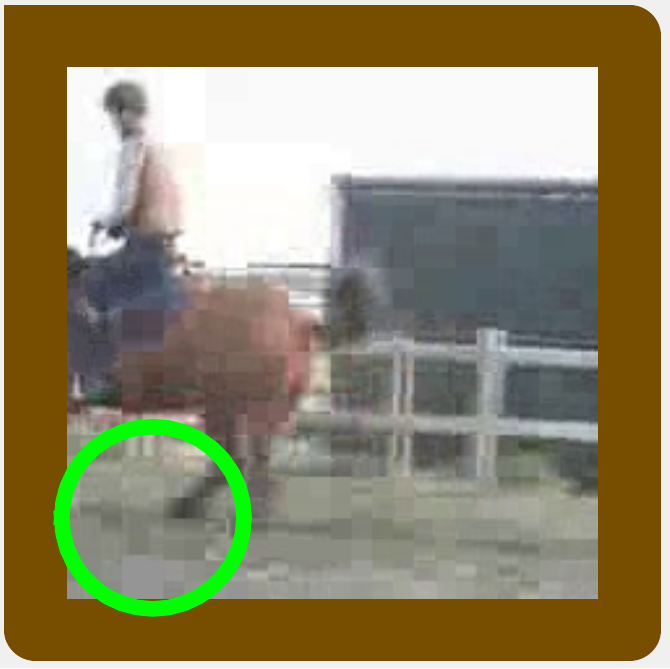} \hspace{-\sep ex}
\includegraphics[width=\as\columnwidth, height=\as\columnwidth]{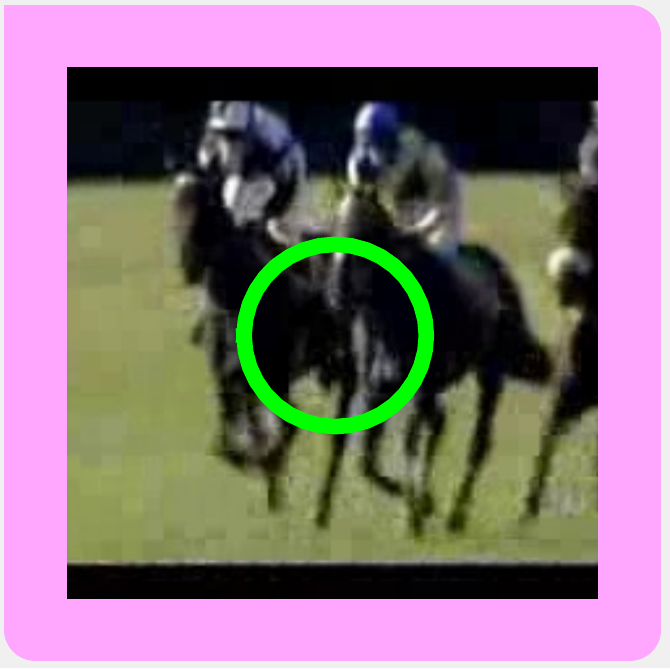} &
\includegraphics[width=\as\columnwidth, height=\as\columnwidth]{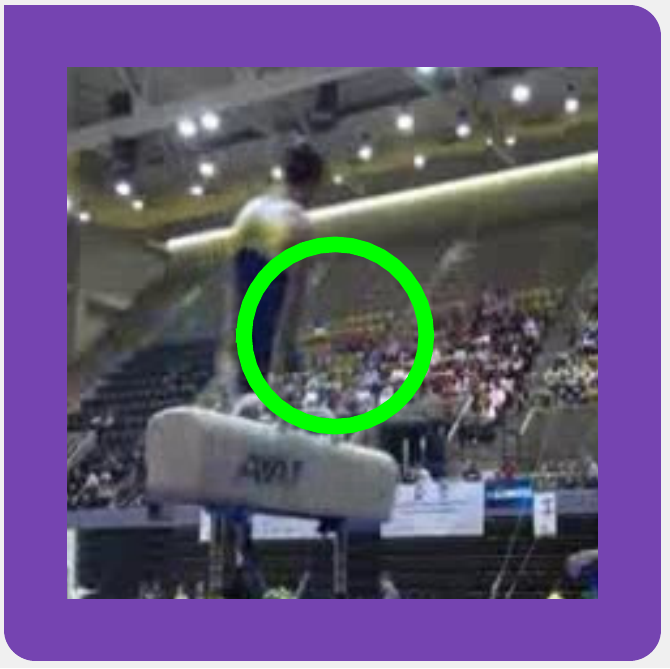} \hspace{-\sep ex}
\includegraphics[width=\as\columnwidth, height=\as\columnwidth]{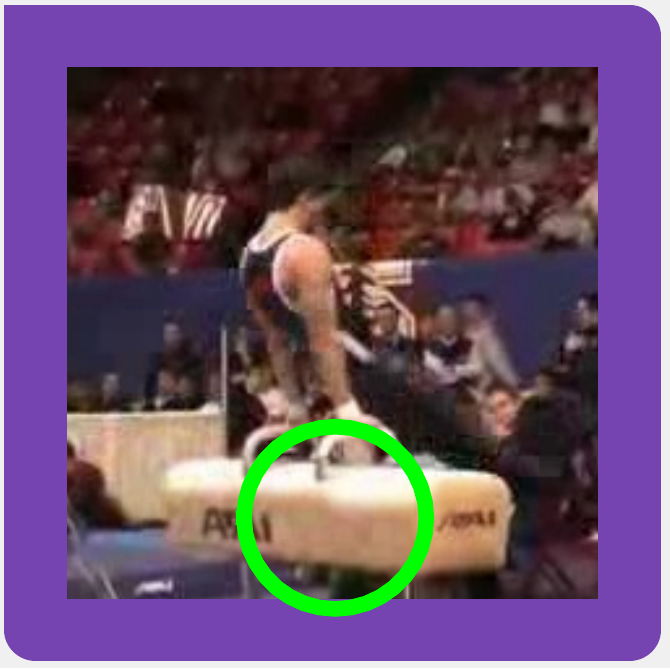} \\[-1ex]
&
\includegraphics[width=\as\columnwidth, height=\as\columnwidth]{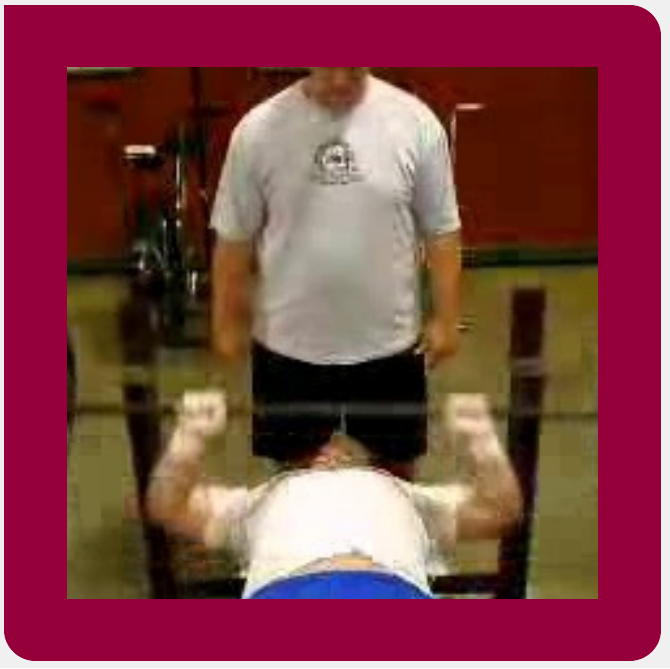} \hspace{-\sep ex}
\includegraphics[width=\as\columnwidth, height=\as\columnwidth]{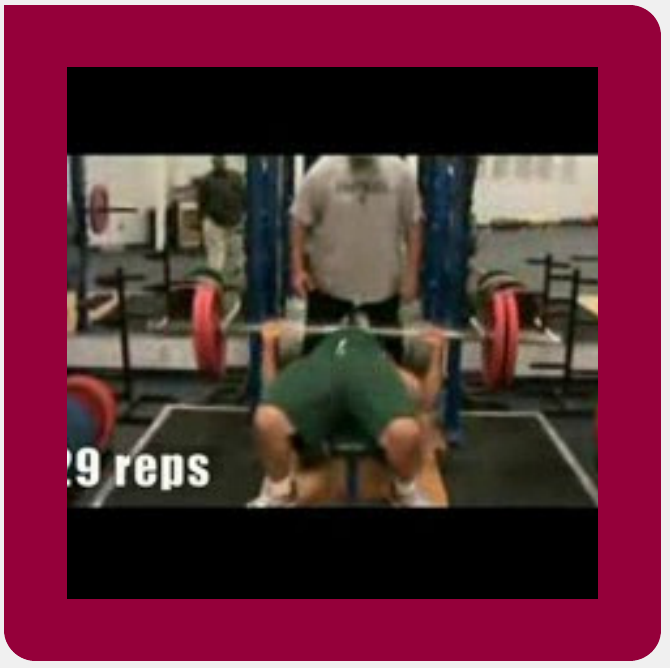} &
\includegraphics[width=\as\columnwidth, height=\as\columnwidth]{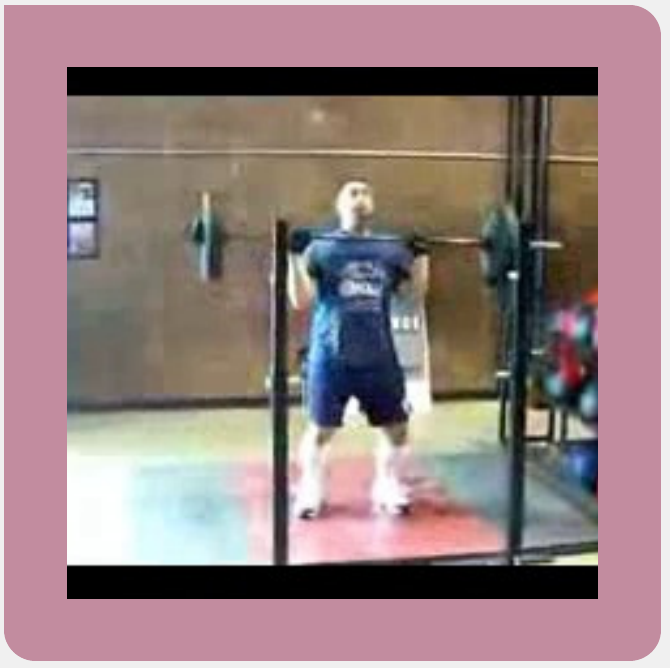} \hspace{-\sep ex}
\includegraphics[width=\as\columnwidth, height=\as\columnwidth]{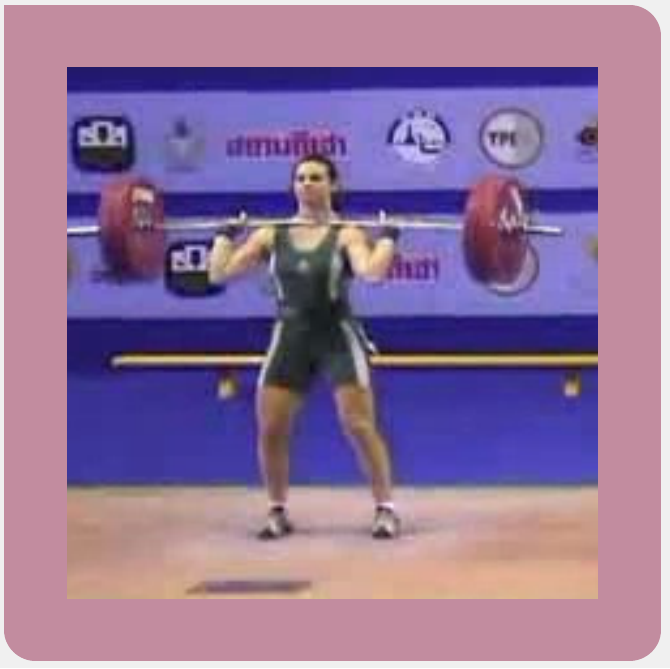}  &&
\includegraphics[width=\as\columnwidth, height=\as\columnwidth]{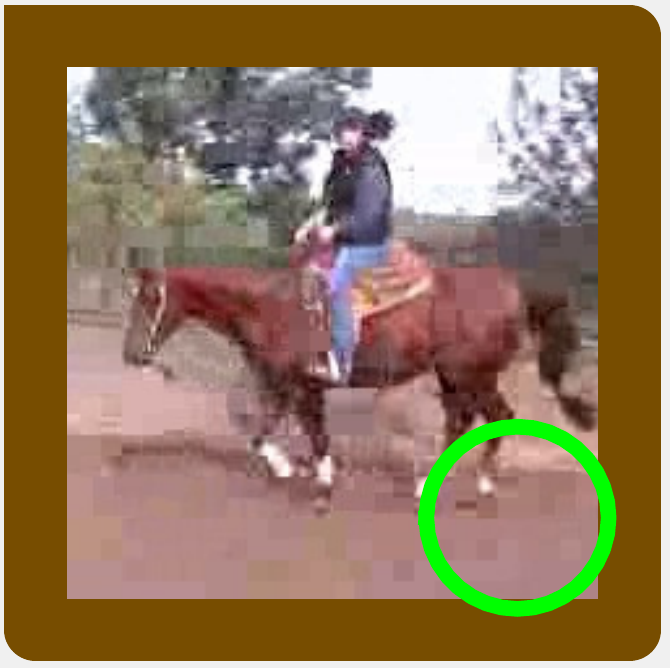} \hspace{-\sep ex}
\includegraphics[width=\as\columnwidth, height=\as\columnwidth]{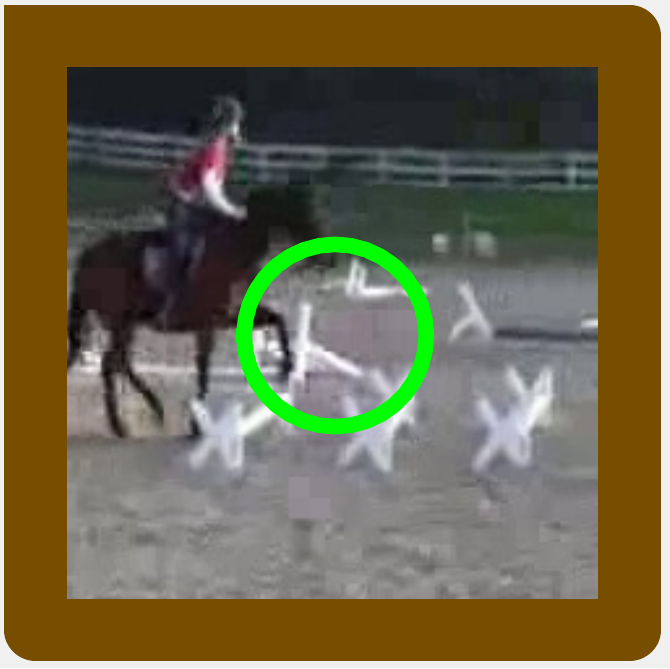} &
\includegraphics[width=\as\columnwidth, height=\as\columnwidth]{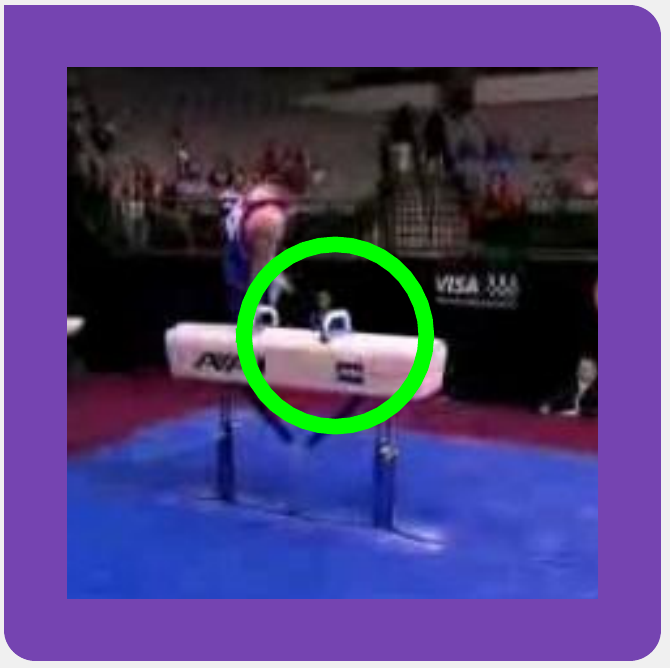} \hspace{-\sep ex}
\includegraphics[width=\as\columnwidth, height=\as\columnwidth]{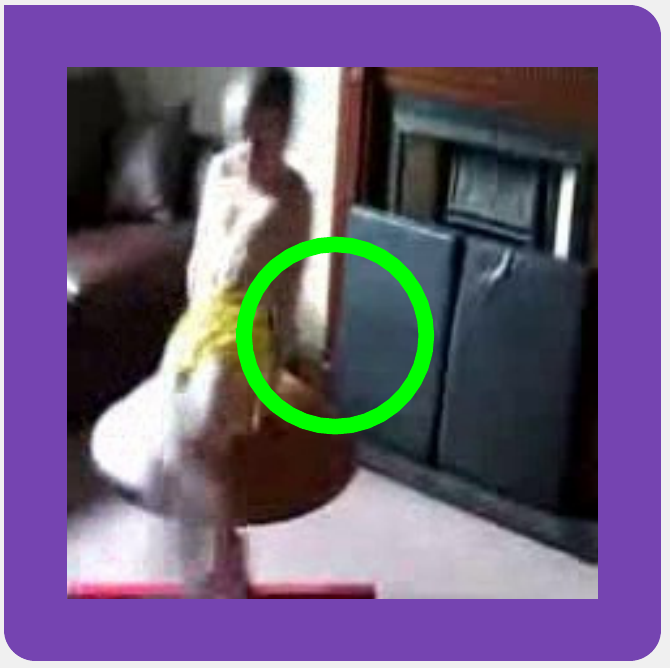} \\[2ex]
\parbox[t]{2mm}{\multirow{3}{*}{\rotatebox[origin=c]{90}{\textbf{Layer 4}}}} &
\includegraphics[width=\as\columnwidth]{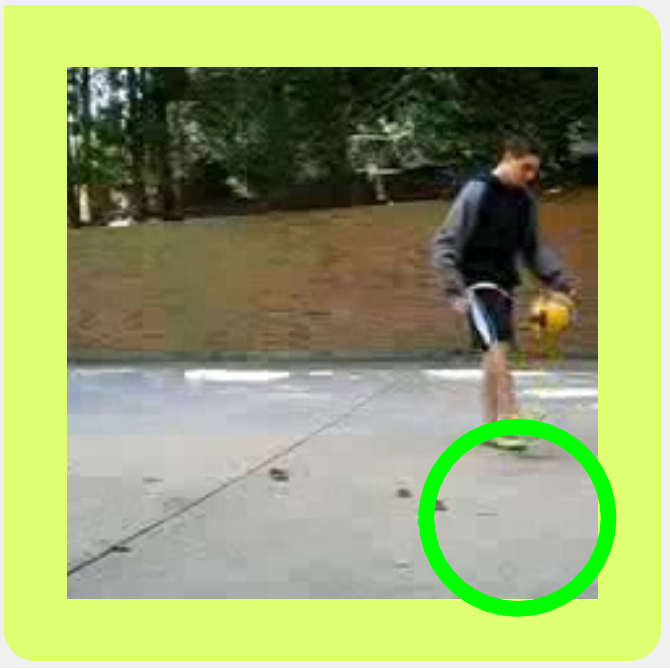} \hspace{-\sep ex}
\includegraphics[width=\as\columnwidth, height=\as\columnwidth]{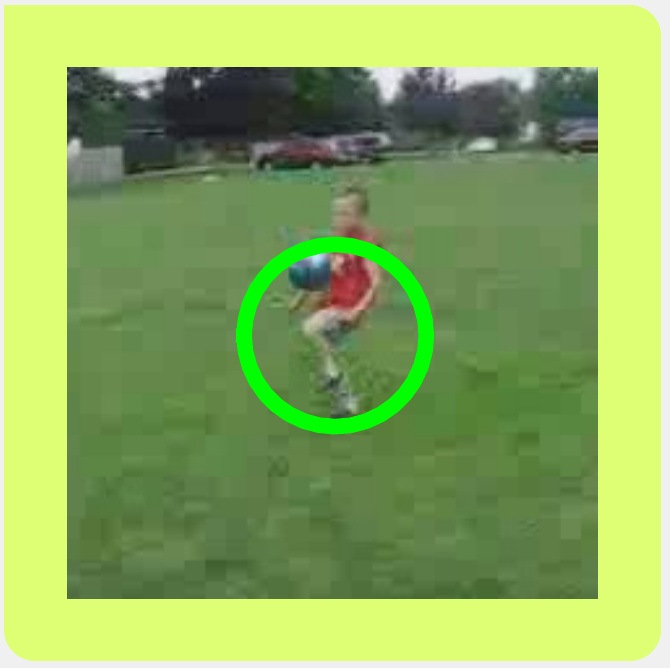} &
\includegraphics[width=\as\columnwidth, height=\as\columnwidth]{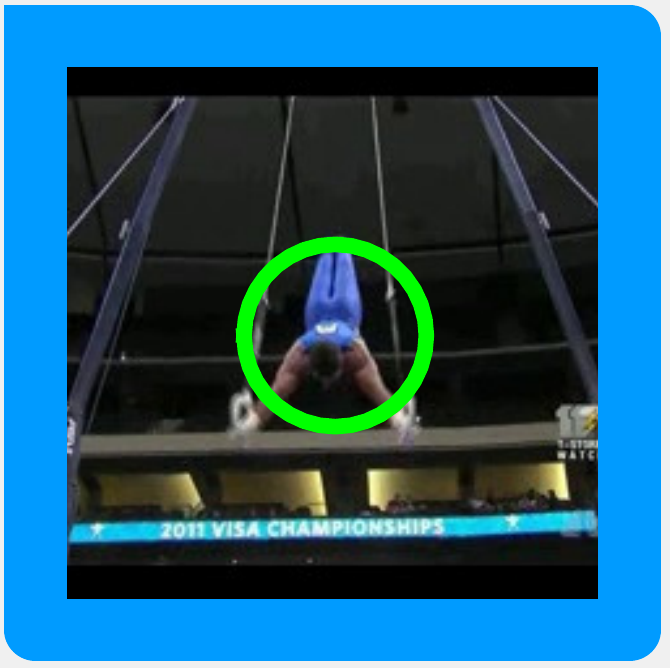} \hspace{-\sep ex}
\includegraphics[width=\as\columnwidth, height=\as\columnwidth]{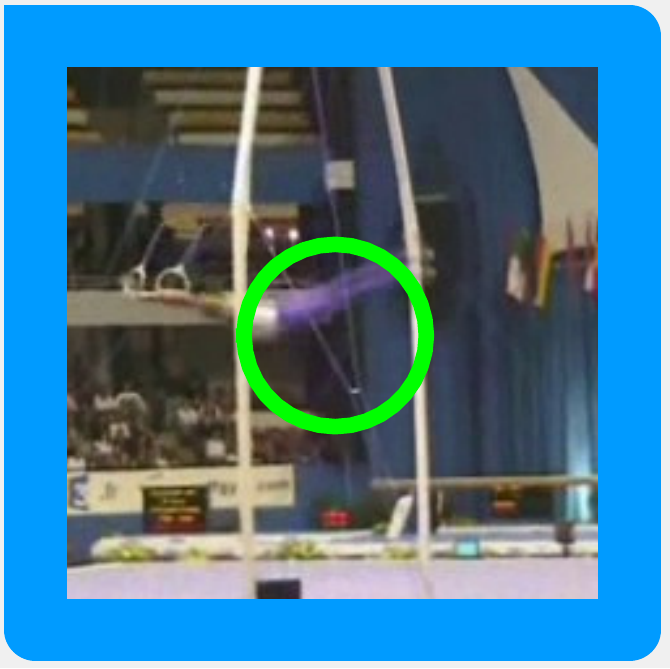}  &&
\includegraphics[width=\as\columnwidth, height=\as\columnwidth]{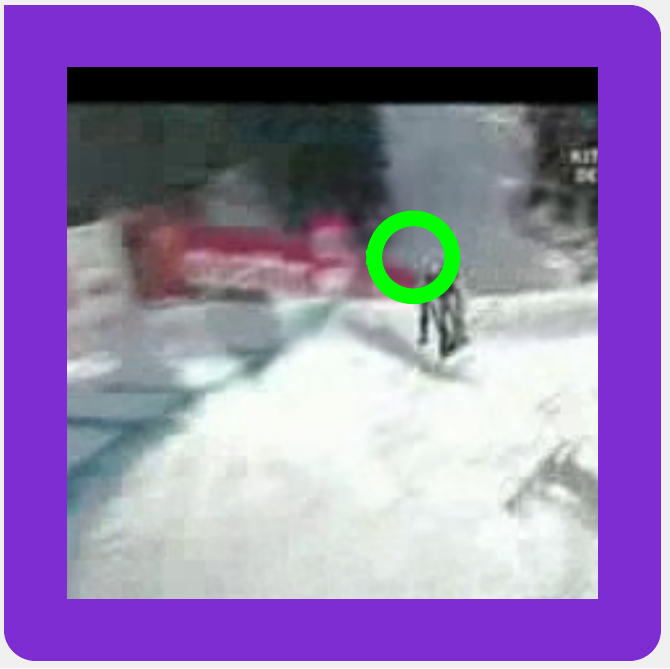} \hspace{-\sep ex}
\includegraphics[width=\as\columnwidth, height=\as\columnwidth]{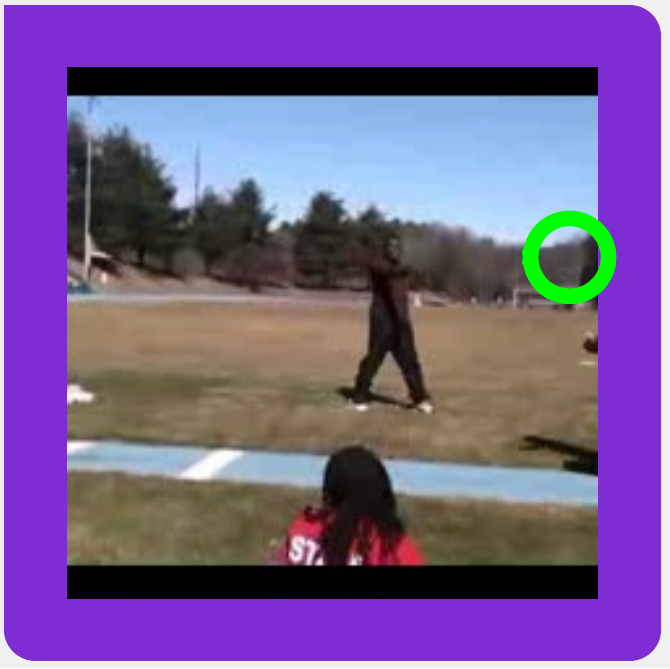} &
\includegraphics[width=\as\columnwidth, height=\as\columnwidth]{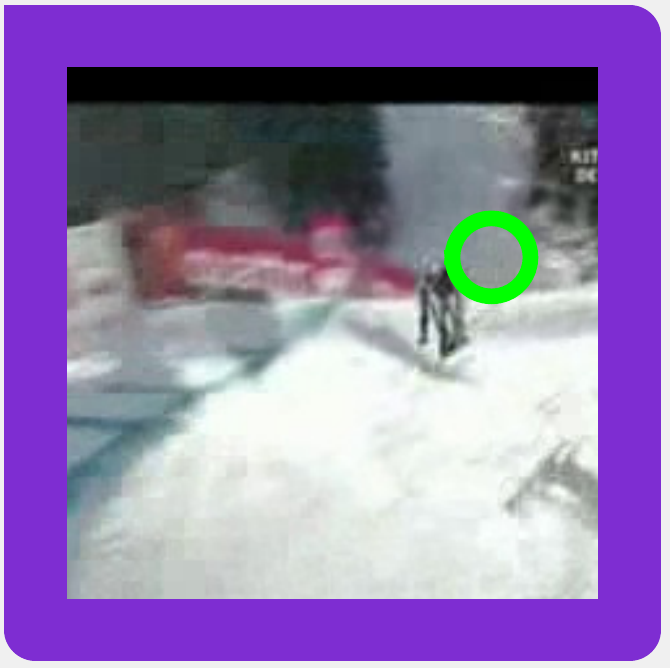} \hspace{-\sep ex}
\includegraphics[width=\as\columnwidth, height=\as\columnwidth]{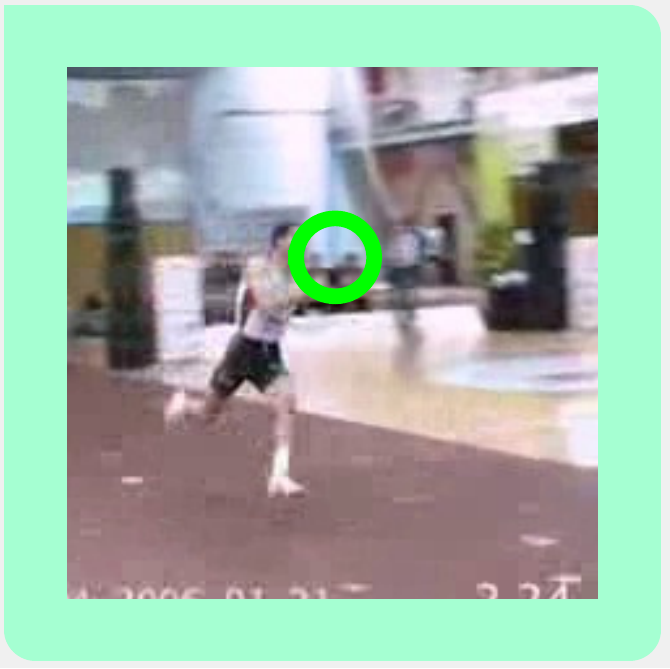} \\[-1ex]
&
\includegraphics[width=\as\columnwidth, height=\as\columnwidth]{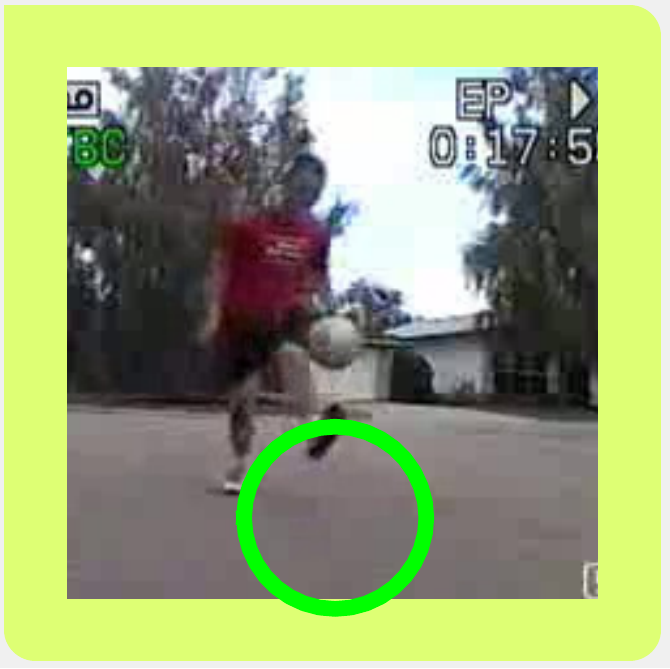} \hspace{-\sep ex}
\includegraphics[width=\as\columnwidth, height=\as\columnwidth]{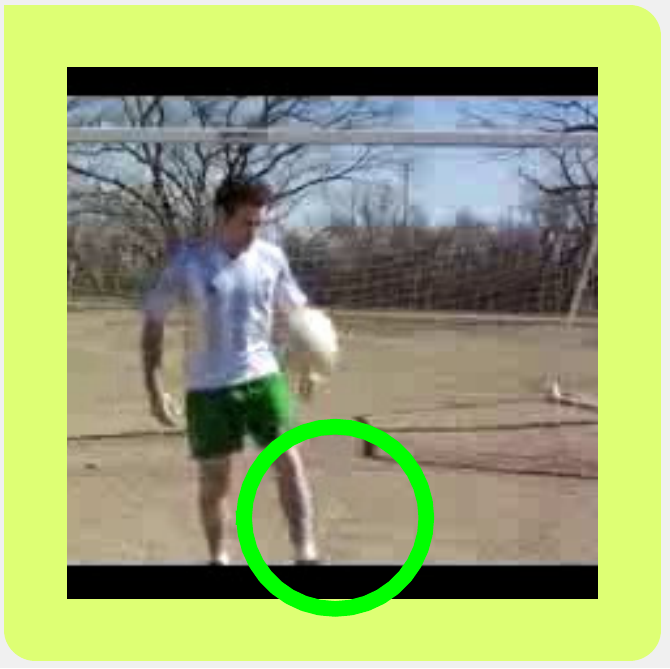} &
\includegraphics[width=\as\columnwidth, height=\as\columnwidth]{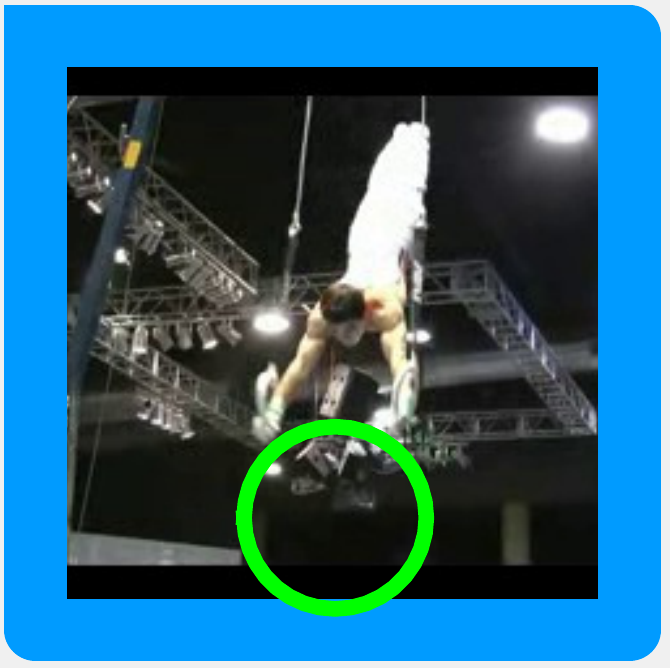} \hspace{-\sep ex}
\includegraphics[width=\as\columnwidth, height=\as\columnwidth]{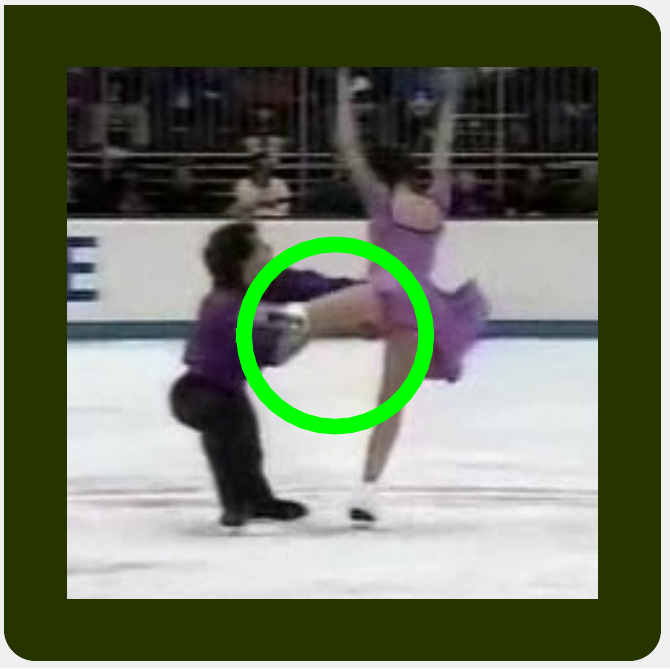}  &&
\includegraphics[width=\as\columnwidth, height=\as\columnwidth]{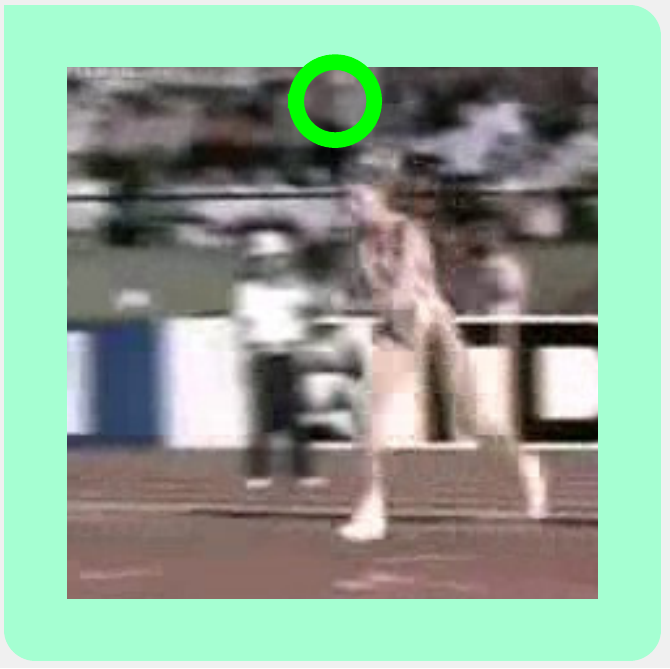} \hspace{-\sep ex}
\includegraphics[width=\as\columnwidth, height=\as\columnwidth]{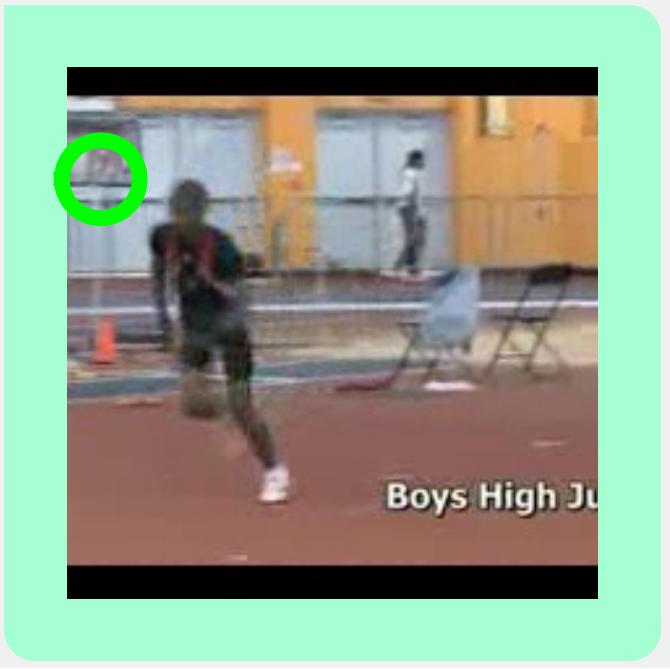} &
\includegraphics[width=\as\columnwidth, height=\as\columnwidth]{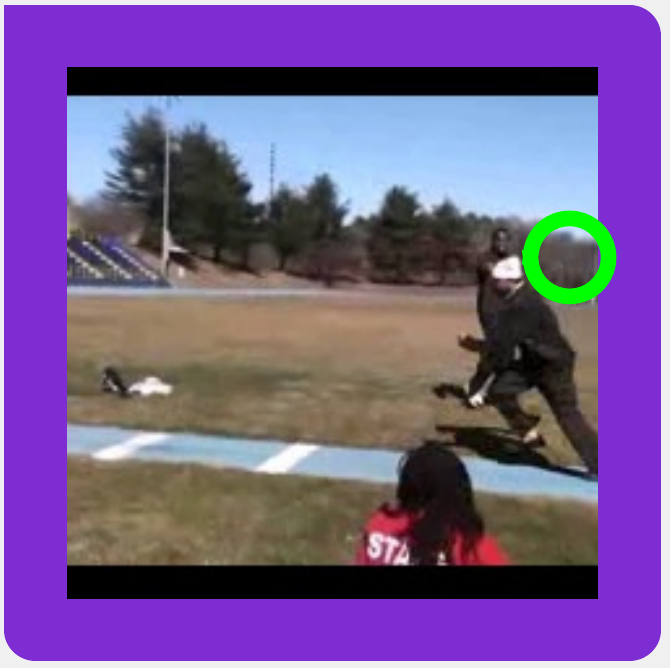} \hspace{-\sep ex}
\includegraphics[width=\as\columnwidth, height=\as\columnwidth]{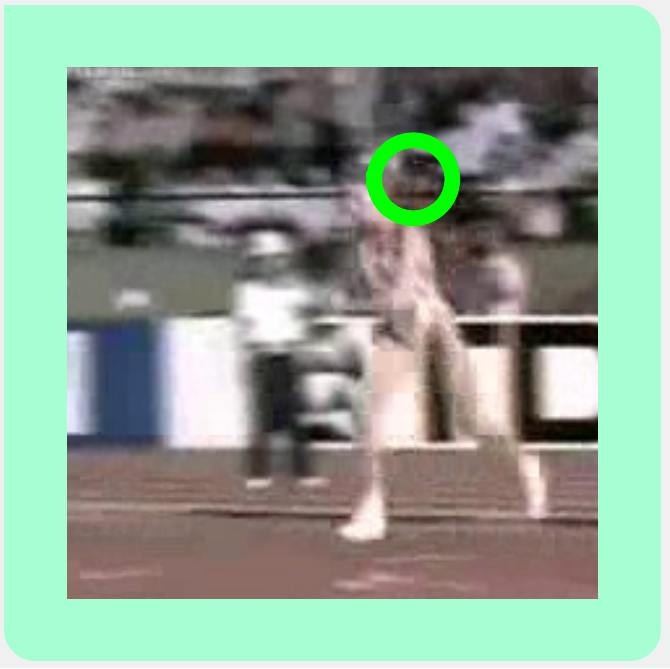} \\[-1ex]
&
\includegraphics[width=\as\columnwidth, height=\as\columnwidth]{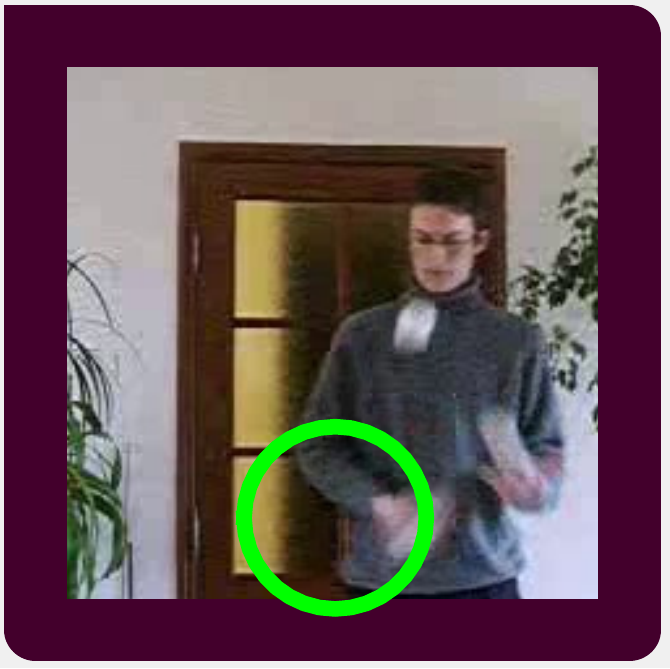} \hspace{-\sep ex}
\includegraphics[width=\as\columnwidth, height=\as\columnwidth]{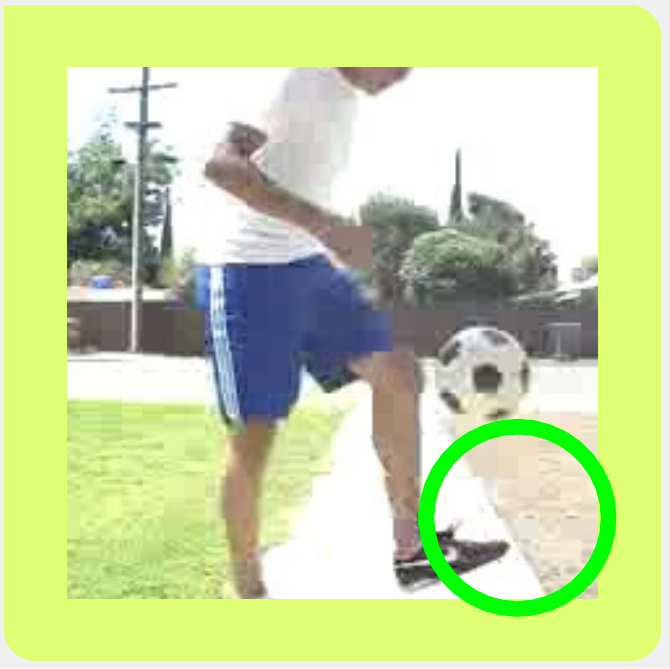} &
\includegraphics[width=\as\columnwidth, height=\as\columnwidth]{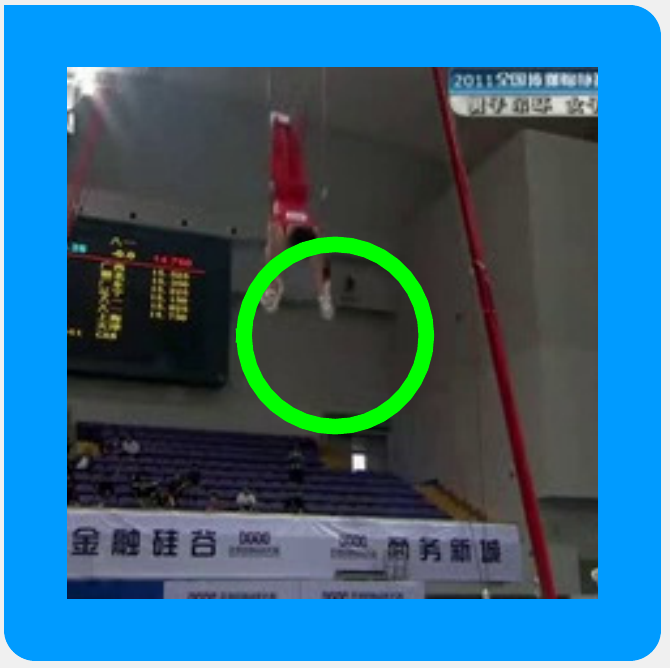} \hspace{-\sep ex}
\includegraphics[width=\as\columnwidth, height=\as\columnwidth]{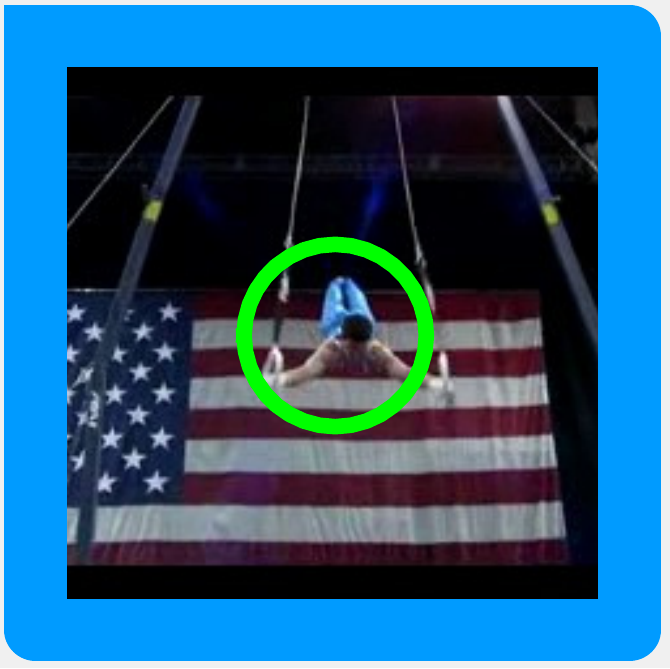}  &&
\includegraphics[width=\as\columnwidth, height=\as\columnwidth]{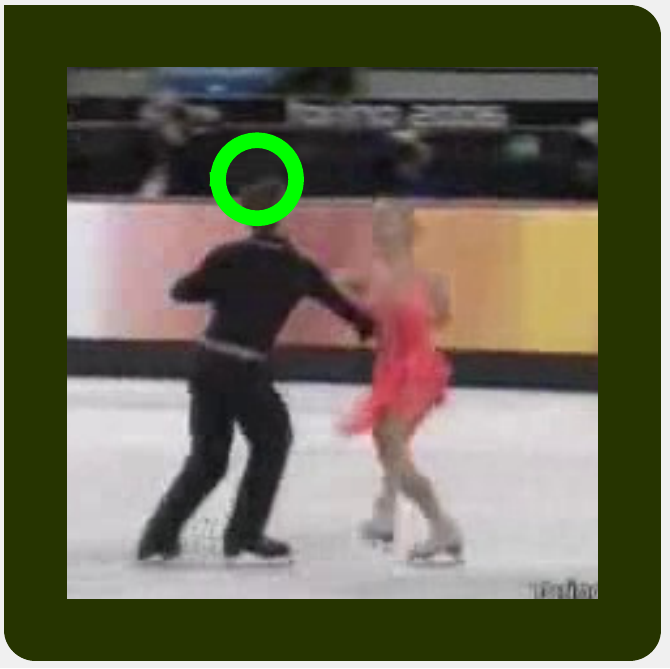} \hspace{-\sep ex}
\includegraphics[width=\as\columnwidth, height=\as\columnwidth]{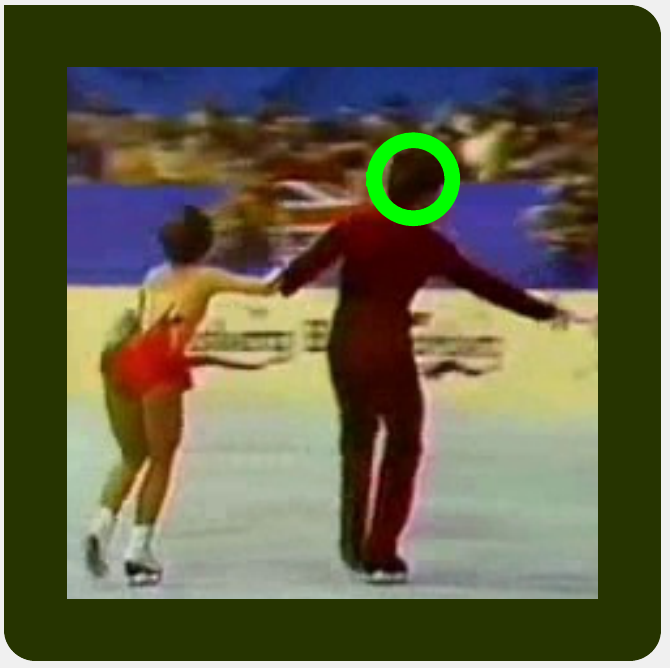} &
\includegraphics[width=\as\columnwidth, height=\as\columnwidth]{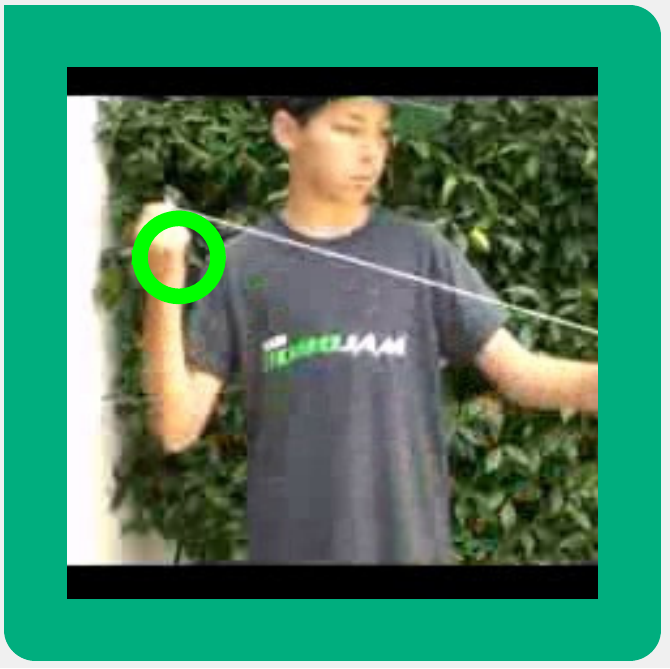} \hspace{-\sep ex}
\includegraphics[width=\as\columnwidth, height=\as\columnwidth]{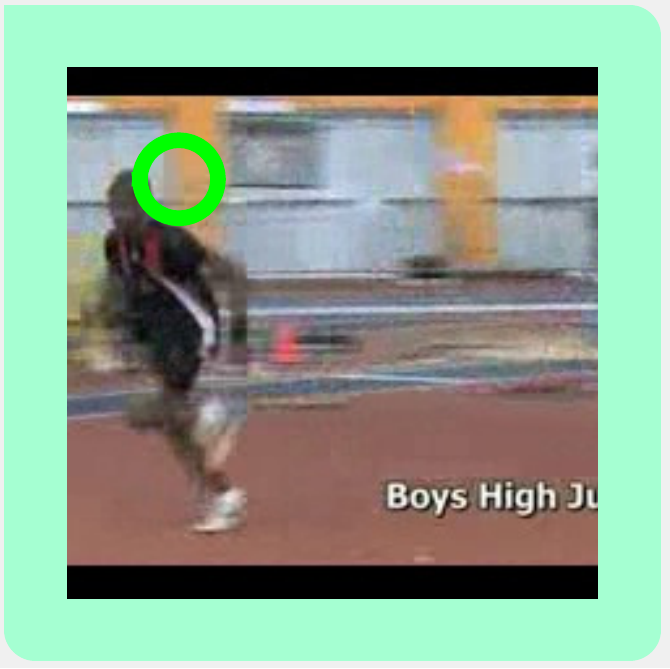} \\
\end{tabular}
 \label{fig:topactivations}
}
\mbox{}\vspace{-.2cm}\\
\caption{Comparison of 100f and 16f networks by looking at the top activations of filters. Better viewed in color.}
\mbox{}\vspace{-.5cm}\\
\end{figure}

\vspace{-0.1cm}
\paragraph*{High-layer filter activations}
We further investigate filters from higher convolutional layers by 
examining their highest activations. For a given layer and a chosen 
filter, we record the maximum activation value for all test 
videos\footnote{UCF101 videos are obtained by clipping different 
parts ({\em video}) from a longer video ({\em group}). We take 
one {\em video} per {\em group} assuming that {\em videos} 
from the same {\em group} would have similar activations and would 
avoid a proper analysis. In total, there are 7 test {\em groups} 
per class; therefore there can be at most 7 {\em videos} belonging 
to a class.} for that filter. We then sort test videos according 
to the activation values and select the top 7 videos. This procedure 
is similar to~\cite{zeiler_vis}. We can expect that a filter should 
be activated by similar action classes especially at the higher 
network layers. Given longer video clips available to the LTC 
networks, we also expect better grouping of actions from the same 
class by filter activations of LTC. We illustrate action classes 
for 30 filters (x-axis) and their top 7 activations (y-axis) for 
the 100f and 16f networks in Figure~\ref{fig:topactivationscolors}. 
Each action class is represented by a unique color. The filters 
are sorted by their purity, i.e.\ the frequency of the dominating 
class. We assign each video the color of its ground truth class 
label. We see that the clustering of videos from the same class 
becomes more clear in higher layers in the network for both 16f 
and 100f networks. However, it is evident that 100f filters have 
more purity than 16f even in L4 and L3. Note that 16f network is 
trained with high resolution ($112\times 112$) flow and 100f 
network with low resolution ($58\times 58$) flow.

Example frames from top-scoring videos for a set of selected 
filters $f$ are shown in Figure~\ref{fig:topactivations} for 
16f and 100f flow networks. We also provide a video on our 
project web page~\cite{projectpage} to show which videos activate 
for these filters. We can observe that for filters $f$ maximizing 
the homogeneity of returned class labels, the top activations for 
filters of the 100f network result in videos with similar action 
classes. The grouping of videos by classes is less prominent for 
activations of the 16f network. This result indicates that the 
LTC networks have higher level of abstraction at corresponding 
convolution layers when compared to networks with smaller temporal extents.

\subsection{Runtime}
\label{ss:runtime}
Training on UCF101 takes 1.9 day for 100f (58x58) networks 
and 1.1 day for 16f (112x112) networks with 0.5 dropout. 
At test time (without flow computation), the 100f and 16f networks 
run at 4452fps and 1128fps respectively on a Titan X GPU and
8 CPU cores
for parallel data loading.
Although it takes more time (roughly 1.6 times) 
to compute one forward pass for 100f, a larger number of frames are 
processed per second. C3D{~\cite{tran_c3d}} reports 313fps 
for a 16f network while using a larger number of parameters. Our proposed 
solution is therefore computationally efficient.

\vspace{-0.2cm}
\section{Conclusions}

This paper introduces and evaluates long-term temporal convolutions (LTC)
and shows that they can significantly improve the
performance. Using space-time convolutions over a large number of video frames, we obtain state of the art performance on two action recognition datasets UCF101 and HMDB51. We also demonstrate the impact of the optical flow quality. In the
presence of limited training data, using flow improves over RGB and
the quality of the flow impacts the results significantly.

\vspace{-.2cm}
\section*{Acknowledgements}
This work was supported by the ERC starting grant \mbox{ACTIVIA}, the ERC advanced grant \mbox{ALLEGRO}, Google and Facebook Research Awards and the MSR-Inria joint lab.

\vspace{-.1cm}
\bibliographystyle{IEEEtran}
\bibliography{references}

\begin{thebibliography}{10}
\providecommand{\url}[1]{#1}
\csname url@samestyle\endcsname
\providecommand{\newblock}{\relax}
\providecommand{\bibinfo}[2]{#2}
\providecommand{\BIBentrySTDinterwordspacing}{\spaceskip=0pt\relax}
\providecommand{\BIBentryALTinterwordstretchfactor}{4}
\providecommand{\BIBentryALTinterwordspacing}{\spaceskip=\fontdimen2\font plus
\BIBentryALTinterwordstretchfactor\fontdimen3\font minus
  \fontdimen4\font\relax}
\providecommand{\BIBforeignlanguage}[2]{{%
\expandafter\ifx\csname l@#1\endcsname\relax
\typeout{** WARNING: IEEEtran.bst: No hyphenation pattern has been}%
\typeout{** loaded for the language `#1'. Using the pattern for}%
\typeout{** the default language instead.}%
\else
\language=\csname l@#1\endcsname
\fi
#2}}
\providecommand{\BIBdecl}{\relax}
\BIBdecl

\bibitem{Tversky02}
B.~Tversky, J.~Morrison, and J.~Zacks, ``On bodies and events,'' in \emph{The
  Imitative Mind}, A.~Meltzoff and W.~Prinz, Eds.\hskip 1em plus 0.5em minus
  0.4em\relax Cambridge University Press, 2002.

\bibitem{Laptev08}
I.~Laptev, M.~Marsza{\l}ek, C.~Schmid, and B.~Rozenfeld, ``Learning realistic
  human actions from movies,'' in \emph{CVPR}, 2008.

\bibitem{Niebles08}
J.~C. Niebles, H.~Wang, and L.~Fei-Fei, ``Unsupervised learning of human action
  categories using spatial-temporal words,'' \emph{IJCV}, vol.~79, no.~3, pp.
  299--318, 2008.

\bibitem{Schuldt04}
C.~Sch{\"u}ldt, I.~Laptev, and B.~Caputo, ``Recognizing human actions: a local
  {SVM} approach,'' in \emph{ICPR}, 2004.

\bibitem{wang_idt}
H.~Wang and C.~Schmid, ``Action recognition with improved trajectories,'' in
  \emph{ICCV}, 2013.

\bibitem{simonyan_twostream}
K.~Simonyan and A.~Zisserman, ``Two-stream convolutional networks for action
  recognition in videos,'' in \emph{NIPS}, 2014.

\bibitem{krizhevsky_imagenet}
A.~Krizhevsky, I.~Sutskever, and G.~E. Hinton, ``{ImageNet} classification with
  deep convolutional neural networks,'' in \emph{NIPS}, 2012.

\bibitem{Deng09}
J.~Deng, W.~Dong, R.~Socher, L.-J. Li, K.~Li, and L.~Fei-Fei, ``{ImageNet}: {A}
  large-scale hierarchical image database,'' in \emph{CVPR}, 2009.

\bibitem{Zhou14}
B.~Zhou, A.~Lapedriza, J.~Xiao, A.~Torralba, and A.~Oliva, ``Learning deep
  features for scene recognition using places database,'' in \emph{NIPS}, 2014.

\bibitem{Girshick14}
R.~Girshick, J.~Donahue, T.~Darrell, and J.~Malik, ``Rich feature hierarchies
  for accurate object detection and semantic segmentation,'' in \emph{CVPR},
  2014.

\bibitem{Taigman14}
Y.~Taigman, M.~Yang, M.~Ranzato, and L.~Wolf, ``{DeepFace}: {C}losing the gap
  to human-level performance in face verification,'' in \emph{CVPR}, 2014.

\bibitem{karpathy_sports1m}
A.~Karpathy, G.~Toderici, S.~Shetty, T.~Leung, R.~Sukthankar, and L.~Fei-Fei,
  ``Large-scale video classification with convolutional neural networks,'' in
  \emph{CVPR}, 2014.

\bibitem{tran_c3d}
D.~Tran, L.~Bourdev, R.~Fergus, L.~Torresani, and M.~Paluri, ``Learning
  spatiotemporal features with {3D} convolutional networks,'' in \emph{ICCV},
  2015.

\bibitem{Donahue15}
J.~Donahue, L.~A. Hendricks, S.~Guadarrama, M.~Rohrbach, S.~Venugopalan,
  K.~Saenko, and T.~Darrell, ``Long-term recurrent convolutional networks for
  visual recognition and description,'' in \emph{CVPR}, 2015.

\bibitem{Ji10}
S.~Ji, W.~Xu, M.~Yang, and K.~Yu, ``{3D} convolutional neural networks for
  human action recognition,'' in \emph{ICML}, 2010.

\bibitem{Taylor10}
G.~W. Taylor, R.~Fergus, Y.~LeCun, and C.~Bregler, ``Convolutional learning of
  spatio-temporal features,'' in \emph{ECCV}, 2010.

\bibitem{projectpage}
\url{http://www.di.ens.fr/willow/research/ltc/}.

\bibitem{Csurka04}
G.~Csurka, C.~Dance, L.~Fan, J.~Willamowski, and C.~Bray, ``Visual
  categorization with bags of keypoints,'' in \emph{ECCVW}, 2004.

\bibitem{Perronnin10}
F.~Perronnin, J.~S{\'a}nchez, and T.~Mensink, ``Improving the {Fisher} kernel
  for large-scale image classification,'' in \emph{ECCV}, 2010.

\bibitem{Fernando15}
B.~Fernando, E.~Gavves, J.~Oramas, A.~Ghodrati, and T.~Tuytelaars, ``Modeling
  video evolution for action recognition,'' in \emph{CVPR}, 2015.

\bibitem{lecun-89e}
Y.~LeCun, B.~Boser, J.~S. Denker, D.~Henderson, R.~Howard, W.~Hubbard, and
  L.~Jackel, ``Backpropagation applied to handwritten zip code recognition,''
  \emph{Neural Computation}, vol.~1, no.~4, pp. 541--551, 1989.

\bibitem{wang_tdd}
L.~Wang, Y.~Qiao, and X.~Tang, ``Action recognition with trajectory-pooled
  deep-convolutional descriptors,'' in \emph{CVPR}, 2015.

\bibitem{wang_deeptwostream}
L.~Wang, Y.~Xiong, Z.~Wang, and Y.~Qiao, ``Towards good practices for very deep
  two-stream convnets,'' in \emph{arXiv:1507.02159}, 2015.

\bibitem{bilen_dynamic}
H.~Bilen, B.~Fernando, E.~Gavves, A.~Vedaldi, and S.~Gould, ``Dynamic image
  networks for action recognition,'' in \emph{CVPR}, 2016.

\bibitem{feichtenhofer_twostreamfusion}
C.~Feichtenhofer, A.~Pinz, and A.~Zisserman, ``Convolutional two-stream network
  fusion for video action recognition,'' in \emph{CVPR}, 2016.

\bibitem{zhang_emv}
B.~Zhang, L.~Wang, Z.~Wang, Y.~Qiao, and H.~Wang, ``Real-time action
  recognition with enhanced motion vector {CNNs},'' in \emph{CVPR}, 2016.

\bibitem{kantorov_efficient}
V.~Kantorov and I.~Laptev, ``Efficient feature extraction, encoding, and
  classification for action recognition,'' in \emph{CVPR}, 2014.

\bibitem{Farneback}
G.~Farneb\"{a}ck, ``Two-frame motion estimation based on polynomial
  expansion,'' in \emph{SCIA}, 2003.

\bibitem{Brox2004}
T.~Brox, A.~Bruhn, N.~Papenberg, and J.~Weickert, ``High accuracy optical flow
  estimation based on a theory for warping,'' in \emph{ECCV}, 2004.

\bibitem{UCF101}
K.~Soomro, A.~Roshan~Zamir, and M.~Shah, ``{UCF101}: A dataset of 101 human
  actions classes from videos in the wild,'' in \emph{CRCV-TR-12-01}, 2012.

\bibitem{HMDB51}
H.~Kuehne, H.~Jhuang, E.~Garrote, T.~Poggio, and T.~Serre, ``{HMDB}: a large
  video database for human motion recognition,'' in \emph{ICCV}, 2011.

\bibitem{MIFS}
Z.-Z. Lan, M.~Lin, X.~Li, A.~G. Hauptmann, and B.~Raj, ``Beyond {Gaussian}
  pyramid: Multi-skip feature stacking for action recognition.'' in
  \emph{CVPR}, 2015.

\bibitem{yue_beyondshort}
J.~Y. Ng, M.~J. Hausknecht, S.~Vijayanarasimhan, O.~Vinyals, R.~Monga, and
  G.~Toderici, ``Beyond short snippets: Deep networks for video
  classification,'' in \emph{CVPR}, 2015.

\bibitem{wang_transformations}
X.~Wang, A.~Farhadi, and A.~Gupta, ``Actions {\textasciitilde}
  transformations,'' in \emph{CVPR}, 2016.

\bibitem{zeiler_vis}
M.~D. Zeiler and R.~Fergus, ``Visualizing and understanding convolutional
  networks,'' in \emph{ECCV}, 2014.

\end{thebibliography}
\end{document}